\RequirePackage{snapshot}
\documentclass[english,10pt,twocolumn,letterpaper]{article}
\usepackage[T1]{fontenc}
\usepackage[latin9]{inputenc}
\usepackage{color}
\usepackage{array}
\usepackage{float}
\usepackage{multirow}
\usepackage{amsmath}
\usepackage{amssymb}
\usepackage{graphicx}
\usepackage[numbers]{natbib}
\usepackage{pbox}
\usepackage{etoolbox}
\usepackage{fancyhdr}
\usepackage[nomessages]{fp}
\patchcmd{\chapter}{plain}{fancy}{}{}

\definecolor{gray}{rgb}{0.5, 0.5, 0.5}

\newtoggle{arxiv}
\toggletrue{arxiv}

\newtoggle{arxiv_flatten}
\toggletrue{arxiv_flatten}

\usepackage{pifont}\newcommand{\cmark}{\ding{51}}\newcommand{\xmark}{\ding{55}}\renewcommand{\cmark}{Yes}\renewcommand{\xmark}{No}
\usepackage{etoc}
\usepackage{pdflscape} 

\usepackage{tocloft}

\makeatletter

\providecommand{\tabularnewline}{\\}

\usepackage{titlesec}

\iftoggle{arxiv}{
    \newcommand{\optvspace}[1]{}
    \newcommand{\arxvspace}[1]{\vspace{#1}}
}{
    \newcommand{\optvspace}[1]{\vspace*{#1}}
    \newcommand{\arxvspace}[1]{}
}

\iftoggle{arxiv}{\usepackage{cvpr}
\titlespacing*{\paragraph}{0pt}{0.2ex plus 0.05ex}{1.5ex}
}{\usepackage{cvpr-compact}
}

\usepackage{times}
\usepackage{epsfig}
\usepackage{graphicx}
\usepackage{amsmath}
\usepackage{amssymb}
\usepackage{subfig}

\usepackage[dvipsnames]{colortbl}

\usepackage{setspace}

\usepackage{enumitem}

\setlist{noitemsep, nolistsep}

\usepackage[pagebackref=true,breaklinks=true,letterpaper=true,colorlinks,bookmarks=false]{hyperref}

\cvprfinalcopy 
\def\cvprPaperID{178} 

\iftoggle{arxiv}{    \fancypagestyle{firststyle}{
    \fancyfoot[L]{\footnotesize 
      \textsf{This paper is published in IEEE CVPR~2017.}
    }
  }
}{  \fancypagestyle{firststyle}{}
\ifcvprfinal\fancyfoot{}\pagestyle{empty}\fi
}

\newcommand{\lowreswarn}{\textsf{\textcolor{gray}{Due to the file size limit on arXiv submissions, images in this file are significantly downsampled. The high-resolution PDF is available at}}\\
\url{http://ytzhang.net/files/publications/2017-cvpr-dbnet-highres.pdf} 
}

\iftoggle{arxiv_flatten}{
\fancyfoot[C]{\thepage \\ \footnotesize \vspace*{1em} \lowreswarn}
}
{}

\definecolor{cellhl}{rgb}{0.95, 0.95, 0.96}

\newcommand{\colcut}{\hspace{-0.3em}}

\usepackage[font=small,labelfont=bf]{caption}

\newcommand{\filluptopage}[1]{  \clearpage
  \loop\ifnum\value{page}<#1\relax
    \null\clearpage
  \repeat
}

\iftrue  
            \newcommand{\todo}[1]{}
        \newcommand{\outline}[1]{}
        \newcommand{\textgray}[1]{}
        \newcommand{\commenttext}[1]{}
        \newcommand{\commentfoot}[1]{}
        \newcommand{\commentselfoot}[2]{}
        \newcommand{\commentselrep}[2]{}
        \newcommand{\topic}[1]{}
        \newcommand{\commenthl}[1]{}
\else 
            \newcommand{\todo}[1]{{\textcolor{red}{[[TODO: {#1}]]}}}
        \newcommand{\outline}[1]{{\textcolor{blue}{[[{#1}]]}}}
        \newcommand{\textgray}[1]{\textcolor{gray}{[[{#1}]]}}
        \newcommand{\commenttext}[1]{\textcolor{red}{[[{#1}]]}}
        \newcommand{\commentfoot}[1]{\footnote{\textcolor{red}{\textit{#1}}}}
        \newcommand{\commentselfoot}[2]{{\textcolor{blue}{#1}}\commenttext{#2}}
        \newcommand{\commentselrep}[2] {{\textcolor{blue}{#1}} {\textcolor{green}{[[\textit{#2}]]}}}
        \newcommand{\topic}[1]{\textcolor{gray}{\textbf{(#1.)}}}
        \newcommand{\commenthl}[1]{\textcolor{blue}{[HL: #1]}}
\fi

\definecolor{darkgreen}{rgb}{0.0,0.70,0}
\definecolor{darkyellow}{rgb}{0.9,0.60,0}

\newcommand{\unitcut}{}

\usepackage{caption}

\newcommand{\repeatcaption}[2]{  \FPeval{\previousfigure}{clip(\thefigure)}
  \caption{(continued from Figure~{\previousfigure}) #2}}

\@ifundefined{showcaptionsetup}{}{ \PassOptionsToPackage{caption=false}{subfig}}
\usepackage{subfig}
\makeatother

\iftoggle{arxiv}{
\newcommand{\mysubfloatskip}{}
}{
\newcommand{\mysubfloatskip}{\vspace{-7pt}}
}

\usepackage{babel}
\begin{document}

\newcommand{\mytitle}{Discriminative Bimodal Networks for \\
Visual Localization and Detection with Natural Language Queries}
\title{\mytitle}

\author{Yuting Zhang, Luyao Yuan, Yijie Guo, Zhiyuan He, I-An Huang, Honglak Lee \vspace{0.2em}
\\
University of Michigan, Ann Arbor, MI, USA \vspace{-0.2em}\\
{\small{} \texttt{\{yutingzh, yuanluya, guoyijie, zhiyuan, huangian, honglak\}@umich.edu}}}
\maketitle

\optvspace{-0.1in}
\begin{abstract}
	\optvspace{-0.15in}
		Associating image regions with text queries has been recently explored as a new way to bridge visual and linguistic representations.
				A few pioneering approaches have been proposed based on recurrent neural language models trained generatively (e.g., generating captions), but achieving somewhat limited localization accuracy. 
		To better address natural-language-based visual entity localization, we propose a discriminative approach. 
        We formulate a discriminative bimodal neural network (\textbf{DBNet}), which can be trained by a classifier with extensive use of negative samples.
			Our training objective encourages better localization on single images, incorporates text phrases in a broad range, and properly pairs image regions with text phrases into positive and negative examples. 
		Experiments on the Visual Genome dataset demonstrate the proposed DBNet significantly outperforms previous state-of-the-art methods both for localization on single images and for detection on multiple images. 
		We we also establish an evaluation protocol for natural-language visual detection. 
    \iftoggle{arxiv}{
}{
    Code is available at: \url{http://ytzhang.net/projects/dbnet}~.
}
\optvspace{-0.3in}
\end{abstract}
 
\etocdepthtag.toc{mtchapter}
\etocsettagdepth{mtchapter}{subsection}
\etocsettagdepth{mtappendix}{none}

\allowdisplaybreaks

\thispagestyle{firststyle}

\unitcut\unitcut\unitcut
\section{Introduction}
\unitcut

Object localization and detection in computer vision are traditionally
limited to a small number of predefined categories (e.g., car, dog, and person),
and category-specific image region classifiers \citep{hog,dpm,rcnn}
serve as object detectors. However, in the real world, the \emph{visual
entities} of interest are much more diverse, including groups of objects
(involved in certain relationships), object parts, and objects with
particular attributes and/or in particular context. 
For scalable annotation, these entities need to be labeled in a more flexible way, such as using text phrases. 

Deep learning has been demonstrated as a unified learning
framework for both text and image representations. 
Significant progress has been made in many related tasks, such as image captioning \citep{show-tell,show-attend-tell,deep-align-gen,m-rnn,mind-eye,conv-lstm-caption,densecap,novel-caption,unambiguous-cap},
visual question answering \citep{vqa,ask-neuron,attend-vqa,dynamic-vqa,module-net},
text-based fine-grained image classification \citep{fine-description},
natural-language object retrieval \citep{natural-obj,unambiguous-cap},
and text-to-image generation \citep{text2image}. 

A few pioneering works \citep{natural-obj,unambiguous-cap} use recurrent neural language models \citep{rnn-gen,rnn-lang,gen-text} and deep image representations \citep{alexnet,vggnet} for localizing the object referred to by a text phrase given a single image (i.e., ``object referring" task~\citep{refer-it}). 
Global spatial context, such as ``a man on the left (of the image)'', has been commonly used to pick up the particular object. 
In contrast, \citet{densecap} takes descriptions without global context\footnote{Only a very small portion of text phrases on the Visual Genome refer to the global context. } as queries for localizing more general visual entities on the Visual Genome dataset~\citep{vg}.

\begin{figure}
\begin{centering}
\includegraphics[width=1\columnwidth]{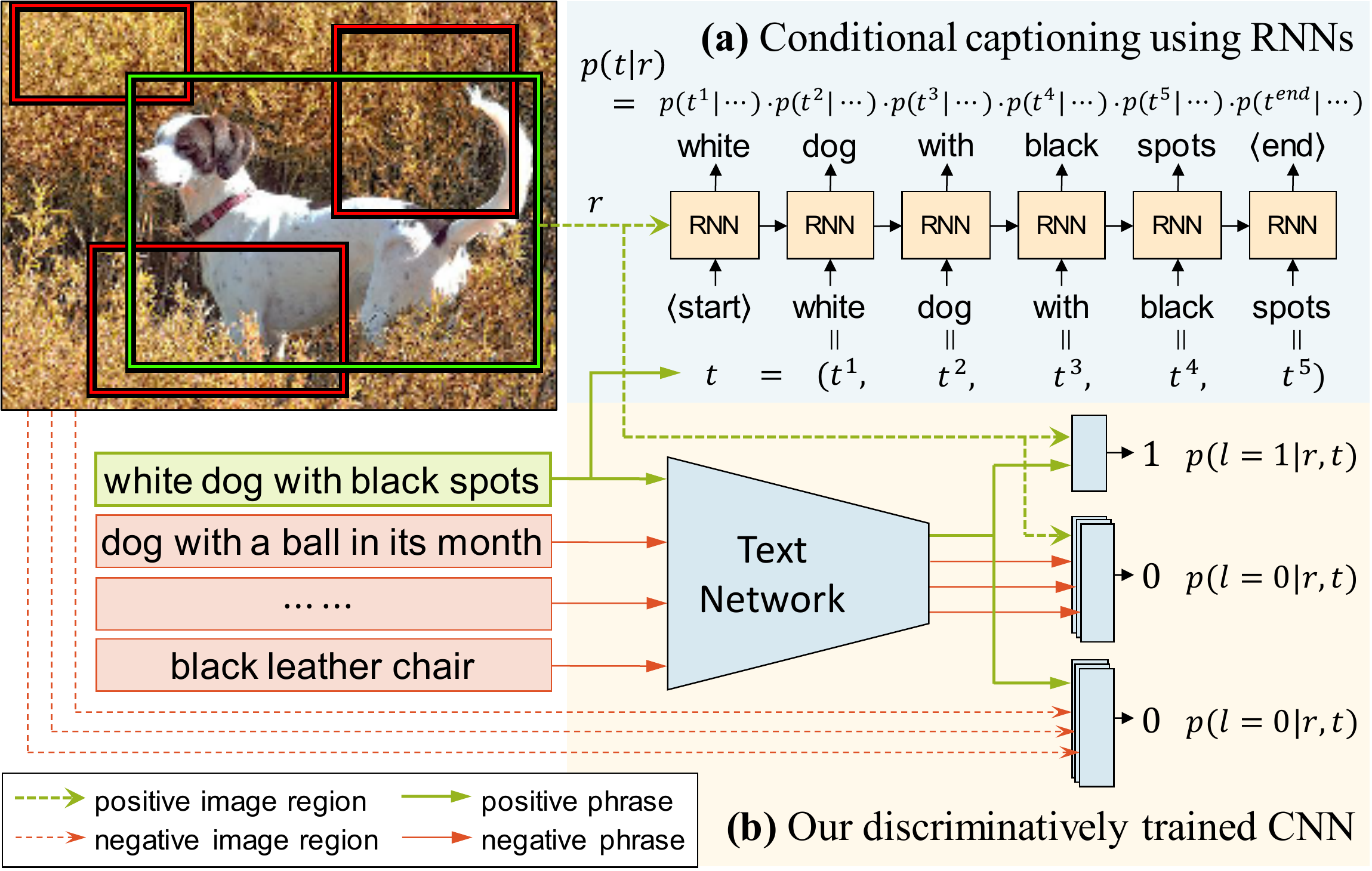}
\par\end{centering}
\unitcut\unitcut
\caption{Comparison between (a) image captioning model and (b) our discriminative architecture
for visual localization. \label{fig:cap-vs-feat}}
\unitcut\unitcut\unitcut\unitcut
\end{figure}

All above existing work performs localization by maximizing the likelihood
to generate the query text given image regions using an image captioning
model (Figure~\ref{fig:cap-vs-feat}a), whose output probability
density needs to be modeled on the virtually infinite space of the natural language. 
Since it is hard to train a classifier on such a huge
structured output space, current captioning models are constrained
to be trained in generative \citep{natural-obj,densecap} or partially discriminative \citep{unambiguous-cap} ways. 
However, as discriminative tasks, localization and detection usually favor models that are trained with a more discriminative objective to better utilize negative samples. 
In this paper, we propose a new deep architecture for natural-language-based visual entity localization, which we call a \emph{discriminative bimodal network} (\textbf{DBNet}). Our architecture uses a binary output space to allow extensive discriminative training, where any negative training sample can be potentially utilized. 
The key idea is to take the text query
as a condition rather than an output and to let the model directly
predict if the text query and image region are compatible (Figure~\ref{fig:cap-vs-feat}b).
In particular, the two pathways of the deep architecture respectively
extract the visual and linguistic representations. 
A discriminative pathway is built upon the two pathways to fuse the bimodal representations for binary classification of the inter-modality compatibility. 

Compared to the estimated probability density in the huge space of the natural language,
the score given by a binary classifier is more likely to be \emph{calibrated}.
In particular, better calibrated scores should be more comparable
across different images and text queries. 
This property makes it possible to learn decision thresholds to determine the existence of visual entities on multiple images and text queries, making the localization model generalizable for detection tasks. 
While a few examples of natural-language visual detection are showcased in \citep{densecap}, we perform more comprehensive quantitive and ablative evaluations. 

In our proposed architecture, we use convolutional
neural networks (CNNs) for both visual and textual representations.
Inspired by fast R-CNN \citep{fast-rcnn}, we use the RoI-pooling
architecture induced from large-scale image classification networks
for efficient feature extraction and model learning on image regions.
For textual representations, we develop a character-level CNN \citep{character-cnn}
for extracting phrase features. A network on top of the image and
language pathways \emph{dynamically} forms classifiers for image region features
 depending on the text features, and it outputs the classifier
responses on all regions of interest. 

Our main contributions are as follows: 
\begin{enumerate}
\item We develop a bimodal deep architecture with a binary output
space to enable fully discriminative training for natural-language
visual localization and detection. 
\item We propose a training objective that extensively pairs text phrases and bounding boxes, where 1)~the discriminative objective is defined over all possible region-text pairs in the entire training set, and 2)~the non-mutually exclusive nature of text phrases is taken into account to avoid ambiguous training samples. 
\item Experimental results on Visual Genome demonstrate that the proposed
DBNet significantly outperforms existing methods based on recurrent neural language models for visual entity localization on single images. 
\item We also establish evaluation methods for natural-language visual detection on multiple images and show state-of-the-art results.
\end{enumerate}
 
\unitcut\unitcut\unitcut\unitcut
\section{Related work}
\unitcut

\paragraph{Object detection.}

Recent success of deep learning on visual object recognition \citep{alexnet,zfnet,vggnet,googlenet,rethink-inception,resnet} constitutes the backbone of the state-of-the-art for object detection \citep{rcnn,overfeat,multibox,fgs-struct,deep-id,yolo,fast-rcnn,faster-rcnn,resnet,rfcn}.
Natural-language visual detection can adapt the deep visual representations and single forward-pass computing framework (e.g., RoI pooling \citep{fast-rcnn}, SPP \citep{spp}, R-FCN \citep{rfcn}) used in existing work of traditional object detection. 
However, natural-language visual detection needs a huge structured label space to represent the natural language, and finding a proper mapping to the huge space from visual representations is difficult.

\paragraph{Image captioning and caption grounding.}

The recurrent neural network (RNN)~\citep{lstm} based language model \citep{rnn-gen,rnn-lang,gen-text} has become the dominant method for captioning images with text \citep{show-tell}. 
Despite differences in details of network architectures, most RNN language models learn the likelihood of picking up a word from a predefined vocabulary given the visual appearance features and previous words (Figure~\ref{fig:cap-vs-feat}a).
\citet{show-attend-tell} introduced an attention mechanism to encourage RNNs to focus on relevant image regions when generating particular words. 
\citet{deep-align-gen} used strong supervision of text-region alignment for well-grounded captioning. 

\paragraph{Object localization by natural language.}

Recent work used the conditional likelihood of captioning an image region with given text for localizing associated objects. 
\citet{natural-obj} proposed the spatial-context recurrent ConvNet (SCRC), which conditioned on both local visual features and global contexts for evaluating given captions. 
\citet{densecap} combined captioning and object proposal in an end-to-end neural network, which can densely caption (DenseCap) image regions and localize objects.  
\citet{unambiguous-cap} trained the captioning model by maximizing the posterior of localizing an object given the text phrase, which reduced the ambiguity of generated captions. 
However, the training objective was limited to figuring out single objects on single images. 
\citet{visual-relationship} simplified and limited text queries to subject-relationship-object (SVO) triplets. 
\citet{grounder} improved localization accuracy with an extra text reconstruction task. 
\citet{nl-seg} extended bounding box localization to instance segmentation using natural language queries. 
\citet{context-in-referral} and \citet{context-for-referral} explicitly modeled context for referral expressions.

\paragraph{Text representation.}

Neural networks can also embed text into a fixed-dimensional feature space. 
Most RNN-based methods (e.g., skip-thought vectors~\citep{skip-thought}) and CNN-based methods~\citep{conv-sentence-model,conv-sentence-cls} use word-level one-hot encoding as the input.
Recently, character-level CNN has also been demonstrated an effective way for paragraph categorization~\citep{character-cnn} and zero-shot image classification~\citep{fine-description}.

\unitcut\unitcut
\section{Discriminative visual-linguistic network}
\unitcut

\commenttext{Do we need to draw a diagram for the two-pathway network?}

The best-performing object detection framework \citep{hog,dpm,rcnn}
in terms of accuracy generally verifies if a candidate image region
belongs to a particular category of interest.  Though recent deep
architectures \citep{multibox,faster-rcnn,densecap} can propose regions
with confidence scores at the same time, a verification model, taking
as input the image features from the exact proposed regions, still
serves as a key to boost the accuracy. 

In this section, we develop a verification model for natural-language
visual localization and detection. Unlike the classifiers for a small
number of predefined categories in traditional object detection, our
model is dynamically adaptable to different text phrases. 

\unitcut
\subsection{Model framework}
\unitcut
Let $x$ be an image, $r$ be the coordinates of a region, and $t$
be a text phrase. The verification model $f(x,t,r;\Theta)\in\mathbb{R}$
outputs the confidence of $r$'s being matched with $t$. Suppose that
$l\in\{1,0\}$ is the binary label indicating if $(t,r)$ is a 
positive or negative region-text pair on $x$. Our verification model learns to fit the probability for $r$ and $t$ being compatible (a
positive pair), i.e., $p(l=1|x,r,t)$.  See Section~\ref{sec:compare-captioning-eq}
in the supplementary materials for a formalized comparison with conditional
captioning models. 

\paragraph{}

To this end, we develop a bimodal deep neural network for our model. In particular,
$f(x,t,r;\Theta)$ is composed of two single-modality pathways followed by a discriminative pathway. 
The image pathway $\boldsymbol{\phi}_{\text{rgn}}(x,r;\Theta_{\text{rgn}})$
extracts the $d_{\text{rgn}}$-dim visual representation on the image
region $r$ on $x$. The language pathway $\boldsymbol{\phi}_{\text{txt}}(t;\Theta_{\text{txt}})$
extracts the $d_{\text{txt}}$-dim textual representation for the
phrase $t$. The discriminative pathway with parameters $\Theta_{\text{dis}}$
dynamically generates a classifier for visual representation according
to the textual representation, and predicts if $r$ and $t$ are matched
on $x$. The full model is specified by $\Theta=(\Theta_{\text{txt}},\Theta_{\text{rgn}},\Theta_{\text{dis}})$.

\unitcut
\subsection{Visual and linguistic pathways}
\unitcut

\paragraph{RoI-pooling image network.}

We suppose the regions of interest are given by an existing region proposal method (e.g., EdgeBox \citep{edgebox}, RPN \citep{faster-rcnn}).
We calculate visual representations for all image regions in one pass using the fast R-CNN RoI-pooling pipeline. 
State-of-the-art image classification networks, including the 16-layer VGGNet \citep{vggnet} and ResNet-101 \citep{resnet}, are used as backbone architectures. 

\paragraph{Character-level textual network.}

For an English text phrase $t$, we encode each of its characters
into a 74-dim one-hot vector, where the alphabet is composed of 74
printable characters including punctuations and the space. Thus, the
$t$ is encoded as a 74-channel sequence by stacking all character
encodings. We use a character-level deep CNN \citep{character-cnn}
to obtain the high-level textual representation of $t$. In particular,
our network has 6 convolutional layers interleaving with 3 max-pooling
layers and followed by 2 fully connected layers (see Section~\ref{sec:textual-pathway}
in the supplementary materials for more details). 
It takes a sequence of a fixed length as the input and produces textual representations of a fixed dimension. 
The input length is set to be long enough (here, 256 characters) to cover possible text phrases.\footnote{The Visual Genome dataset has more than 2.8M unique phrases, whose median length in character is 29. Less than 500 phrases has more than 100 characters.} To avoid empty tailing characters in the input, we replicate the
text phrase until reaching the input length limit. 

We empirically found that the very sparse input can easily lead to over-sparse intermediate
activations, which can create a large portion of ``dead'' ReLUs
and finally result in a degenerate solution. To avoid this problem,
we adopt the Leaky ReLU (LReLU) \citep{lrelu} to keep all hidden
units active in the character-level CNN. 

Other text embedding methods \citep{skip-thought,conv-sentence-model,conv-sentence-cls} also can be used in the DBNet framework. 
We use the character-level CNN because of its simplicity and flexibility. 
Compared to word-based models, it uses lower-dimensional input vectors and has no constraint on the word vocabulary size. 
Compared to RNNs, it easily allows deeper architectures. 

\unitcut
\subsection{Discriminative pathway}
\label{sec:dis-pathway}
\unitcut

The discriminative pathway first forms a linear classifier using the textual representation of the phrase $t$. Its linear combination weights and bias are 
\begin{align}
\mathbf{w}(t) & =\mathbf{A}_{\mathbf{w}}^{\top}\boldsymbol{\phi}_{\text{txt}}(t;\Theta_{\text{txt}}),\label{eq:dy-kernel}\\
b(t) & =\mathbf{a}_{b}^{\top}\boldsymbol{\phi}_{\text{txt}}(t;\Theta_{\text{txt}}),\label{eq:dy-bias}
\end{align}
where $\mathbf{A}_{\mathbf{w}}\in\mathbb{R}^{d_{\text{txt}}\times d_{\text{rgn}}}$,
$\mathbf{a}_{\mathbf{b}}\in\mathbb{R}^{d_{\text{txt}}}$, and $\Theta_{\text{dis}}=(\mathbf{A}_{\mathbf{w}},\mathbf{a}_{b})$.
This classifier is applied to the visual representation of the image
region $r$ on $x$, obtaining the verification confidence predicted
by our model: 
\begin{equation}
f(x,r,t;\Theta)=\mathbf{w}(t)^{\top}\boldsymbol{\phi}_{\text{rgn}}(x,r;\Theta_{\text{rgn}})+b(t).\label{eq:dy-classifier}
\end{equation}
Compared to the basic form of the bilinear function $\boldsymbol{\phi}_{\text{txt}}^{\top}(t;\Theta_{\text{txt}})\mathbf{A}_{\mathbf{w}}\boldsymbol{\phi}_{\text{rgn}}(x,r;\Theta_{\text{rgn}})$,
our discriminative pathway includes an additional linear term as the text-dependent
bias for the visual representation classifier. 

As a natural way for modeling the cross-modality correlation, multiplication is also a source of instability for training. To improve
the training stability, we introduce a regularization term $\Gamma_{\text{dynamic}}=\Vert\mathbf{w}(t)\Vert_{2}^{2}+\vert b(t)\vert^{2}$
for the dynamic classifier, besides the network weight decay
$\Gamma_{\text{decay}}$ for $\Theta$.
 
\unitcut\unitcut
\section{Model learning}
\unitcut

In DBNet, we drive the training of the proposed two-pathway bimodal CNN with a binary classification objective. We pair image regions and text phrases as training samples. 
We define the ground truth binary label for each training region-text pair (Section~\ref{sec:gt-label}), and propose a weighted training loss function (Section~\ref{sec:training-loss}).

\paragraph{Training samples.}

Given $M$ training images $x_{1},x_{2},\ldots,$ $x_{M}$, let $\mathcal{G}_{i}=\{(r_{ij},t_{ij})\}_{j=1}^{N_{i}}$
be the set of ground truth annotations for $x_{i}$, where $N_{i}$
is the number of annotations, $r_{ij}$ is the coordinate
of the $j$\textsuperscript{th} region, and $t_{ij}$
is the text phrase corresponding to $r_{ij}$. When one region is
paired with multiple phrases, we take each pair as a separate
entry in $\mathcal{G}_{i}$. 

We denote the set of all regions considered on $x_{i}$ by $\mathcal{R}_{i}$,
which includes both annotated regions $\bigcup_{j=1}^{N_{i}}\{r_{ij}\}$
and regions given by proposal methods \citep{selective-search,edgebox,faster-rcnn}.
We write $\mathcal{T}_{i}=\bigcup\{t_{ij}\}_{j=1}^{N_{i}}$ for the
set of annotated text phrases on $x_{i}$, and $\mathcal{T}=\bigcup_{i=1}^{M}\mathcal{T}_{i}$
for all training text phrases.  

\unitcut
\subsection{Ground truth labels}
\label{sec:gt-label}
\unitcut

\begin{figure}
\begin{centering}
\includegraphics[width=1\columnwidth]{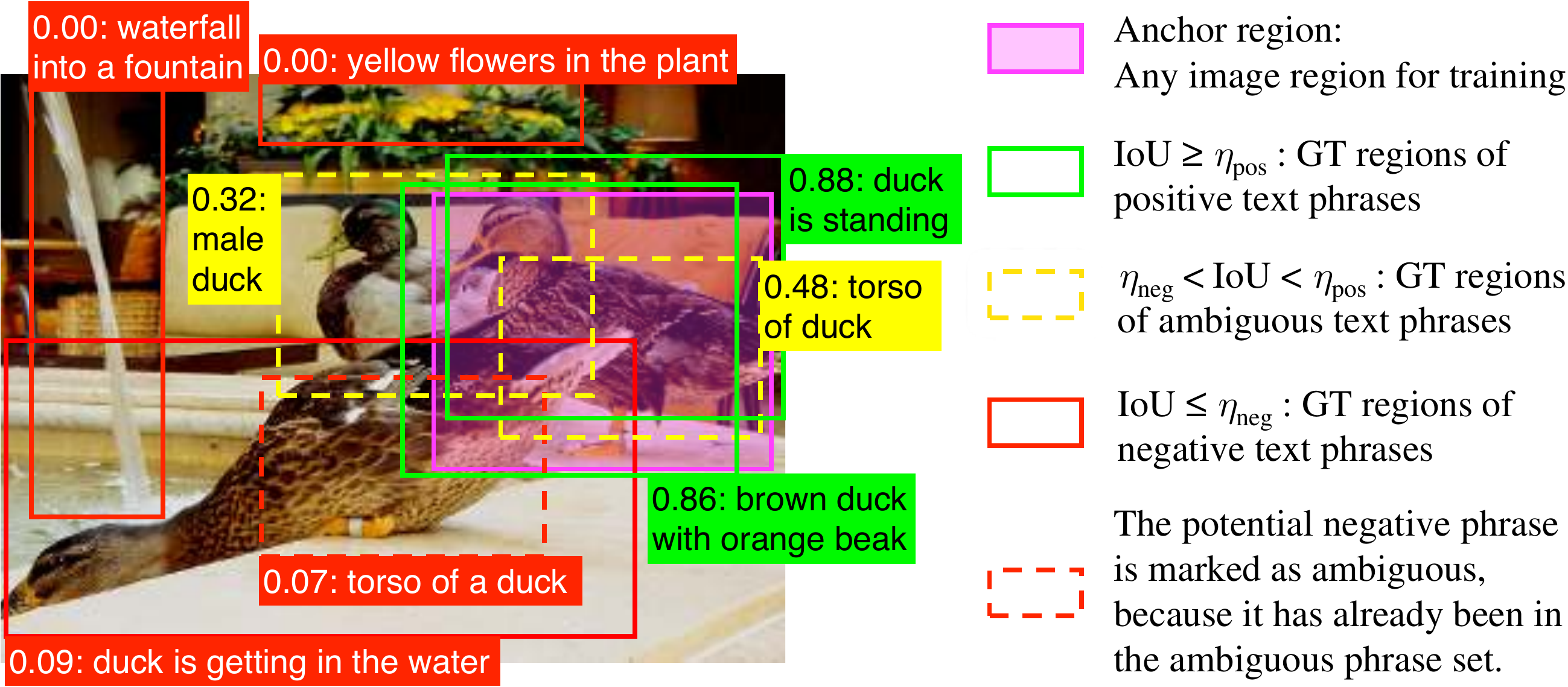}
\par\end{centering}
\caption{Ground truth labels for region-text pairs (given an arbitrary image region). 
Phrases are categorized into positive, ambiguous, and negative sets based on the given region's overlap with ground truth boxes (measured by IoU and displayed as the numbers in front of the text phrases).
Ambiguous phrases augmented by text similarity is not shown here (see the video in the supplementary materials for an illustration). 
For visual clarity, $\eta_{\text{neg}}=0.3$ and $\eta_{\text{pos}}=0.7$, which are different from the rest of the paper.
}

\label{fig:per-region}
\end{figure}

\paragraph{Labeling criterion.}

We assign each possible training region-text pair with a ground truth label for binary classification. 
For a region $r$ on the image $x_i$  and a text phrase $t\in \mathcal{T}_i$, we take the largest overlap between $r$ and $t$'s ground truth regions as evidence to determine $(r,t)$'s label.
Let $\operatorname{IoU}(\cdot,\cdot)$ denote the intersection over union.  The \emph{largest overlap} is defined as
\begin{equation}
\nu_{i}(r,t)=\max_{r'\in\mathcal{R}_{i}}\{\operatorname{IoU}(r',r):(r',t)\in\mathcal{G}_{i}\}. \label{eq:best-iou}
\end{equation}
In object detection on a limited number of categories (i.e., $\mathcal{T}_i$ consists of category labels), $\nu_i(r,t)$ is usually reliable enough for assigning binary training labels, given the (almost) complete ground truth annotations for all categories. 

In contrast, text phrase annotations are inevitably incomplete in the training set. 
One image region can have an intractable number of valid textual descriptions, including different points of focus and paraphrases of the same description, so annotating all of them is infeasible. 
Consequently, $\nu_i(r,t)$ cannot always reflect the consistency between an image region and a text phrase. 
To obtain reliable training labels, we define positive labels in a conservative manner; 
and then, we combine text similarity together with spatial IoU to establish the ambiguous text phrase set that reflects potential ``false negative'' labels. 
We provide detailed definitions below.

\paragraph{Positive phrases.}

For a region $r$ on $x_i$, its positive text phrases (i.e., phrases assigned with positive labels) constitute the set
\begin{equation}
\mathcal{P}_{i}(r)=\{t\in\mathcal{T}_{i}:\nu_{i}(r,t)\ge\eta_{\text{pos}}\}, \label{eq:pos-txt}
\end{equation}
where $\eta_{\text{pos}}$ is a high enough IoU threshold ($=0.9$) to determine positive labels. 
Some positive phrases may be missing due to incomplete annotations.
However, we do not try to recover them (e.g., using text similarity), as ``false positive'' training labels may be introduced by doing so. 

\paragraph{Ambiguous phrases.}

Still for the region $r$, we collect the text phrases whose ground truth regions have moderate (neither too large nor too small) overlap with $r$ into a set  
\begin{equation}
\mathcal{U}_{i}(r)=\{t\in\mathcal{T}_{i}:\eta_{\text{neg}}<\nu_{i}(r,t)<\eta_{\text{pos}}\}, \label{eq:nearby-txt}
\end{equation}
where $\eta_{\text{neg}}$ is the IoU lower bound ($=0.1$). 
When $r$'s largest IoU with the ground truths of a phrase $t$ lies in $(\eta_{\text{neg}},\eta_{\text{pos}})$, it is uncertain whether $t$ is positive or negative. In other words, $t$ is \emph{ambiguous} with respect to the region $r$. 

Note that $\mathcal{U}_{i}(r)$ only contains phrases from $\mathcal{T}_i$. 
To cover all possible ambiguous phrases from the full set $\mathcal{T}$, we use a text similarity measurement $\operatorname{sim}(\cdot,\cdot)$ to augment $\mathcal{U}_{i}(r)$ to the finalized ambiguous phrase set  
\begin{equation}
\mathcal{A}_{i}(r)=\{t\in\mathcal{T}:\exists t'\in\mathcal{U}_{i}(r),\operatorname{sim}(t,t')>\tau\}\backslash\mathcal{P}_{i}(r), \label{eq:amb-txt}
\end{equation}
where we use the METEOR~\citep{meteor} similarity for $\operatorname{sim}(\cdot,\cdot)$ and set the text similarity threshold $\tau=0.3$.\footnote{If the METEOR similarity of two phrases is greater than 0.3, they are usually very similar. In Visual Genome, $\sim$0.25\% of all possible pairs formed by the text phrases that occur $\ge$20 times can pass this threshold.}

\paragraph{Labels for region-text pairs.}

For any image region $r$ on $x_i$ and any phrase $t \in \mathcal{T}$, the ground truth label of $(r,t)$ is
\begin{equation}
y_{i}(r,t)=\begin{cases}
1, & t\in\mathcal{P}_{i}(r),\\
\mathtt{\langle uncertain\rangle}, & t\in\mathcal{A}_{i}(r),\\
0, & \text{otherwise,}
\end{cases}\label{eq:gt-label}
\end{equation}
where the pairs of a region and its ambiguous text phrases are assigned with the ``uncertain'' label to avoid false negative labels.  
Figure~\ref{fig:per-region} illustrates the region-text label for an arbitrary training image region.

\unitcut
\subsection{Weighted training loss}
\label{sec:training-loss}
\unitcut

\commenttext{Do we need to use boldface for r and t? If so, also update Figure 1}

\paragraph{Effective training sets.}

On the image $x_{i}$, 
the effective set of training region-text pairs is 
\begin{equation}
\mathcal{S}_{i}=\{(r,t)\in\mathcal{R}_{i}\times\mathcal{T}:y_{i}(r,t)\neq\mathtt{\langle uncertain\rangle}\},\label{eq:effective-set}
\end{equation}
where, as previously defined, $\mathcal{R}_{i}$ consists of annotated and proposed regions, and $\mathcal{T}$ consists of all phrases from the training set. 
We exclude samples of uncertain labels. 

We partition $\mathcal{S}_{i}$ into three subsets according to the value of $y_{i}(r,t)$ and the origin of the phrase $t$: $\mathcal{S}_{i}^{\text{pos}}$ for $y_i(r,t)=1$, $\mathcal{S}_{i}^{\text{neg}}$ for $y_i(r,t)=0 \land t \in \mathcal{T}_i$, and $\mathcal{S}_{i}^{\text{rest}}$ for all negative region-text pairs containing phrases from the rest of the training set (i.e., not from $x_i$).

\paragraph{Per-image training loss}

Let $f_{i}(r,t)=f(x_{i},r,t;\Theta)\in\mathbb{R}$ for notation convenience; 
and, let $\ell(\cdot,\cdot)$ be a binary classification loss, in particular, the cross-entropy loss of logistic regression. 
We define the training loss on $x_{i}$ as the summation of three parts:
\begin{equation}
L_{i}=\lambda_{\text{pos}}L_{i}^{\text{pos}}+\lambda_{\text{neg}}L_{i}^{\text{neg}}+\lambda_{\text{rest}}L_{i}^{\text{rest}},\label{eq:loss-decomp}
\end{equation}
\begin{align}
L_{i}^{\text{pos}} & =\frac{1}{|\mathcal{S}_{i}^{\text{pos}}|}\sum_{(r,t)\in\mathcal{S}_{i}^{\text{pos}}}\ell\left(f_{i}(r,t),1\right),\label{eq:loss-pos}\\
L_{i}^{\text{neg}} & =\frac{1}{|\mathcal{S}_{i}^{\text{neg}}|}\sum_{(r,t)\in\mathcal{S}_{i}^{\text{neg}}}\ell\left(f_{i}(r,t),0\right),\label{eq:loss-neg}\\
L_{i}^{\text{rest}} & =\frac{\sum_{(r,t)\in\mathcal{S}_{i}^{\text{rest}}}\mathrm{freq}(t)\cdot\ell\left(f_{i}(r,t),0\right)}{\sum_{(r,t)\in\mathcal{S}_{i}^{\text{rest}}}\mathrm{freq}(t)},\label{eq:loss-rest}
\end{align}
where $\mathrm{freq}(t)$ is $t$'s frequency of occurrences in the training set. 
We normalize and re-weight the loss for each of the three subsets of $\mathcal{S}_i$ separately.
In particular, we set $\lambda_{\text{pos}}=\lambda_{\text{neg}}+\lambda_{\text{rest}}=1$ to balance the positive and negative training loss. 
The values of $\lambda_{\text{neg}}$ and $\lambda_{\text{rest}}$ are implicitly determined by the numbers of text phrases that we choose inside and outside $x_i$ during stochastic optimization. 

The training loss functions in most existing work on natural-language visual localization \citep{natural-obj,densecap} use only positive samples for training, which is similar to solely using $L_{i}^{\text{pos}}$.
The method in \citep{unambiguous-cap} also considers the negative case (similar to $L_{i}^{\text{neg}}$), but it is less flexible and not extensible to the case of $L_{i}^{\text{rest}}$. 
The recurrent neural language model can encourage a certain amount of discriminativeness on word selection, but not on entire text phrases as ours.

\begin{table*}
\vspace*{-0.1in}
\begingroup
\renewcommand{\colcut}{\hspace{0em}}
\small{} 
\begin{centering}
\par\end{centering}
\begin{centering}
\begin{tabular}{>{\raggedright}p{4em}|>{\raggedright}p{5em}|>{\colcut}l<{\colcut}|>{\colcut}c<{\colcut}>{\colcut}c<{\colcut}>{\colcut}c<{\colcut}>{\colcut}c<{\colcut}>{\colcut}c<{\colcut}>{\colcut}c<{\colcut}>{\colcut}c<{\colcut}|>{\colcut}c<{\colcut}>{\colcut}c<{\colcut}}
\hline 
Region & Visual & Localization & \multicolumn{7}{c|}{Recall / \% for IoU@} & Median  & Mean \tabularnewline
\cline{4-10} 
proposal & network &  model & 0.1 & 0.2 & 0.3 & 0.4 & 0.5 & 0.6 & 0.7 & IoU & IoU\tabularnewline
\hline 
\hline 
\multirow{2}{4em}{DC-RPN 500} & \multirow{2}{5em}{16-layer VGGNet} & DenseCap  & 52.5 & 38.9 & 27.0 & 17.1 & \textcolor{white}{0}9.5 & \textcolor{white}{0}4.3 & \textcolor{white}{0}1.5 & 0.117 & 0.184\tabularnewline
 &  & DBNet & \textbf{57.4} & \textbf{46.9} & \textbf{37.8} & \textbf{29.4} & \textbf{21.3} & \textbf{13.6} & \textbf{\textcolor{white}{0}}\textbf{7.0} & \textbf{0.168} & \textbf{0.250}\tabularnewline
\hline 
\multirow{6}{4em}{EdgeBox 500} & \multirow{5}{5em}{16-layer VGGNet} & DenseCap & 48.8 & 36.2 & 25.7 & 16.9 & 10.1 & \textcolor{white}{0}5.4 & \textcolor{white}{0}2.4 & 0.092 & 0.178\tabularnewline
 &  & SCRC  & 52.0 & 39.1 & 27.8 & 18.4 & 11.0 & \textcolor{white}{0}5.8 & \textcolor{white}{0}2.5 & 0.115 & 0.189\tabularnewline
 &  & DBNet w/o bias term  & 52.3 & 43.8 & 36.3 & 29.3 & 22.4 & 15.7 & \textcolor{white}{0}9.4 & 0.124 & 0.246\tabularnewline
 &  & DBNet w/o VOC pretraining  & 54.3 & 45.0 & 36.6 & 28.8 & 21.3 & 14.4 & \textcolor{white}{0}8.2 & 0.144 & 0.245\tabularnewline
 &  & DBNet  & \textbf{54.8} & \textbf{45.9} & \textbf{38.3} & \textbf{30.9} & \textbf{23.7} & \textbf{16.6} & \textbf{\textcolor{white}{0}}\textbf{9.9} & \textbf{0.152} & \textbf{0.258}\tabularnewline
\cline{2-12} 
 & ResNet-101 & DBNet  & \textbf{59.6} & \textbf{50.5} & \textbf{42.3} & \textbf{34.3} & \textbf{26.4} & \textbf{18.6} & \textbf{11.2} & \textbf{0.205} & \textbf{0.284}\tabularnewline
\hline 
\end{tabular}
\par\end{centering}

\unitcut
\unitcut
\vspace*{0.05in}

\endgroup

\caption{Single-image object localization accuracy on the Visual Genome dataset.
Any text phrase annotated on a test image is taken as a query for
that image. ``IoU@'' denotes the overlapping threshold for determining
the recall of ground truth boxes. DC-RPN is the region proposal network
from DenseCap. \label{tab:loc}}
\unitcut
\unitcut
\unitcut
\unitcut
\unitcut
\vspace*{0.02in}
\end{table*}

\paragraph{Full training objective.}

Summing up the training loss for all images together with weight decay
for the whole neural network and the regularization for the
text-specific dynamic classifier (Section~\ref{sec:dis-pathway}), the full training objective is:
\begin{equation}
\min_{\Theta}\frac{1}{M}\sum_{i=1}^{M}L_{i}+\beta_{1}\Gamma_{\text{decay}}+\beta_{2}\Gamma_{\text{dynamic}},\label{eq:full-loss}
\end{equation}
where we set $\beta_{1}=5\times10^{-4}$ and $\beta_{2}=10^{-8}$. 
Model optimization is in Section~\ref{sec:optimization} of the supplementary materials.

\todo{explain the numbers?}

\unitcut\unitcut
\section{Experiments}
\unitcut\unitcut

\paragraph{Dataset.}

We evaluated the proposed DBNet on the Visual Genome dataset~\citep{vg}.
It contains 108,077 images, where $\sim$5M regions are annotated
with text phrases in order to densely cover a wide range of visual entities.

We split the Visual Genome datasets in the same way as in~\citep{densecap}:
77,398 images for training, 5,000 for validation (tuning model parameters),
and 5000 for testing; the remaining 20,679 images were not included (following \cite{densecap}).

The text phrases were annotated from crowd sourcing and 
included a significant portion of misspelled words. 
We corrected misspelled words using the Enchant spell checker~\citep{enchant} from AbiWord. After
that, there were 2,113,688 unique phrases in the training set and
180,363 unique phrases in the testing set. In the test set, about
one third (61,048) of the phrases appeared in the training set, and
the remaining two thirds (119,315) were unseen. About 43 unique phrases
were annotated with ground truth regions per image. All experimental
results are reported on this dataset. 

\paragraph{Models.}

We constructed the fast R-CNN~\citep{fast-rcnn}-style visual pathway of DBNet based on either the 16-layer VGGNet (Model-D in \citep{vggnet}) or ResNet-101~\citep{resnet}. 
In most experiments, we used VGGNet for fair comparison with existing works (which also use VGGNet) and less evaluation time. 
ResNet-101 was used to further improve the accuracy. 

We compared DBNet with two image captioning based localization models: DenseCap~\citep{densecap} and SCRC~\citep{natural-obj}. 
In DBNet, the visual pathway was pretrained for object detection using the faster R-CNN~\citep{faster-rcnn} on the PASCAL VOC 2007+2012 trainval set~\citep{voc}.
The linguistic pathway was randomly initialized. 
Pretrained VGGNet on ImageNet ILSVRC classification dataset~\citep{imagenet} was used to initialize DenseCap, and the model was trained to match the dense captioning accuracy reported by \citet{densecap}. 
We found that the faster R-CNN pretraining did not benefit DenseCap (see Section~\ref{sec:pretraining} of the supplementary materials). The SCRC model was additionally pretrained for image captioning on MS COCO~\citep{coco} in the same way as \citet{natural-obj} did.  

We trained all models using the training set on Visual Genome and evaluated them for both localization on single images and detection on multiple images. 
We also assessed the usefulness of the major components of our DBNet. 

\unitcut
\subsection{Single image localization}
\unitcut

In the localization task, we took all ground truth text phrases annotated on an image as queries to localize the associated objects by maximizing the network response over proposed image regions. 

\paragraph{Evaluation metrics. }

We used the same region proposal method to propose bounding boxes for all models, and we used the non-maximum suppression (NMS) with the IoU threshold $0.3$ to localize a few boxes. 
The performance was evaluated by the recall of ground truth regions of the query phrase (see Section~\ref{sec:def-recall-precision} of the supplementary materials for a discussion on recall and precision for localization tasks).
If one of the proposed bounding boxes with the top-$k$ network responses had a large enough overlap (determined by an IoU threshold) with the ground truth bounding box, we took it as a successful localization.
If multiple ground truth boxes were on the same image, we only required the localized boxes to match one of them. The final recall was
averaged over all test cases, i.e., per image and text phrase.
Median and mean overlap (IoU) between the top-1 localized box and the ground truth were also considered. 

\begin{table}
\vspace*{0.05in}
\begingroup
\renewcommand{\colcut}{\hspace{0em}}
\small{} 
\begin{centering}
\begin{tabular}{l|ccc|c}
\hline 
DenseCap & \multicolumn{3}{c|}{Recall / \% for IoU@} & Median\tabularnewline
\cline{2-4} 
performance & 0.1 & 0.3 & 0.5 & IoU\tabularnewline
\hline 
\hline 
Small test set in \citep{densecap} & 56.0 & 34.5 & 15.3 & 0.137\tabularnewline
Test set in this paper & 50.5 & 24.7 & \textcolor{white}{0}8.1 & 0.103\tabularnewline
\hline 
\end{tabular}
\par\end{centering}
\endgroup

\unitcut\unitcut
\vspace*{0.03in}
\caption{Localization accuracy of DenseCap on the small test set (1000 images
and 100 test queries) used in \citep{densecap} and the full test
set (5000 images and $>$0.2M queries) used in this paper. $1000$
boxes (at most) per image are proposed using the DenseCap RPN. \label{tab:densecap-loc}}
\unitcut\unitcut\unitcut\unitcut
\vspace*{0.05in}
\end{table}

\paragraph{DBNet outperforms captioning models.}

We summarize the top-1 localization performance of different methods in Table~\ref{tab:loc}, where $500$ bounding boxes were proposed for testing.
DBNet outperforms DenseCap and SCRC under all metrics.
In particular, DBNet's recall was more than twice as high as the other two methods for the IoU threshold at $0.5$ (commonly used for object detection~\citep{voc,coco}) and about $4$ times higher for IoU at $0.7$ (for high-precision localization~\citep{kitti,fgs-struct}). 

\citet{densecap} reported DenseCap's localization accuracy on a much smaller test set (1000 images and 100 test queries in total), which is not comparable to our exhaustive test settings (Table~\ref{tab:densecap-loc} for comparison). 
We also note that different region proposal methods (EdgeBox and DenseCap RPN) did not make a big difference on the localization performance.
We used EdgeBox for the rest of our evaluation. 

\begin{figure}[tb]
\vspace*{-0.2in}
\unitcut
\unitcut

\begin{centering}
\hfill{}\subfloat[IoU@0.5]{\begin{centering}
\includegraphics[width=0.45\columnwidth]{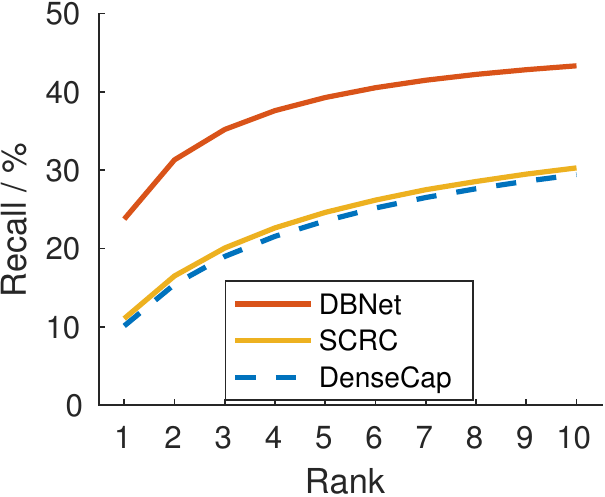}
\par\end{centering}
}\hfill{}\hfill{}\subfloat[IoU@0.7]{\begin{centering}
\includegraphics[width=0.45\columnwidth]{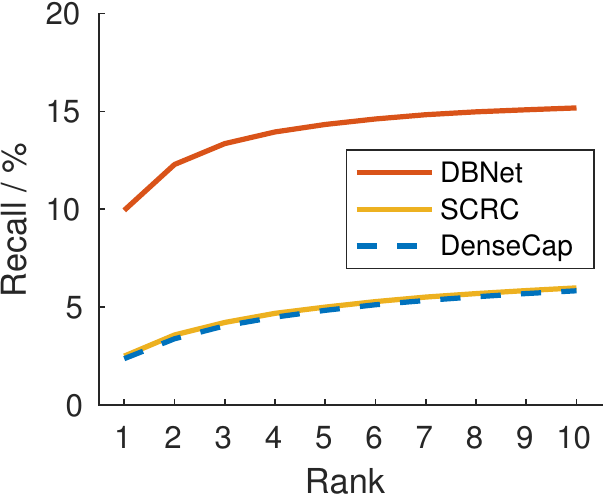}
\par\end{centering}
}\hfill{}
\par\end{centering}
\caption{Top-$k$ localization recall under two overlapping thresholds. VGGNet
and EdgeBox 500 are used in all methods. \label{fig:top-k-loc}}
\unitcut
\unitcut
\unitcut
\unitcut
\end{figure}

Figure~\ref{fig:top-k-loc} shows the top-$k$ recall ($k=1,2,\ldots,10$) in curves. SCRC is slightly better than DenseCap, possibly due to the global context features used in SCRC. 
DBNet outperforms both consistently with a significant margin, thanks to the effectiveness of discriminative training.

\begin{figure}
\begin{centering}
\includegraphics[width=1\columnwidth]{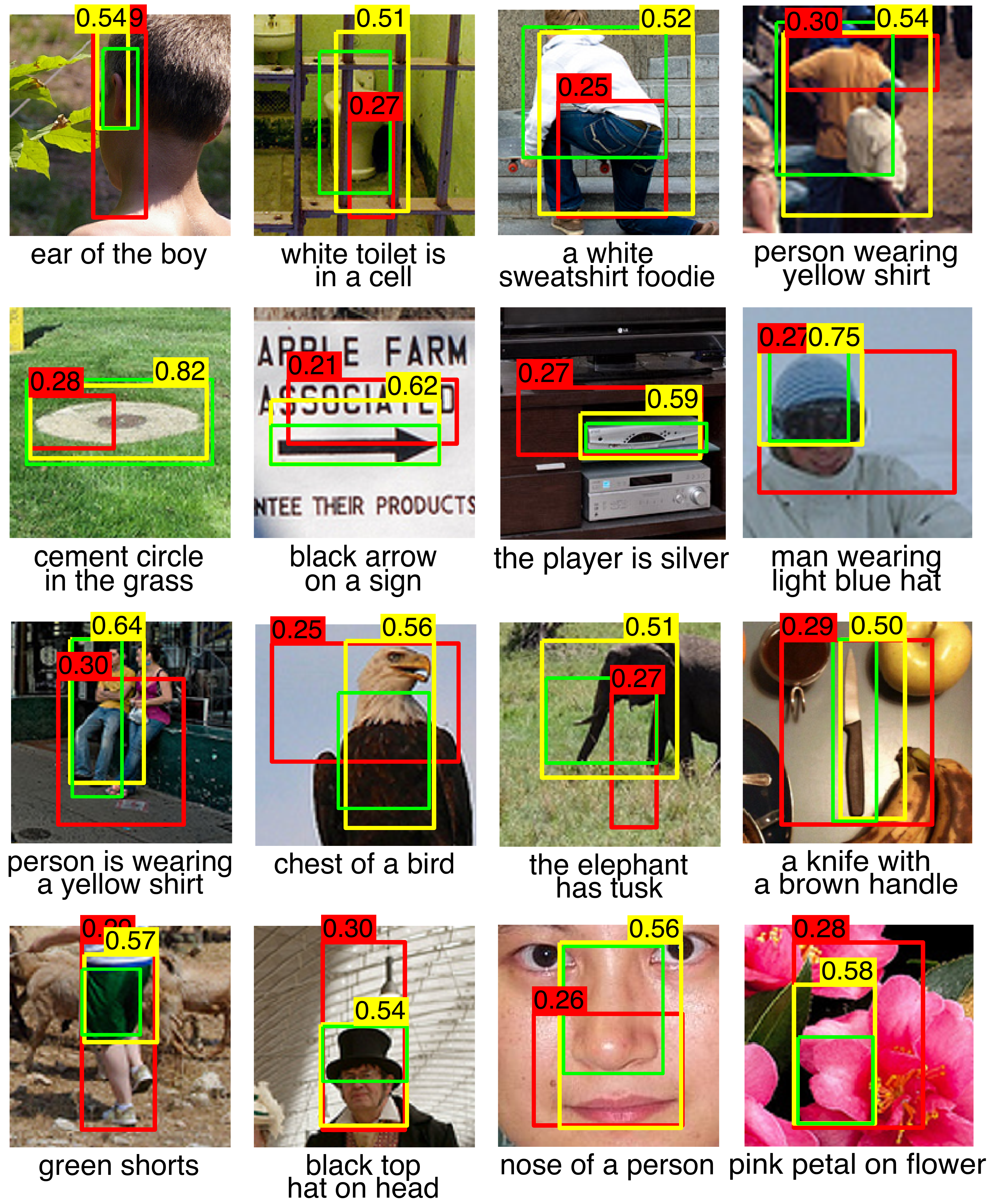}
\par\end{centering}
\unitcut\unitcut
\caption{Qualitative comparison between DBNet and DenseCap on localization task. 
\textcolor{darkgreen}{Green boxes:} ground truth; \textcolor{red}{Red boxes:} DenseCap; \textcolor{darkyellow}{Yellow boxes:} DBNet.
\label{fig:loc-vis} }
\unitcut\unitcut\unitcut\unitcut
\end{figure}

\paragraph{Dynamic bias term improves performance. }

The text-dependent bias term introduced in (\ref{eq:dy-bias}) and
(\ref{eq:dy-classifier}) makes our method for fusing visual and linguistic
representations different from the basic bilinear functions (e.g.,
used in \citep{fine-description}) and more similar to a visual feature
classifier. As in Table~\ref{tab:loc}, this dynamic bias term led
to $>20\%$ relative improvement on median IoU and $\sim5\%$ ($2.5\%\sim0.5\%$
absolute) relative improvement on recall at all IoU thresholds. 

\paragraph{Transferring knowledge benefits localization accuracy.}

Pretraining the visual pathway of DBNet for object detection on PASCAL
VOC showed minor benefit on recall at lower IoU thresholds, but it
brought $10\%$ and $17\%$ relative improvement to the recall for
the IoU threshold at $0.5$ and $0.7$, respectively. See Section~\ref{sec:pretraining}
in the supplementary materials for more results, where we showed that DenseCap did not get benefit from the same technique.

\paragraph{Qualitative results. }

We visually compared the localization results of DBNet and DenseCap in Figure~\ref{fig:loc-vis}. 
In many cases, DBNet localized the queried entities at more reasonable locations. 
More examples are provided in Section~\ref{sec:more-vis-loc} of the supplementary materials.

\paragraph{More quantitative results. }
In the supplementary materials, we studied the performance improvement of the learned models over random guessing and the upper bound performance due to the limitation of region proposal methods (Section~\ref{sec:acc-bound}). 
We also evaluated DBNet using queries in a constrained form (Section~\ref{sec:obj-rel-obj}), where the high query complexity was demonstrated as a significant source of failures for natural language visual localization.

\unitcut
\subsection{Detection on multiple images}
\unitcut

In the detection task, the model needs to verify the existence and quantity of queried visual entities in addition to localizing them, if any. 
Text phrases not associated with any image regions can exist in the query set of an image, and evaluation metrics can be defined by extending those used in traditional object detection. 

\paragraph{Query sets. }

Due to the huge total number of possible query phrases, it is practical to test only a subset of phrases on a test image. 
We developed query sets in three difficulty levels ($0,1,2$). 
For a text phrase, a test image is \emph{positive} if at least one ground truth region exists for the phrase; otherwise, the image is \emph{negative}. 
\begin{itemize}
\item \emph{Level-0:} The query set was the same as in the localization task, so
every text phrase was tested only on its positive images ($\sim$43
phrases per image). 
\item \emph{Level-1:} For each text phrase, we randomly chose the same number
of negative images and the positive images ($\sim$92 phrases per image).
\item \emph{Level-2:} The number of negative images was either 5 times the number of positive images or 20 (whichever was larger) for each test phrase ($\sim$775 phrases per image).
This set included relatively more negative images (compared to positive images) for infrequent phrases.
\end{itemize}
As the level went up, it became more challenging for a detector to
maintain its precision, as more negative test cases are included.
In the level-1 and level-2 sets, text phrases depicting obvious non-object ``stuff'', such as sky, were removed to better fit the detection task. 
Then, 176,794 phrases (59,303 seen and 117,491 unseen) remained.

\begin{table}
\begingroup
\renewcommand{\colcut}{\hspace{-0.1em}}
\small{} 
\begin{centering}
\subfloat[\textbf{Level-0:} Only positive images per text phrase. ]{\begin{centering}
\begin{tabular}{>{\colcut}l<{\colcut}|>{\colcut}c<{\colcut}>{\colcut}c<{\colcut}|>{\colcut}c<{\colcut}>{\colcut}c<{\colcut}|>{\colcut}c<{\colcut}>{\colcut}c<{\colcut}}
\hline 
Average  & \multicolumn{2}{c|}{IoU@0.3} & \multicolumn{2}{c|}{IoU@0.5} & \multicolumn{2}{c}{IoU@0.7}\tabularnewline
\cline{2-7} 
precision / \% & mAP & gAP & mAP & gAP & mAP & gAP\tabularnewline
\hline 
\hline 
DenseCap  & 36.2 & \textcolor{white}{0}1.8 & 15.7 & \textcolor{white}{0}0.5 & \textcolor{white}{0}3.4 & \textcolor{white}{0}0.0\tabularnewline
SCRC & 38.5 & \textcolor{white}{0}2.2 & 16.5 & \textcolor{white}{0}0.5 & \textcolor{white}{0}3.4 & \textcolor{white}{0}0.0\tabularnewline
DBNet & \textbf{48.1} & \textbf{23.1} & \textbf{30.0} & \textbf{10.8} & \textbf{11.6} & \textbf{\textcolor{white}{0}}\textbf{2.1}\tabularnewline
\hline 
DBNet w/ Res & \textbf{51.1} & \textbf{24.2} & \textbf{32.6} & \textbf{11.5} & \textbf{12.9} & \textbf{\textcolor{white}{0}}\textbf{2.2}\tabularnewline
\hline 
\end{tabular}
\par\end{centering}
}
\par\end{centering}
\mysubfloatskip
\begin{centering}
\subfloat[\textbf{Level-1:} The ratio between the positive and negative images is 1:1 per text phrase. ]
{\begin{centering}
\begin{tabular}{>{\colcut}l<{\colcut}|>{\colcut}c<{\colcut}>{\colcut}c<{\colcut}|>{\colcut}c<{\colcut}>{\colcut}c<{\colcut}|>{\colcut}c<{\colcut}>{\colcut}c<{\colcut}}
\hline 
Average  & \multicolumn{2}{c|}{IoU@0.3} & \multicolumn{2}{c|}{IoU@0.5} & \multicolumn{2}{c}{IoU@0.7}\tabularnewline
\cline{2-7} 
precision / \% & mAP & gAP & mAP & gAP & mAP & gAP\tabularnewline
\hline 
\hline 
DenseCap   & 22.9 & \textcolor{white}{0}1.0 & 10.0 & \textcolor{white}{0}0.3 & \textcolor{white}{0}2.1 & \textcolor{white}{0}0.0\tabularnewline
SCRC  & 37.5 & \textcolor{white}{0}1.7 & 16.3 & \textcolor{white}{0}0.4 & \textcolor{white}{0}3.4 & \textcolor{white}{0}0.0\tabularnewline
DBNet & \textbf{45.5} & \textbf{21.0} & \textbf{28.8} & \textbf{\textcolor{white}{0}}\textbf{9.9} & \textbf{11.4} & \textbf{\textcolor{white}{0}}\textbf{2.0}\tabularnewline
\hline 
DBNet w/ Res & \textbf{48.3} & \textbf{22.2} & \textbf{31.2} & \textbf{10.7} & \textbf{12.6} & \textbf{\textcolor{white}{0}}\textbf{2.1}\tabularnewline
\hline 
\end{tabular}
\par\end{centering}
}
\par\end{centering}
\mysubfloatskip
\begin{centering}
\subfloat[\textbf{Level-2:} The ratio between the positive and negative images is at least 1:5 (minimum 20 negative images and 1:5 otherwise) per text phrase. ]
{\begin{centering}
\begin{tabular}{>{\colcut}l<{\colcut}|>{\colcut}c<{\colcut}>{\colcut}c<{\colcut}|>{\colcut}c<{\colcut}>{\colcut}c<{\colcut}|>{\colcut}c<{\colcut}>{\colcut}c<{\colcut}}
\hline 
Average  & \multicolumn{2}{c|}{IoU@0.3} & \multicolumn{2}{c|}{IoU@0.5} & \multicolumn{2}{c}{IoU@0.7}\tabularnewline
\cline{2-7} 
precision / \% & mAP & gAP & mAP & gAP & mAP & gAP\tabularnewline
\hline 
\hline 
DenseCap  & \textcolor{white}{0}4.1 & \textcolor{white}{0}0.1 & \textcolor{white}{0}1.7 & \textcolor{white}{0}0.0 & \textcolor{white}{0}0.3 & \textcolor{white}{0}0.0\tabularnewline
DBNet  & \textbf{26.7} & \textbf{\textcolor{white}{0}}\textbf{8.0} & \textbf{17.7} & \textbf{\textcolor{white}{0}}\textbf{3.9} & \textbf{\textcolor{white}{0}}\textbf{7.6} & \textbf{\textcolor{white}{0}}\textbf{0.9}\tabularnewline
\hline 
DBNet w/ Res & \textbf{29.7} & \textbf{\textcolor{white}{0}}\textbf{9.0} & \textbf{19.8} & \textbf{\textcolor{white}{0}}\textbf{4.3} & \textbf{\textcolor{white}{0}}\textbf{8.5} & \textbf{\textcolor{white}{0}}\textbf{0.9}\tabularnewline
\hline 
\end{tabular}
\par\end{centering}
}
\par\end{centering}
\endgroup
\unitcut
\caption{Detection average precision using query set of three levels of
difficulties. mAP: mean AP over all text phrases. gAP: AP over all
test cases. VGGNet is the default visual CNN for all methods. ``DBNet
w/ Res'' denotes our DBNet with ResNet-101.\label{tab:det-ap}}
\unitcut\unitcut\unitcut
\end{table}

\paragraph{Evaluation metrics. }

We measured the detection performance by average precision (AP). In
particular, we computed AP independently for each query phrase (comparable
to a category in traditional object detection~\citep{voc}) over
its test images, and reported the \emph{mean AP} (\textbf{mAP}) over
all query phrases. Like traditional object detection, the score
threshold for a detected region is category/phrase-specific. 

For more practical natural-language visual detection, where the query text may not be known in advance, we also directly computed
AP over all test cases. We term it \emph{global AP} (\textbf{gAP}),
which implies a universal decision threshold for any query phrase. 
Table~\ref{tab:det-ap} summarizes mAPs and gAPs under different
overlapping thresholds for all models. 

\paragraph{DBNet shows higher per-phrase performance. }

DBNet achieved consistently stronger performance than DenseCap and SCRC
in terms of mAP, indicating that DBNet produced more accurate
detection per given phrase. Even for the challenging IoU threshold
of 0.7, DBNet still showed reasonable performance. The mAP results
suggest the effectiveness of discriminative training. 

\begin{figure*}
\vspace*{-0.1in}
\begin{centering}
\includegraphics[height=0.167\textwidth]{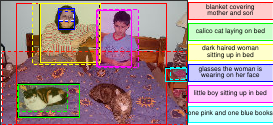}\hspace*{\fill}\includegraphics[height=0.167\textwidth]{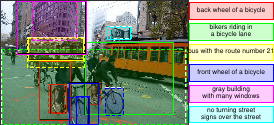}\hspace*{\fill}\includegraphics[height=0.167\textwidth]{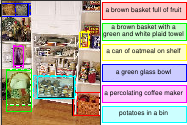}
\par\end{centering}
\medskip{}

\begin{centering}
\includegraphics[height=0.167\textwidth]{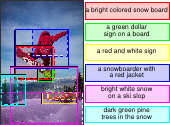}\hspace*{\fill}\includegraphics[height=0.167\textwidth]{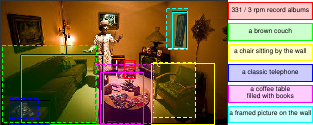}\hspace*{\fill}\includegraphics[height=0.167\textwidth]{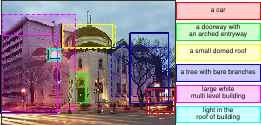}
\par\end{centering}
\caption{Qualitative detection results of DBNet with ResNet-101. 
We show detection results of six different text phrases on each image. 
For each image, the colors of bounding boxes correspond to the colors of text tags on the right. 
The semi-transparent boxes with dashed boundaries are ground truth regions, and the boxes with solid boundaries are detection results.
\label{fig:det-vis}}
\end{figure*}

\begin{table*}
\vspace*{0.07in}
\begingroup
\newcommand{\colcutx}{\hspace{0em}}
\renewcommand{\colcut}{\hspace{0em}}
\small{} 
\begin{centering}
\par\end{centering}
\begin{centering}
\begin{tabular}{>{\colcutx}c<{\colcutx}|>{\colcutx}c<{\colcutx}|>{\colcutx}c<{\colcutx}|>{\colcut}c<{\colcut}>{\colcut}c<{\colcut}>{\colcut}c<{\colcut}|>{\colcut}c<{\colcut}>{\colcut}c<{\colcut}|>{\colcut}c<{\colcut}>{\colcut}c<{\colcut}>{\colcut}c<{\colcut}|>{\colcut}c<{\colcut}>{\colcut}c<{\colcut}>{\colcut}c<{\colcut}}
\hline 
Prune & Phrases & Finetune & \multicolumn{5}{c|}{Localization} & \multicolumn{6}{c}{Detection (Level-1)}\tabularnewline
\cline{4-14} 
ambiguous & from other & visual & \multicolumn{3}{c|}{Recall / \% for IoU@} & Median & Mean & \multicolumn{3}{c|}{mAP / \% for IoU@} & \multicolumn{3}{c}{gAP / \% for IoU@}\tabularnewline
\cline{4-6} \cline{9-14} 
phrases & images & pathway & 0.3 & 0.5 & 0.7 & IoU & IoU & 0.3 & 0.5 & 0.7 & 0.3 & 0.5 & 0.7\tabularnewline
\hline 
\hline 
 \xmark & \xmark & \xmark & 30.6 & 17.5 & \textcolor{white}{0}7.8  & 0.066  & 0.211 & 35.5 & 22.0 & \textcolor{white}{0}8.6 & \textcolor{white}{0}8.3 & \textcolor{white}{0}3.1 & \textcolor{white}{0}0.4\tabularnewline
\cmark & \xmark & \xmark & 34.5 & 21.2 & \textcolor{white}{0}9.0 & 0.113 & 0.237 & 39.0 & 24.6 & \textcolor{white}{0}9.7 & 15.5 & \textcolor{white}{0}7.4 & \textcolor{white}{0}1.6\tabularnewline
\cmark & \cmark & \xmark & 34.7 & 21.1 & \textcolor{white}{0}8.8 & 0.119 & 0.238 & 41.3 & 25.6 & 10.0 & 17.2 & \textcolor{white}{0}7.9 & \textcolor{white}{0}1.6\tabularnewline
\cmark & \cmark & \cmark & \textbf{38.3} & \textbf{23.7} & \textbf{\textcolor{white}{0}}\textbf{9.9} & \textbf{0.152} & \textbf{0.258} & \textbf{45.5} & \textbf{28.8} & \textbf{11.4} & \textbf{21.0} & \textbf{\textcolor{white}{0}}\textbf{9.9} & \textbf{\textcolor{white}{0}}\textbf{2.0}\tabularnewline
\hline 
\end{tabular}
\par\end{centering}
\begin{centering}
\par\end{centering}
\begin{centering}
\par\end{centering}
\endgroup

\caption{Ablation study of DBNet's major components. The visual pathway is based on the 16-layer VGGNet. \label{tab:ablation}}
\vspace*{0.08in}
\end{table*}

\paragraph{DBNet scores are better ``calibrated''. }

Achieving good performance in gAP is challenging as it assumes a phrase-agnostic, universal decision threshold.
For IoU at 0.3 and 0.5, DenseCap
and SCRC showed very low performance in terms of gAP, and DBNet dramatically
($10 \sim20\times$) outperformed them. 
For IoU at 0.7, DenseCap and SCRC were unsuccessful, while DBNet could produce a certain degree of positive results. 
The gAP results suggest that the responses of DBNet are much better calibrated among different text phrases than captioning models, supporting our hypothesis that distributions on a binary decision space are easier to model than those on the huge natural language space. 

\paragraph{Robustness to negative and rare cases. }

The performance of all models dropped as the query set became more difficult. 
SCRC appeared to be more robust than DenseCap for negative test cases (level-1 performance).
DBNet showed superior performance in all difficulty levels. 
Particularly for the level-2 query set, DenseCap's performance dropped significantly compared to the level-1 case, which suggests that it probably failed at handling rare phrases (note that relatively more negative images are included in the level-2 set for rare phrases).  
For IoU at 0.5 and 0.7, DBNet's level-2 performance was
even better than the level-0 performance of DenseCap and SCRC. 
We did not test SCRC on the level-2 query set because of its high time consumption.\footnote{For level-2 query set, DBNet and DenseCap cost $\sim$0.5 min to process one image (775 queries) when using the VGGNet and a Titan X card. 
SCRC takes nearly $10$ minutes with the same setting. 
In addition, DBNet took 2--3 seconds to process one image when using level-0 query set.}

\paragraph{Qualitative results. }

We showed qualitative results of DBNet detection on selected examples in Figure~\ref{fig:det-vis}.
More comprehensive (random and failed) examples are provided in Section~\ref{sec:more-vis-det} of the supplementary materials.
Our DBNet could detect diverse visual entities, including objects
with attributes (e.g., ``a bright colored snow board''), objects
in context (e.g., ``little boy sitting up in bed''), object parts
(e.g., ``front wheel of a bicycle''), and groups of objects (e.g.,``bikers riding in a bicycle lane'').

\unitcut
\subsection{Ablation study on training strategy}
\label{sec:ablation}
\unitcut

We did ablation studies for three components of our DBNet training strategy:
1)~pruning ambiguous phrases ($\mathcal{A}_i(r)$ defined in Eq.~(\ref{eq:amb-txt})),
2)~training with negative phrases from other images ($L_{i}^{\text{rest}}$),
and 3)~finetuning the visual pathway. 

As shown in Table~\ref{tab:ablation}, the performance of the most basic training strategy is better than DenseCap and SCRC, due to the effectiveness of discriminative training. 
Ambiguous phrase pruning led to significant performance gain, by improving the correctness of training labels, where no ``pruning ambiguous phrases'' means setting $\mathcal{A}_i(r)=\emptyset$.  
More quantitative analysis on tuning the text similarity threshold $\tau$ are provided in Section~\ref{sec:text-sim-thresh} of the supplementary materials. 
Inter-image negative phrases did not benefit localization performance, since localization is a single-image task. 
However, this mechanism improved the detection performance by making the model more robust to diverse negative cases. As expected in most vision tasks, finetuning pretrained classification network boosted the performance of our models. 
In addition, upgrading the VGGNet-based visual pathway to ResNet-101 led to another clear gain in DBNet's performance (Table~\ref{tab:loc} and \ref{tab:det-ap}).

\unitcut
\section{Conclusion}
\unitcut

We demonstrated the importance of discriminative learning for natural-language visual localization.
We proposed the discriminative bimodal neural network (DBNet) to allow flexible discriminative training objectives.
We further developed a comprehensive training strategy to extensively and properly leverage negative observations on training data. 
DBNet significantly outperformed the previous state-of-the-art based on caption generation models. 
We also proposed quantitative measurement protocols for natural-language visual detection. 
DBNet showed more robustness against rare queries compared to existing methods and produced detection scores with better calibration over various text queries. 
Our method can be potentially improved by combining its discriminative objective with a generative objective, such as image captioning. 
\iftoggle{arxiv}{\section*{Acknowledgements}
}{
\subsection*{Acknowledgements}
}
This work was funded by Software R{\&}D Center, Samsung Electronics Co., Ltd, as well as ONR N00014-13-1-0762, NSF CAREER IIS-1453651, and Sloan Research Fellowship. We thank NVIDIA for donating K40c and TITAN X GPUs. 
We also thank Kibok Lee, Binghao Deng, Jimei Yang, and Ruben Villegas for helpful discussions.
\iftoggle{arxiv}{}{}
 
\begingroup
\setstretch{1}
\setlength{\bibsep}{2pt}
\iftoggle{arxiv}{}{
\small{}
}

\bibliographystyle{abbrvnat}
\bibliography{ref}

\endgroup

\iftoggle{arxiv}{  }{\filluptopage{11}
\pagestyle{plain}
\setcounter{page}{1}
}

\newcommand{\subf}[2]{  {\tiny\begin{tabular}[t]{@{}c@{}}
  #1\\#2
  \end{tabular}}}      
\onecolumn

\appendix

\begin{center}
\textsf{\textbf{\Large{}Supplementary Materials}}
\par\end{center}

\vspace{0.5em}

\begin{center}
\textbf{\Large{}Discriminative Bimodal Networks for 
\vspace{0.4em} \\
Visual Localization and Detection with Natural Language Queries}
\par\end{center}{\Large \par}

\vspace{1em}

\begin{center}
{\large{}Yuting Zhang, Luyao Yuan, Yijie Guo, Zhiyuan He, I-An Huang, Honglak Lee}
\vspace{0.5em}

{\large{}University of Michigan, Ann Arbor, MI, USA}
\vspace{0.5em}

{\small{} \texttt{\{yutingzh, yuanluya, guoyijie, zhiyuan, huangian, honglak\}@umich.edu}}
\vspace{1.5em}
\par\end{center}

\etocdepthtag.toc{mtappendix}
\etocsettagdepth{mtchapter}{none}
\etocsettagdepth{mtappendix}{subsection}

\begingroup
\hypersetup{allcolors=black}

\renewcommand\cftsecfont{\mdseries}
\renewcommand{\cftsecleader}{\mdseries\cftdotfill{\cftdotsep}}
\renewcommand\cftsecpagefont{\mdseries}
\renewcommand{\cftsecafterpnum}{}
\setlength{\cftsecindent}{1em}
\setlength{\cftsubsecindent}{2.5em}
\setlength{\cftbeforesecskip}{0.3em}
\addtolength{\cftsecnumwidth}{2pt}
\addtolength{\cftsubsecnumwidth}{2pt}

\vspace{-1em}

\tableofcontents
\endgroup

\section{CNN architecture for the linguistic pathway \label{sec:textual-pathway}}

We summarize the CNN architecture used for the linguistic pathway
in Table~\ref{tab:textual-cnn}. 

\begin{table}[H]
\begingroup
\renewcommand{\colcut}{\hspace{0em}}
\small{} 
\begin{centering}
\begin{tabular}{c|c|c|c|c|c|c}
\hline 
Layer ID & Type & Kernel size & Output channels & Pooling size & Output length & Activation\tabularnewline
\hline 
\hline 
0 & input & n/a & 74 & none & 256 & none\tabularnewline
\hline 
1 & convolution & 7 & 256 & 2 & 128 & LReLU (leakage $=0.1$)\tabularnewline
\hline 
2 & convolution & 7 & 256 & none & 128 & LReLU (leakage $=0.1$)\tabularnewline
\hline 
3 & convolution & 3 & 256 & none & 128 & LReLU (leakage $=0.1$)\tabularnewline
\hline 
4 & convolution & 3 & 256 & 2 & 64 & LReLU (leakage $=0.1$)\tabularnewline
\hline 
5 & convolution & 3 & 512 & none & 64 & LReLU (leakage $=0.1$)\tabularnewline
\hline 
6 & convolution & 3 & 512 & 2 & 32 & LReLU (leakage $=0.1$)\tabularnewline
\hline 
7 & inner-product & n/a & 2048 & n/a & n/a & LReLU (leakage $=0.1$)\tabularnewline
\hline 
8 & inner-product & n/a & 2048 & n/a & n/a & LReLU (leakage $=0.1$)\tabularnewline
\hline 
\end{tabular}
\par\end{centering}
\endgroup

\caption{CNN architecture for the linguistic pathway. \label{tab:textual-cnn}}
\end{table}

\section{Formalized comparison with conditional generative models}
\label{sec:compare-captioning-eq}

In contrast to our discriminative framework, which fits $p(l|x,r,t)$, existing methods on natural-language visual localization \citep{natural-obj,densecap,unambiguous-cap} use the conditional caption generation model, where $f(x,t,r;\Theta)$ resembles $p(t|x,r)$. 
In \citep{natural-obj,densecap}, the models are trained by maximizing $p(t|x,r)$. 
In \citep{unambiguous-cap}, the model is trained instead by maximizing $p(r|x,t)$. 
However, it still resembles $p(t|x,r)$, and $p(r|x,t)$ is calculated via Bayes' theorem. 

Since the space of the natural language is intractable, accurately modeling $p(t|x,r)$ is extremely difficult. 
Even considering only the plausible text phrases for $r$ on $x$, the modes of $p(t|x,r)$ are still hard to be properly lifted and balanced due to the lack of enough training samples to cover all valid descriptions. 
The generative modeling for text phrases may fundamentally limit the discriminative power of the existing model. 

In contrast, our model takes both $r$ and $t$ as conditional variables. 
The conditional distribution on $l$ is much easier to model due to the small binary label space, and it also naturally admits discriminative training. 
The power of deep distributed representations can also be leveraged for generalizing textual representations to less frequent phrases. 
 
\section{Model optimization}
\label{sec:optimization}

The training objective is optimized by back-propagation~\citep{backprop} using the mini-batch stochastic gradient descent (SGD) with momentum $0.9$. We use the basic SGD for the visual pathway and Adam~\citep{adam} for the rest of the network. 

We use EdgeBox~\citep{edgebox} to propose 1000 boxes per image (in addition to the boxes annotated with text phrases) during training.
For each image per iteration, we always include the top 50 proposed boxes in the SGD, and randomly sample another 50 out of the remaining 950 box proposals for diversity and efficiency. 

To calculate $L_{i}^{\text{rest}}$ exactly, we need to extract features from all text phrases ($>$2.8M in Visual Genome) in the training set and combine them with almost every image regions in the mini-batch, which is impractical. 
Following the stochastic optimization framework, we randomly sample a few text phrases according to their frequencies of occurrence in the training set. 
This stochastic optimization procedure is consistent with (\ref{eq:loss-rest}). 

In each iteration, we sample $2$ images when using the 16-layer VGGNet and $1$ image when using ResNet-101 on a single Titan X. The representations for each unique phrase and each unique image region is computed once per iteration. 
We partition a DBNet into sub-networks for the visual and textual pathways, and for the discriminative pathway. 
The batch size for those sub-networks are different and determined by inputs, e.g., the numbers of text phrases, bounding boxes, and effective region-text pairs. 
When using 2 images per iteration, the batch size for the discriminative pathway is $\sim$10K, where we feed all effective region-text pairs, as defined in (\ref{eq:effective-set}) , to the discriminative pathway. 
The large batch size is needed for efficient and stable optimization.
Our Caffe \citep{caffe} and MATLAB based implementation supports dynamic and arbitrarily large batch sizes for sub-networks.  
The initial learning rates when using different visual pathways are summarized in Table~\ref{tab:dbnet-lr}. 

\begin{table}[H]
\begingroup
\small{} 
\renewcommand{\arraystretch}{1.2}
\begin{centering}
\begin{tabular}{c|c|c|c}
\hline 
\multicolumn{2}{c|}{Sub-networks \textbackslash{} Models} & 16-layer VGGNet & ResNet-101\tabularnewline
\hline 
\hline 
Visual & Before RoI-pooling & $10^{-3}$ & $10^{-3}$\tabularnewline
\cline{2-4} 
pathway & After RoI-pooling & $10^{-3}$ & $10^{-4}$\tabularnewline
\hline 
\multicolumn{2}{c|}{Remainder} & $10^{-4}$ & $10^{-5}$\tabularnewline
\hline 
\end{tabular}
\par\end{centering}
\endgroup

\caption{Learning rates for DBNet training}
\label{tab:dbnet-lr}

\end{table}

We trained the VGG-based DBNet for approximately 10 days (3--4 days without finetuning the visual network, 4--5 days for the whole network, and 1--2 days with the decreased learning rate).
DenseCap could get converged in $\sim$4 days, but further training did not improve the results. 
Given DBNet's much higher accuracy, the extra training time was worthwhile.   
 
\section{Discussion on recall and precision for localization}
\label{sec:def-recall-precision}

Table~\ref{tab:loc}, \ref{tab:densecap-loc}, and \ref{tab:ablation} report the recall for the localization tasks, where each text phrase is localized with the bounding box of the highest score.
Given an IoU threshold, the localized bounding box is either correct or not. 
As no decision threshold exists in this setting, we can calculate only the \emph{accuracy}, but not a precision-recall curve. 
Following the convention in DenseCap and SCRC, we call this accuracy the ``(rank-1) \emph{recall}'', since it reflects if any ground-truth region can be recalled by the top-scored box.
In Figure~\ref{fig:top-k-loc}, assuming one ground-truth region per image (i.e., ordinary localization settings), we have $\mathrm{precision}=\mathrm{recall}/\mathrm{rank}$. 
Note that rank-1 precision is the same as rank-1 recall.
 
\section{More quantitative results}
\label{sec:more-quant} 

We provide more quantitative analysis in this section, including the impact of pretraining on other datasets, random and upper-bound localization performance, localization with controlled queries, and an ablative study on the text similarity threshold for determining the ambiguous text phrase set. 

\subsection{Pretraining on different datasets}
\label{sec:pretraining}

We trained DBNet and DenseCap using various pretrained visual networks. 
In particular, we used the 16-layer VGGNet in two settings: 1)~pretrained on ImageNet ILSVRC 2012 for image classification (VGGNet-CLS)~\citep{imagenet} and 2)~further pretrained on the PASCAL VOC~\citep{voc} for object detection using faster R-CNN~\citep{faster-rcnn}. 
We compared DBNet and DenseCap trained with these two pretrained networks and tested them with two different region proposal methods (i.e., DenseCap RPN and EdgeBox).
As shown in Table~\ref{tab:pretrain}, VOC pretraining was beneficial for DBNet, but it was not beneficial for DenseCap. 
Thus, we used the ImageNet pretrained VGGNet for DenseCap in the main paper. 

\begin{table}[H]
\begingroup
\renewcommand{\colcut}{\hspace{0em}}
\newcommand{\colcutx}{\hspace{0em}}
\small{} 
\begin{centering}
\begin{tabular}{>{\raggedright}p{4em}|>{\colcut}l<{\colcut}|>{\colcut}c<{\colcut}>{\colcut}c<{\colcut}>{\colcut}c<{\colcut}>{\colcut}c<{\colcut}>{\colcut}c<{\colcut}>{\colcut}c<{\colcut}>{\colcut}c<{\colcut}|>{\colcut}c<{\colcut}>{\colcut}c<{\colcut}}
\hline 
Region & Localization & \multicolumn{7}{c|}{Accuracy / \% for IoU@} & Median  & Mean \tabularnewline
\cline{3-9} 
proposal &  model & 0.1 & 0.2 & 0.3 & 0.4 & 0.5 & 0.6 & 0.7 & IoU & IoU\tabularnewline
\hline 
\hline 
\multirow{4}{4em}{DC-RPN 500} & DenseCap (VGGNet-CLS)  & 52.5 & 38.9 & 27.0 & 17.1 & \textcolor{white}{0}9.5 & \textcolor{white}{0}4.3 & \textcolor{white}{0}1.5 & 0.117 & 0.184\tabularnewline
 & DenseCap (VGGNet-DET)  & 49.4 & 36.9 & 26.0 & 16.7 & \textcolor{white}{0}9.3 & \textcolor{white}{0}4.3 & \textcolor{white}{0}1.5 & 0.096 & 0.176\tabularnewline
 & DBNet (VGGNet-CLS)  & \textbf{57.7} & \textbf{46.9} & 37.0 & 27.9 & 19.5 & 11.7 & \textcolor{white}{0}5.6 & 0.169 & 0.242\tabularnewline
 & DBNet (VGGNet-DET)  & 57.4 & \textbf{46.9} & \textbf{37.8} & \textbf{29.4} & \textbf{21.3} & \textbf{13.6} & \textbf{\textcolor{white}{0}}\textbf{7.0} & \textbf{0.168} & \textbf{0.250}\tabularnewline
\hline 
\multirow{4}{4em}{EdgeBox 500} & DenseCap (VGGNet-CLS) & 48.8 & 36.2 & 25.7 & 16.9 & 10.1 & \textcolor{white}{0}5.4 & \textcolor{white}{0}2.4 & 0.092 & 0.178\tabularnewline
 & DenseCap (VGGNet-DET) & 46.6 & 34.8 & 24.9 & 16.6 & 10.0 & \textcolor{white}{0}5.2 & \textcolor{white}{0}2.2 & 0.076 & 0.171\tabularnewline
 & DBNet (VGGNet-CLS)  & 54.3 & 45.0 & 36.6 & 28.8 & 21.3 & 14.4 & \textcolor{white}{0}8.2 & 0.144 & 0.245\tabularnewline
 & DBNet (VGGNet-DET) & \textbf{54.8} & \textbf{45.9} & \textbf{38.3} & \textbf{30.9} & \textbf{23.7} & \textbf{16.6} & \textbf{\textcolor{white}{0}}\textbf{9.9} & \textbf{0.152} & \textbf{0.258}\tabularnewline
\hline 
\end{tabular}
\par\end{centering}
\endgroup

\caption{Localization performance for DBNet and DenseCap with different pretrained models on Visual Genome. VGGNet-CLS: the 16-layer VGGNet pretrained on ImageNet ILSVRC 2012 dataset. VGGNet-DET: the 16-layer VGGNet further pretrained on PASCAL VOC07+12 trainval set.}
\label{tab:pretrain}

\end{table}

\subsection{Random and oracle localization performance}
\label{sec:acc-bound}

Given proposed image regions, we performed localization for text phrases with random guessing and the oracle detector. 
For random guessing, we randomly chose a proposed region and took it as the localization results. 
For more accurate evaluation, we averaged the results over all possible cases (i.e., enumerating over all proposed boxes).
For the oracle detector, it always picked up the proposed region that had the largest overlap with a ground truth region, providing the performance upper bound due to the limitation of the region proposal method, as in \citep{fgs-struct}. 

As shown in Table~\ref{tab:loc-bound}, the trained models (DBNet, SCRC, DenseCap) significantly outperformed  random guessing, which suggests that promising models can be developed using deep neural networks. 
However, the the performance of DBNet had a large gap with the oracle detector, which indicates that more advanced methods need to be developed in the further to better address the natural language visual localization problem. 

\begin{table}[H]
\begingroup
\renewcommand{\colcut}{\hspace{0em}}
\small{} 
\begin{centering}
\par\end{centering}
\begin{centering}
\begin{tabular}{>{\colcut}l<{\colcut}|>{\colcut}r<{\colcut}>{\colcut}r<{\colcut}>{\colcut}r<{\colcut}>{\colcut}r<{\colcut}>{\colcut}r<{\colcut}>{\colcut}r<{\colcut}>{\colcut}r<{\colcut}|>{\colcut}c<{\colcut}>{\colcut}c<{\colcut}}
\hline 
\multirow{2}{2em}{Model} & \multicolumn{7}{c|}{Recall / \% for IoU@} & Median  & Mean \tabularnewline
\cline{2-8} 
 & 0.1 & 0.2 & 0.3 & 0.4 & 0.5 & 0.6 & 0.7 & IoU & IoU\tabularnewline
\hline 
\hline 
Random & 19.0 & 10.0 & 5.2 & 2.6 & 1.2 & \textcolor{white}{0}0.5 & \textcolor{white}{0}0.2 & 0.041 & 0.056\tabularnewline
\hline
DenseCap & 48.8 & 36.2 & 25.7 & 16.9 & 10.1 & \textcolor{white}{0}5.4 & \textcolor{white}{0}2.4 & 0.092 & 0.178\tabularnewline
SCRC  & 52.0 & 39.1 & 27.8 & 18.4 & 11.0 & \textcolor{white}{0}5.8 & \textcolor{white}{0}2.5 & 0.115 & 0.189\tabularnewline
 DBNet  & \textbf{54.8} & \textbf{45.9} & \textbf{38.3} & \textbf{30.9} & \textbf{23.7} & \textbf{16.6} & \textbf{\textcolor{white}{0}}\textbf{9.9} & \textbf{0.152} & \textbf{0.258}\tabularnewline
\hline
 Oracle & 94.0 & 87.3 & 80.4 & 73.1 & 65.1 & \textcolor{white}{0}55.8 & \textcolor{white}{0}42.4 & 0.650 & 0.572\tabularnewline
\hline 
\end{tabular}
\par\end{centering}

\endgroup

\caption{Single-image object localization accuracy on the Visual Genome dataset for random guess, oracle detector, and trained models. 
EdgeBox is used to propose 500 regions per image. 
\emph{Random}: a proposed region is randomly chosen as the localization for a text phrase and the performance is averaged over all possibilities;
\emph{Oracle}: the proposed region that has the largest overlap with the ground box(es) is taken as the localization for a text phrase. \label{tab:loc-bound}}

\end{table}

\subsection{Localization using constrained queries}
\label{sec:obj-rel-obj}

Pairwise relationships describe a particular type of visual entities, i.e., two objects interacting with each other in a certain way. 
As the basic building block of more complicated parsing structures, the pairwise relationship is worth evaluating as a special case. 
The Visual Genome dataset has pairwise object relationship annotations, independent from the text phrase annotations. 
To fit ``object-relationship-object'' (Obj-Rel-Obj) triplets into our model, we represented a triplet in a SVO (subject-verb-object) text phrase, and took the bounding box enclosing the two objects as the ground truth region for the SVO phrase.  
During the training time, we used both the original text phrase annotations and the SVO phrases derived from the relationship annotations to keep sufficient diversity of the text descriptions. 
During the testing time, we used only the SVO phrases to focus on the localization of pairwise relationships.
The training and testing sets of images were the same as in the other experiments. 

As reported in Table~\ref{tab:loc}, the localization recall for the IoU threshold at $0.5$ was close to $50\%$.  
The groups of two objects were easier to localize than general visual entities, since they were more clearly defined and generally context-free. 
In particular, DBNet's performance (recall and median/mean IoU) for Obj-Rel-Obj was approximately twice as high as that for general text phrases. 
The above experimental results demonstrate the effectiveness of DBNet for localizing object relationships. 
The results also demonstrate the complexity of the text quires (e.g., using all human-annotated phrases v.s. obj-rel-obj pairs) as a significant source of failures. 

\begin{table}[H]
\begingroup
\renewcommand{\colcut}{\hspace{0em}}
\small{} 
\begin{centering}
\par\end{centering}
\begin{centering}
\begin{tabular}{>{\raggedright}p{4em}|>{\raggedright}p{5em}|>{\colcut}l<{\colcut}|>{\colcut}c<{\colcut}>{\colcut}c<{\colcut}>{\colcut}c<{\colcut}>{\colcut}c<{\colcut}>{\colcut}c<{\colcut}>{\colcut}c<{\colcut}>{\colcut}c<{\colcut}|>{\colcut}c<{\colcut}>{\colcut}c<{\colcut}}
\hline 
Region & Visual & Localization & \multicolumn{7}{c|}{Recall / \% for IoU@} & Median  & Mean \tabularnewline
\cline{4-10} 
proposal & network &  model & 0.1 & 0.2 & 0.3 & 0.4 & 0.5 & 0.6 & 0.7 & IoU & IoU\tabularnewline
\hline 
\hline 
\multirow{2}{4em}{EdgeBox 500} & \multirow{2}{5em}{16-layer VGGNet} & DBNet (all phrases)  & \textbf{54.8} & \textbf{45.9} & \textbf{38.3} & \textbf{30.9} & \textbf{23.7} & \textbf{16.6} & \textbf{\textcolor{white}{0}}\textbf{9.9} & \textbf{0.152} & \textbf{0.258}\tabularnewline
\cline{3-12}
 & & DBNet (Obj-Rel-Obj) & \textbf{81.8} & \textbf{75.1} & \textbf{67.3} & \textbf{57.8} & \textbf{46.8} & \textbf{35.4} & \textbf{23.1} & \textbf{0.471} & \textbf{0.448} \\
\hline 
\end{tabular}
\par\end{centering}

\endgroup

\caption{Single-image object localization accuracy on the Visual Genome dataset.
Any text phrase annotated on a test image is taken as a query for that image. 
``IoU@'' denotes the overlapping threshold for determining the recall of ground truth boxes. DC-RPN is the region proposal network from DenseCap. \label{tab:obj-rel-obj}}

\end{table}

\subsection{Ablative study on the text similarity threshold}
\label{sec:text-sim-thresh}

As discussed in Section~\ref{sec:ablation}, removing ambiguous training samples are important.
The ambiguous sample pruning depends on 1)~overlaps between proposed regions and ground truth regions, and 2)~text similarity. 
While the image region overlaps have been commonly considered in traditional object detection, the text similarity is specific to natural language visual localization and detection. 

In Table~\ref{tab:ambiguous-text}, we reported the localization performance of DBNet under different values of the text similarity threshold $\tau$ (defined in Eq.~\eqref{eq:amb-txt}), where we considered a controlled setting with neither text phrases from other images nor the visual pathway finetuning. 
DBNet achieved the best performance with the default parameter $\tau=0.3$.
Suboptimal $\tau$ caused approximately $0.5\%$--$1\%$ decrease in localization recall and $0.01$ decrease in median/mean IoU. 

\begin{table}[H]
\vspace{0.5em}
\small

\begin{center}
\begin{tabular}{c|c|c|c|c|c|cc}
\hline 
Phrases from & Finetuning & \multirow{2}{*}{$\tau$} & \multicolumn{3}{c|}{Recall / \% for IoU@} & Median & Mean\tabularnewline
\cline{4-6} 
other images & visual pathway &  & 0.3 & 0.5 & 0.7 & IoU & IoU\tabularnewline
\hline 
\hline 
No & No & 0.1 & 33.6 & 20.6 & \textcolor{white}{0}8.6 & 0.101 & 0.231\tabularnewline
No & No & 0.2 & 33.0 & 20.2 & \textcolor{white}{0}8.5 & 0.094 & 0.227\tabularnewline
No & No & 0.3 & \textbf{34.5} & \textbf{21.2} & \textcolor{white}{0}\textbf{9.0} & \textbf{0.113} & \textbf{0.237}\tabularnewline
No & No & 0.4 & 33.0 & 20.2 & \textcolor{white}{0}8.4 & 0.093 & 0.227\tabularnewline
No & No & 0.5 & 32.8 & 20.2 & \textcolor{white}{0}8.4 & 0.091 & 0.226\tabularnewline
\hline 
\end{tabular}
\par\end{center}

\caption{Ablative study on text similarity threshold $\tau$ in Eq.~\eqref{eq:amb-txt}. }
\label{tab:ambiguous-text}

\end{table}

Since the above controlled setting excluded text phrases from the rest of the training set, the localization performance was not too sensitive to the value of $\tau$ due to the limited number of phrases. 
When the text phrases from the whole training set are included in the training loss on a single image, the choice of $\tau$ can have a more obvious impact. 
For example, setting $\tau=0$ can disable the inclusion of text phrases from other images in any case. 

\vfill{}

\iftoggle{arxiv}{\clearpage}{}

\section{More qualitative comparison for localization}
\label{sec:more-vis-loc}

More quantitative localization results were shown in this section.
We compared DBNet with DenseCap (Figure~\ref{fig:loc-vis-densecap} in Section~\ref{sec:more-vis-loc-densecap}) and SCRC (Figure~\ref{fig:loc-vis-scrc} in Section~\ref{sec:more-vis-loc-scrc}), respectively. 
For each test example, we cropped the image to make the figure focus on the localized region. 
We used a green box for the ground truth region, a red box for DenseCap/SCRC, and a yellow box for our DBNet.  

In the examples that we showed, at least one of the two methods (DBNet and DenseCap/SCRC) can localize the text query to an image region that has $\text{IoU}>0.2$ overlap with the ground truth region.  
Besides this constraint, all examples were chosen randomly. 
While DenseCap and SCRC outperformed DBNet in a few cases, DBNet significantly outperformed those two methods most of the time. 

\vfill{}

\newcommand{\locviswidth}{0.12\textwidth}

\newcommand{\locviscaption}[1]{Qualitative comparison between DBNet and #1 on localization task. 
Examples are randomly sampled. 
\textcolor{darkgreen}{Green boxes:} ground truth; \textcolor{red}{Red boxes:} #1; \textcolor{darkyellow}{Yellow boxes:} DBNet. The numbers are IoU with ground truth boxes. }

\iftoggle{arxiv}{
\vfill

{
\begin{center}
\LARGE See results on the next page.  
\end{center}
}

\vfill
\vfill
}{}

\clearpage
\subsection{More qualitative comparison with DenseCap}
\label{sec:more-vis-loc-densecap}

\begin{figure}[H]
\centering
\begin{tabular}{cccccc}
\centering
\subf{\includegraphics[width=\locviswidth]{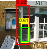}}
     { \small a small chalk\\ \small board in the window}
&
\subf{\includegraphics[width=\locviswidth]{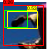}}
     {\small a foot hang \\ \small over the side}
&
\subf{\includegraphics[width=\locviswidth]{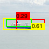}}
     {\small blue and white ship \\ \small on the water}
&
\subf{\includegraphics[width=\locviswidth]{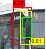}}
     {\small side mirror \\ \small of the bus}
&
\subf{\includegraphics[width=\locviswidth]{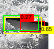}}
     {\small part of\\ \small ocean surfboard}
&
\subf{\includegraphics[width=\locviswidth]{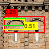}}
     {\small arch over \\ \small window column}

\\

\subf{\includegraphics[width=\locviswidth]{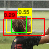}}
     {\small the short hair \\ \small of the player}
&
\subf{\includegraphics[width=\locviswidth]{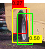}}
     {\small the back wheel \\ \small of the bus}
&
\subf{\includegraphics[width=\locviswidth]{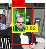}}
     {\small a man wearing \\ \small sunglasses}
&
\subf{\includegraphics[width=\locviswidth]{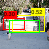}}
     {\small leash on dog pulling\\ \small person on skateboard}
&
\subf{\includegraphics[width=\locviswidth]{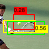}}
     {\small racket in tennis \\ \small player's hand}
&
\subf{\includegraphics[width=\locviswidth]{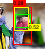}}
     {\small child is wearing \\ \small pink gloves}
\\

\subf{\includegraphics[width=\locviswidth]{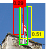}}
     {\small flag on top of \\ \small building}
&
\subf{\includegraphics[width=\locviswidth]{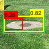}}
     {\small cement circle\\ \small in the glass}
&
\subf{\includegraphics[width=\locviswidth]{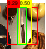}}
     {\small a knife with \\ \small a brown handle}
&
\subf{\includegraphics[width=\locviswidth]{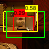}}
     {\small silver stereo\\ \small on table}
&
\subf{\includegraphics[width=\locviswidth]{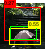}}
     {\small umbrella covering\\ \small the people}
&
\subf{\includegraphics[width=\locviswidth]{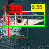}}
     {\small the man \\ \small is surfing}
    
\\
\subf{\includegraphics[width=\locviswidth]{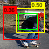}}
     {\small a black \\ \small tee shirt}
&
\subf{\includegraphics[width=\locviswidth]{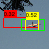}}
     {\small a airplane is \\ \small in the sky}
&
\subf{\includegraphics[width=\locviswidth]{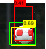}}
     {\small a red and \\ \small white ball}
&
\subf{\includegraphics[width=\locviswidth]{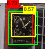}}
     {\small the clock says\\ \small 11/08}
&
\subf{\includegraphics[width=\locviswidth]{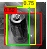}}
     {\small drink can sitting\\ \small on the sink}
&
\subf{\includegraphics[width=\locviswidth]{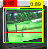}}
     {\small video game on \\ \small the tv screen}

\\

\subf{\includegraphics[width=\locviswidth]{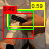}}
     {\small cell phone \\ \small in hand}
&
\subf{\includegraphics[width=\locviswidth]{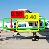}}
     {\small side windows \\ \small of a plane}
&
\subf{\includegraphics[width=\locviswidth]{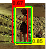}}
     {\small a woman waiting at \\ \small the train station}
&
\subf{\includegraphics[width=\locviswidth]{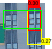}}
     {\small the building\\ \small has windows}
&
\subf{\includegraphics[width=\locviswidth]{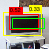}}
     {\small black screen\\ \small on television}
&
\subf{\includegraphics[width=\locviswidth]{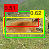}}
     {\small light brown cow \\ \small on ground}

\\

\subf{\includegraphics[width=\locviswidth]{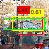}}
     {\small a brown roof \\ \small of a building}
&
\subf{\includegraphics[width=\locviswidth]{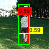}}
     {\small girl about to \\ \small throw frisbee}
&
\subf{\includegraphics[width=\locviswidth]{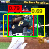}}
     {\small dark shirt with \\ \small yellow writing}
&
\subf{\includegraphics[width=\locviswidth]{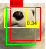}}
     {\small the bear has\\ \small a black nose}
&
\subf{\includegraphics[width=\locviswidth]{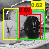}}
     {\small the clock\\ \small is black}
&
\subf{\includegraphics[width=\locviswidth]{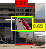}}
     {\small a red stripe \\ \small on the plane}

\\

\subf{\includegraphics[width=\locviswidth]{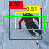}}
     {\small the red jacket \\ \small of the skier}
&
\subf{\includegraphics[width=\locviswidth]{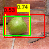}}
     {\small the apple \\ \small is green}
&
\subf{\includegraphics[width=\locviswidth]{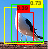}}
     {\small the bird stands \\ \small in the sun}
&
\subf{\includegraphics[width=\locviswidth]{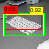}}
     {\small white keys\\ \small of a keyboard}
&
\subf{\includegraphics[width=\locviswidth]{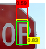}}
     {\small a letter p\\ \small written in white}
&
\subf{\includegraphics[width=\locviswidth]{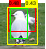}}
     {\small a fat seagull \\ \small standing}

\\

\end{tabular}
\caption{\locviscaption{DenseCap}
\label{fig:loc-vis-densecap} }
\end{figure}

\clearpage
\subsection{More qualitative comparison with SCRC}
\label{sec:more-vis-loc-scrc}

\begin{figure}[H]
\centering
\begin{tabular}{cccccc}
\centering
\subf{\includegraphics[width=\locviswidth]{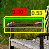}}
     { \small left armrest \\ \small of the bench}
&
\subf{\includegraphics[width=\locviswidth]{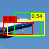}}
     {\small the nose of \\ \small the airplane}
&
\subf{\includegraphics[width=\locviswidth]{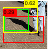}}
     {\small shadow \\ \small from the man}
&
\subf{\includegraphics[width=\locviswidth]{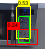}}
     {\small the microwave\\ \small has buttons}
&
\subf{\includegraphics[width=\locviswidth]{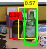}}
     {\small this is \\ \small a bottle}
&
\subf{\includegraphics[width=\locviswidth]{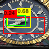}}
     {\small gray hour hand\\ \small on clock}
\\

\subf{\includegraphics[width=\locviswidth]{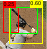}}
     {\small a black\\ \small baseball bat}
&
\subf{\includegraphics[width=\locviswidth]{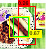}}
     {\small the weiner is \\ \small in the bun}
&
\subf{\includegraphics[width=\locviswidth]{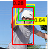}}
     {\small a guy is wearing\\ \small a white helmet}
&
\subf{\includegraphics[width=\locviswidth]{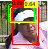}}
     {\small the visor\\ \small is white}
&
\subf{\includegraphics[width=\locviswidth]{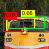}}
     {\small a window\\ \small on the train}
&
\subf{\includegraphics[width=\locviswidth]{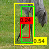}}
     {\small a bird's \\ \small tiny leg}
    
\\
\subf{\includegraphics[width=\locviswidth]{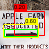}}
     {\small arrow\\ \small pointing right}
&
\subf{\includegraphics[width=\locviswidth]{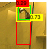}}
     {\small the shower\\ \small nozzle}
&
\subf{\includegraphics[width=\locviswidth]{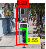}}
     {\small blue jeans \\ \small on young woman}
&
\subf{\includegraphics[width=\locviswidth]{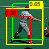}}
     {\small the uniform\\ \small is grey}
&
\subf{\includegraphics[width=\locviswidth]{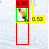}}
     {\small man in yellow\\ \small snowboarding}
&
\subf{\includegraphics[width=\locviswidth]{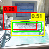}}
     {\small a white\\ \small computer keyboard}
\\

\subf{\includegraphics[width=\locviswidth]{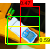}}
     {\small a remote control \\ \small on coffee table}
&
\subf{\includegraphics[width=\locviswidth]{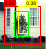}}
     {\small a tree near \\ \small a house}
&
\subf{\includegraphics[width=\locviswidth]{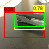}}
     {\small gray stapler}
&
\subf{\includegraphics[width=\locviswidth]{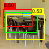}}
     {\small this is a\\ \small dining table}
&
\subf{\includegraphics[width=\locviswidth]{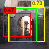}}
     {\small the arch of\\ \small a building}
&
\subf{\includegraphics[width=\locviswidth]{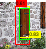}}
     {\small the door on \\ \small the stone cottage}
\\

\subf{\includegraphics[width=\locviswidth]{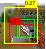}}
     {\small the silver \\ \small long train}
&
\subf{\includegraphics[width=\locviswidth]{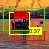}}
     {\small boat in the \\ \small middle of water}
&
\subf{\includegraphics[width=\locviswidth]{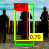}}
     {\small man wearing \\ \small wet suit}
&
\subf{\includegraphics[width=\locviswidth]{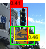}}
     {\small yellow directional sign\\ \small on street}
&
\subf{\includegraphics[width=\locviswidth]{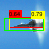}}
     {\small the jet is\\ \small made of steel}
&
\subf{\includegraphics[width=\locviswidth]{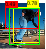}}
     {\small bearded man with \\ \small a white hat}
\\

\subf{\includegraphics[width=\locviswidth]{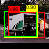}}
     {\small partially loaded\\ \small moving van}
&
\subf{\includegraphics[width=\locviswidth]{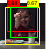}}
     {\small young boy \\ \small pointing at camera}
&
\subf{\includegraphics[width=\locviswidth]{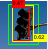}}
     {\small black traffic \\ \small light}
&
\subf{\includegraphics[width=\locviswidth]{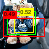}}
     {\small blue and white\\ \small stripe outfit}
&
\subf{\includegraphics[width=\locviswidth]{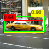}}
     {\small yellow taxi cab\\ \small on the street}
&
\subf{\includegraphics[width=\locviswidth]{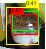}}
     {\small granola in \\ \small yogurt cup}
\\
\subf{\includegraphics[width=\locviswidth]{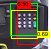}}
     {\small keyboard of \\ \small street meter}
&
\subf{\includegraphics[width=\locviswidth]{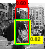}}
     {\small young man carrying\\ \small backpack}
&
\subf{\includegraphics[width=\locviswidth]{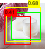}}
     {\small two candle\\ \small holders}
&
\subf{\includegraphics[width=\locviswidth]{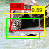}}
     {\small trunck of\\ \small elephant}
&
\subf{\includegraphics[width=\locviswidth]{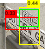}}
     {\small directional street sign\\ \small 1600 block}
&
\subf{\includegraphics[width=\locviswidth]{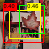}}
     {\small paper note shaped \\ \small like autumn leaf}
\\

\end{tabular}
\caption{\locviscaption{SCRC}
\label{fig:loc-vis-scrc} }
\end{figure}

\section{Qualitative Comparison for Detection}
\label{sec:more-vis-det}

\newcommand{\detvisheight}{0.22\textwidth}
\newcommand{\detvistextheight}{\detvisheight}
\newcommand{\detvistextwidth}{0.15\textwidth}
\newcommand{\detvistexthalfheight}{0.1\textwidth}

\newcommand{\detviscaption}[2]{Qualitative detection results of DBNet, DenseCap, and SCRC #1. 
Detection results of #2 different text phrases are shown for each image. 
The colors of the bounding boxes correspond to the colors of text phrases on the left. 
The semi-transparent boxes with dashed boundaries are ground truth regions, and the boxes with solid boundaries are detection results of three models.}

\newcommand{\detviscase}{}

\newcommand{\failviscaption}{
Random failure examples. 
\textcolor{darkgreen}{Green boxes with solid boundary:}  successful detection ($IoU \geq 0.5$); \textcolor{darkgreen}{Green boxes with dashed boundary:} ground truth with matched detection; \textcolor{red}{Red boxes:} false alarm; \textcolor{darkyellow}{Yellow boxes with dashed boundary:} missed ground truth (without matched detection);  \textcolor{blue}{Blue boxes:} inaccurately localized detection ($0 < IoU < 0.5$). }

In this section, we showed more qualitative results for visual entity detection with various phrases. 
As opposed to the localization task, a decision threshold was needed to decide if the visual entity of interest exists or not. 
We determined this threshold either using prior knowledge on the ground truth regions (Section~\ref{sec:more-vis-det-known-num}) or based on the precision of the detector (Section~\ref{sec:more-vis-det-phrase-dependent} and Section~\ref{sec:more-vis-det-failures}).

In Section~\ref{sec:more-vis-det-known-num}, we showed the same number of detected regions as the ground truth regions for all methods. 
We visualized randomly chosen testing images and phrases under the constraint that at least one of DBNet, DenseCap, or SCRC could get sufficiently accurate detection results (IoU with a ground truh is greater than $0.4$).

In Section~\ref{sec:more-vis-det-phrase-dependent}, we found a decision threshold for each text phrase to make the detection precision (for the IoU threshold at $0.5$) equal to $0.5$. 
If not applicable, we excluded that phrase from visualization.  
We randomly chose testing images and phrases to visualize. 

In Section~\ref{sec:more-vis-det-failures}, we used the same decision threshold as in Section~\ref{sec:more-vis-det-phrase-dependent}. 
However, we focused on visualizing failed detection cases. 
In particular, we randomly chose testing images and phrases under the constraint that at least one of DBNet, DenseCap, and SCRC gave significantly wrong detection results (IoU with any ground truth is less than $0.2$). 
The failure types were also displayed in the figures.

\vfill

{
\begin{center}
\LARGE See results on the next page.  
\end{center}
}

\vfill
\vfill

\clearpage

\begingroup

\iftoggle{arxiv_flatten}
{
    \fancyfoot[C]{$\begin{array}{c}\end{array}$ \\ \footnotesize \vspace*{1em} \lowreswarn}
}
{
    \pagestyle{empty}
}

\begin{landscape}

\subsection{Random detection results with known number of ground truths}
\label{sec:more-vis-det-known-num}

In Figure~\ref{fig:det-vis-known-num}, the number of ground truth entities on the image was supposed to be known in advance. 
All three methods (DBNet, DenseCap, and SCRC) could perform similarly for detecting queried visual entities under a loose standard for localization accuracy (e.g., counting a detected box as a true positive even if it overlaps slightly with the ground truth box). 
The localization accuracy of DBNet was usually more accurate.

\renewcommand{\detviscase}{when the number of ground truth is known}

\begin{centering}
\vspace*{\fill}
\begin{figure}[H]
\begin{tabular}{cccc}
\centering

\large Text phrases
&
\large DBNet
&
\large DenseCap
&
\large SCRC

\\
\hline
\subf{\includegraphics[height=\detvisheight]{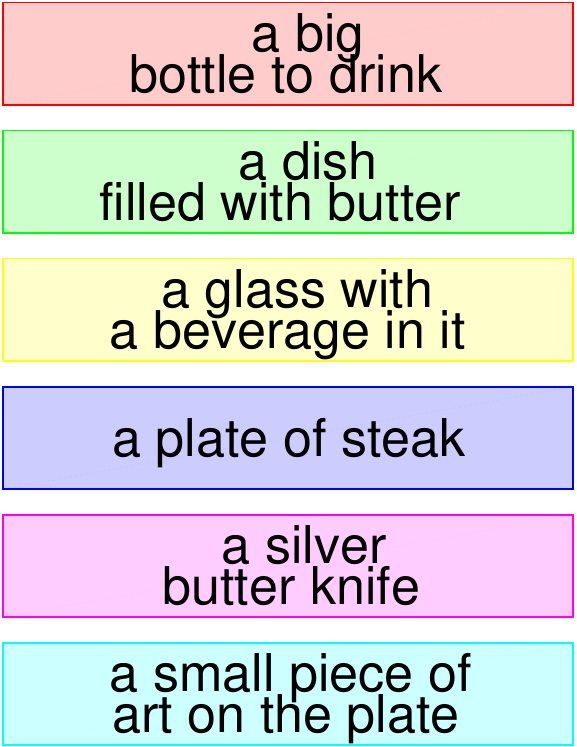}}
     {}
&
\subf{\includegraphics[height=\detvisheight]{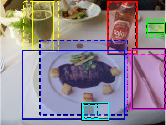}}
     {}    
&
\subf{\includegraphics[height=\detvisheight]{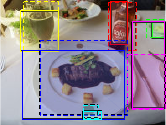}}
     {} 
&
\subf{\includegraphics[height=\detvisheight]{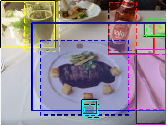}}
     {}
\\
\hline
\subf{\includegraphics[height=\detvisheight]{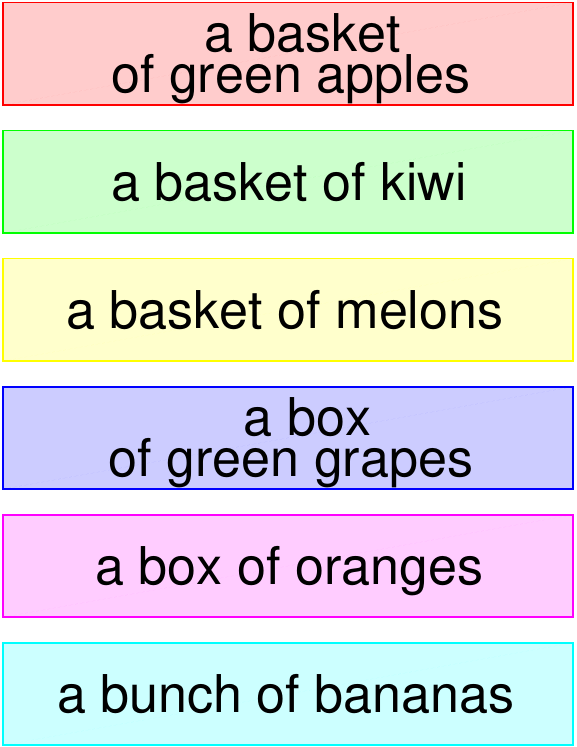}}
     {}
&
\subf{\includegraphics[height=\detvisheight]{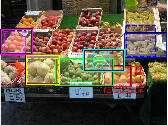}}
     {}    
&
\subf{\includegraphics[height=\detvisheight]{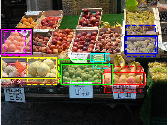}}
     {} 
&
\subf{\includegraphics[height=\detvisheight]{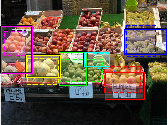}}
     {}
\\
\hline
\subf{\includegraphics[height=\detvisheight]{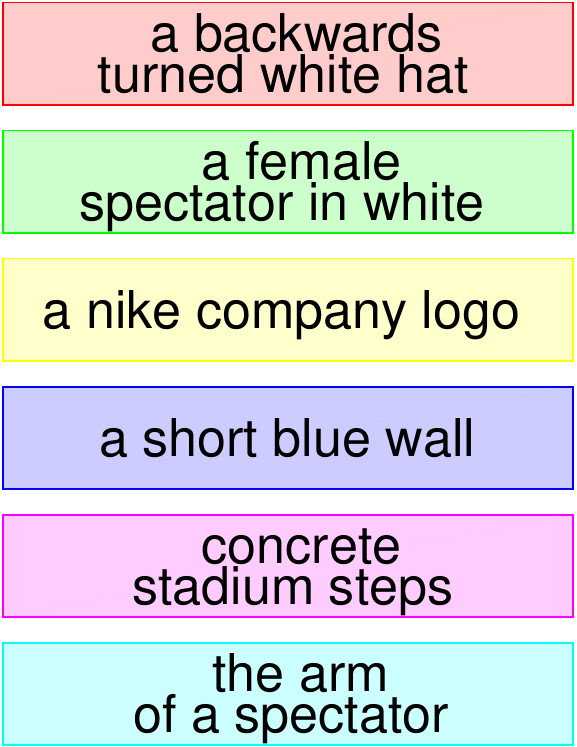}}
     {}
&
\subf{\includegraphics[height=\detvisheight]{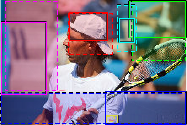}}
     {}    
&
\subf{\includegraphics[height=\detvisheight]{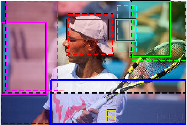}}
     {} 
&
\subf{\includegraphics[height=\detvisheight]{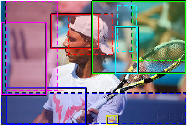}}
     {}
\end{tabular}
\caption{\detviscaption{\detviscase}{six}
\label{fig:det-vis-known-num} }
\end{figure}
\vspace*{\fill}
\end{centering}
\clearpage

\vspace*{\fill}
\begin{figure}[H]
\centering
\begin{tabular}{cccc}
\centering

\large Text phrases
&
\large DBNet
&
\large DenseCap
&
\large SCRC

\\
\hline
\subf{\includegraphics[height=\detvisheight]{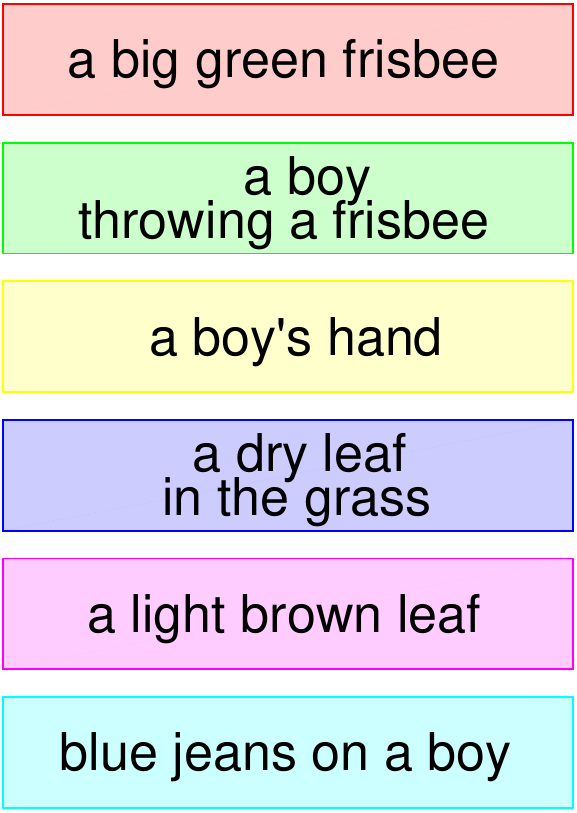}}
     {}
&
\subf{\includegraphics[height=\detvisheight]{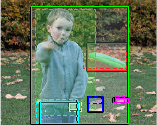}}
     {}    
&
\subf{\includegraphics[height=\detvisheight]{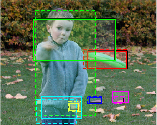}}
     {} 
&
\subf{\includegraphics[height=\detvisheight]{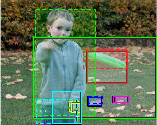}}
     {}
\\
\hline
\subf{\includegraphics[height=\detvisheight]{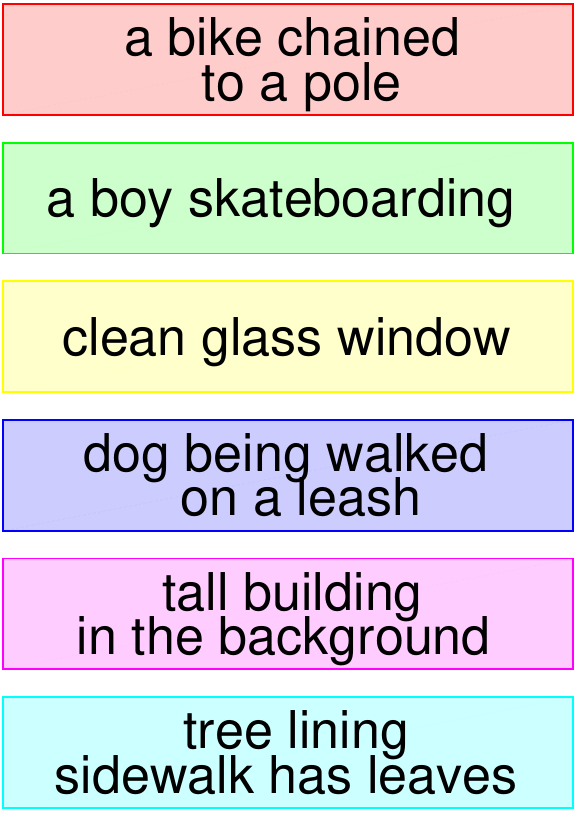}}
     {}
&
\subf{\includegraphics[height=\detvisheight]{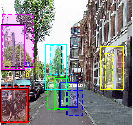}}
     {}    
&
\subf{\includegraphics[height=\detvisheight]{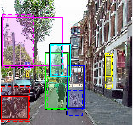}}
     {} 
&
\subf{\includegraphics[height=\detvisheight]{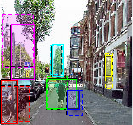}}
     {}
\\
\hline
\subf{\includegraphics[height=\detvisheight]{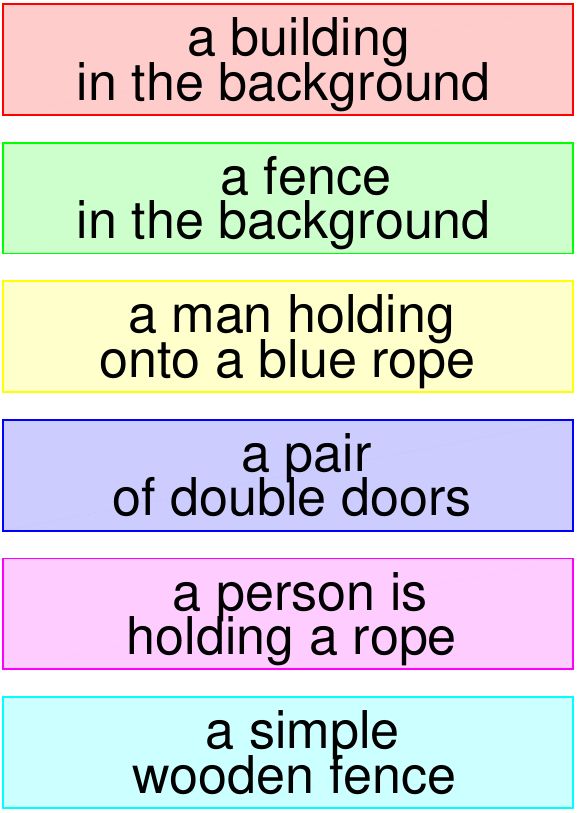}}
     {}
&
\subf{\includegraphics[height=\detvisheight]{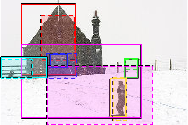}}
     {}    
&
\subf{\includegraphics[height=\detvisheight]{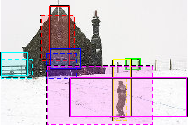}}
     {} 
&
\subf{\includegraphics[height=\detvisheight]{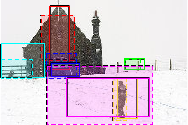}}
     {}
\end{tabular}
\repeatcaption{fig:det-vis-known-num}{\detviscaption{\detviscase}{six}}
\end{figure}
\vspace*{\fill}
\clearpage

\vspace*{\fill}
\begin{figure}[H]
\centering
\begin{tabular}{cccc}
\centering

\large Text phrases
&
\large DBNet
&
\large DenseCap
&
\large SCRC

\\
\hline
\subf{\includegraphics[height=\detvisheight]{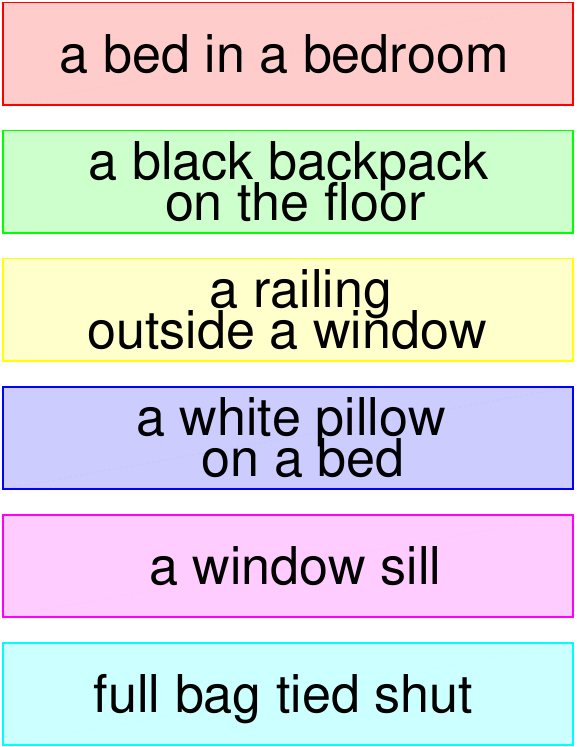}}
     {}
&
\subf{\includegraphics[height=\detvisheight]{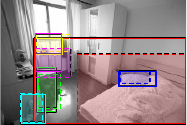}}
     {}    
&
\subf{\includegraphics[height=\detvisheight]{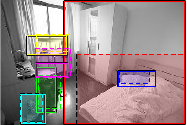}}
     {} 
&
\subf{\includegraphics[height=\detvisheight]{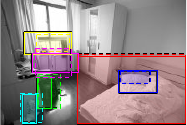}}
     {}
\\
\hline
\subf{\includegraphics[height=\detvisheight]{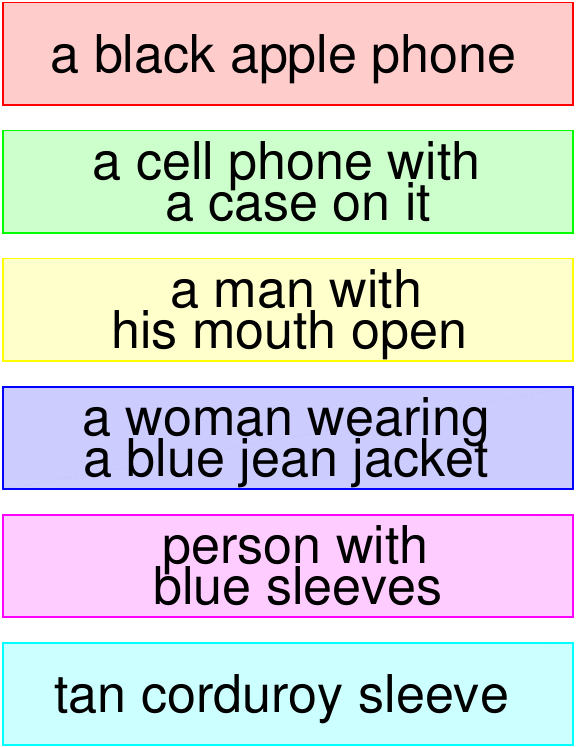}}
     {}
&
\subf{\includegraphics[height=\detvisheight]{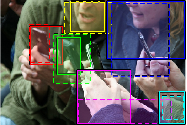}}
     {}    
&
\subf{\includegraphics[height=\detvisheight]{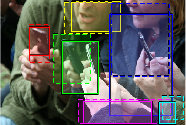}}
     {} 
&
\subf{\includegraphics[height=\detvisheight]{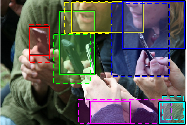}}
     {}
\\
\hline
\subf{\includegraphics[height=\detvisheight]{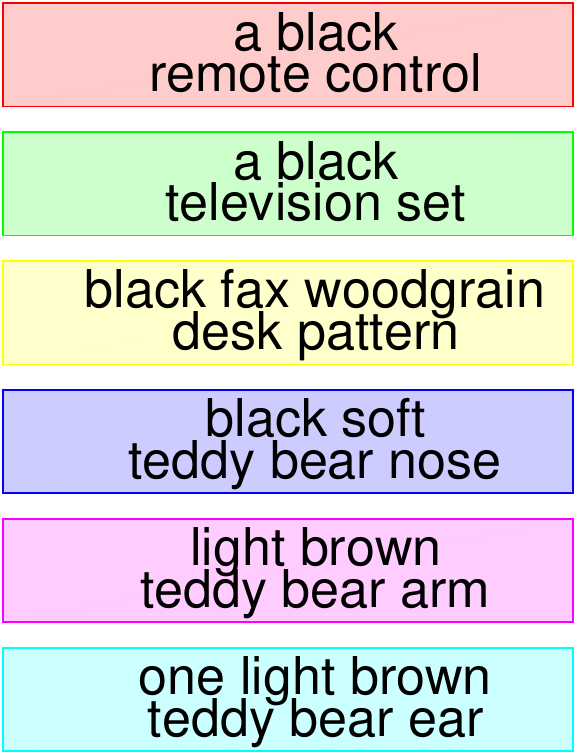}}
     {}
&
\subf{\includegraphics[height=\detvisheight]{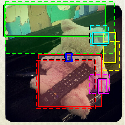}}
     {}    
&
\subf{\includegraphics[height=\detvisheight]{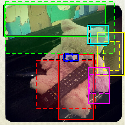}}
     {} 
&
\subf{\includegraphics[height=\detvisheight]{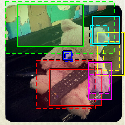}}
     {}
\end{tabular}
\repeatcaption{fig:det-vis-known-num}{\detviscaption{\detviscase}{six}}
\end{figure}
\vspace*{\fill}

\clearpage
 
\subsection{Random detection results with phrase-dependent thresholds}
\label{sec:more-vis-det-phrase-dependent}

In Figure~\ref{fig:det-vis-phrase-dependent}, we used phrase-dependent decision thresholds to determine how many regions were detected on an image. 
We set the threshold to make the detection precision for the IoU threshold at $0.5$ equal to $0.5$ when applicable. 
DBNet outperformed DenseCap and SCRC significantly. 
DenseCap and SCRC resulted in many cases of false alarms or miss detection.  
Note that DBNet could usually achieve the $0.5$ precision with a reasonable recall level, but DenseCap and SCRC might either fail achieving the $0.5$ precision at all or give a low recall.

\renewcommand{\detvistextheight}{0.15\textwidth}

\renewcommand{\detviscase}{using phrase-dependent detection threshold}

\begin{centering}
\vspace*{\fill}
\begin{figure}[H]
\begin{tabular}{cccc}
\centering

\large Text phrases
&
\large DBNet
&
\large DenseCap
&
\large SCRC

\\
\hline
\subf{\includegraphics[height=\detvistextheight]{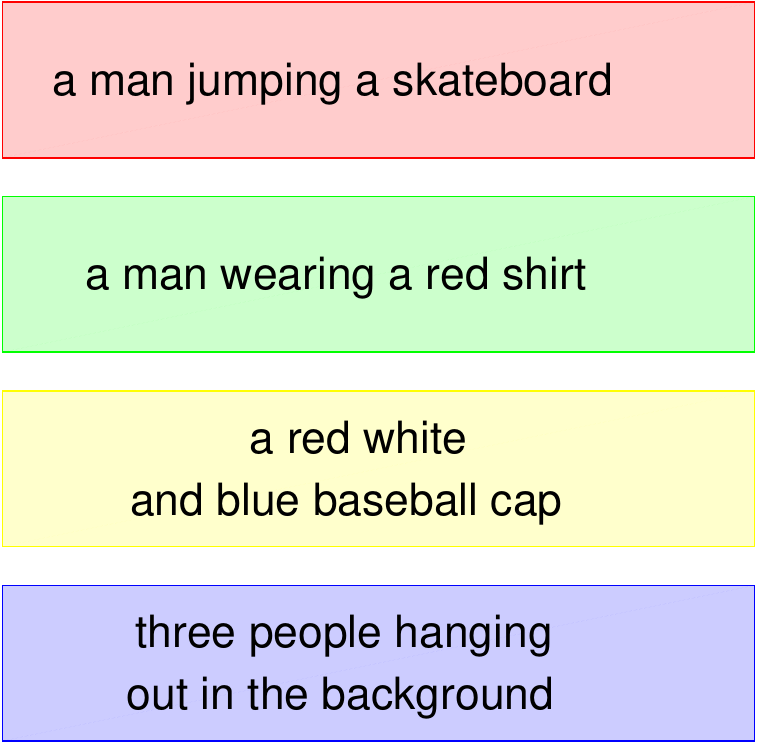}}
     {}
&
\subf{\includegraphics[height=\detvisheight]{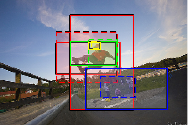}}
     {}    
&
\subf{\includegraphics[height=\detvisheight]{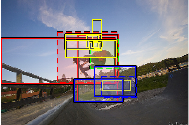}}
     {} 
&
\subf{\includegraphics[height=\detvisheight]{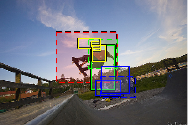}}
     {}
\\
\hline
\subf{\includegraphics[height=\detvistextheight]{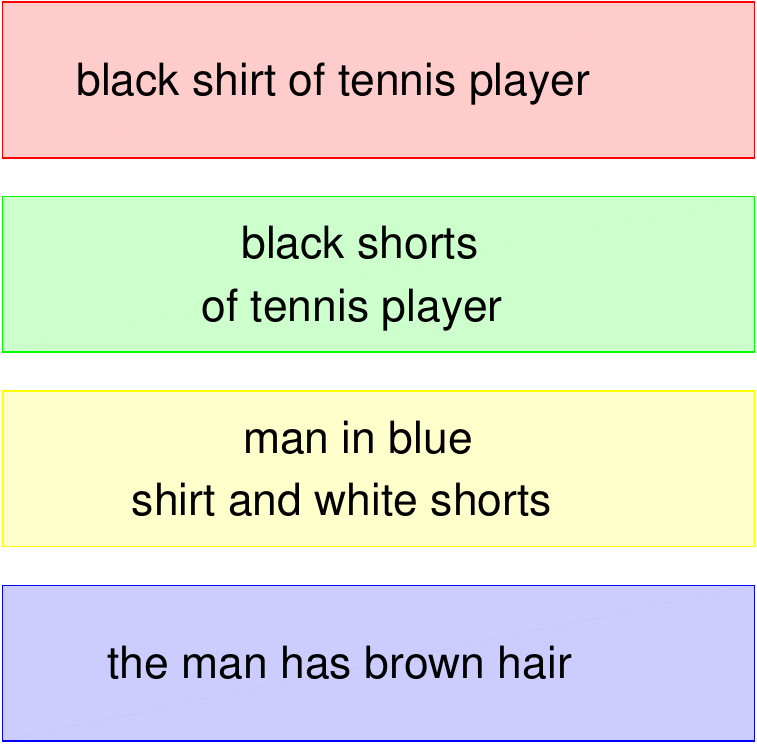}}
     {}
&
\subf{\includegraphics[height=\detvisheight]{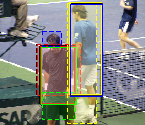}}
     {}    
&
\subf{\includegraphics[height=\detvisheight]{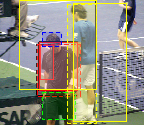}}
     {} 
&
\subf{\includegraphics[height=\detvisheight]{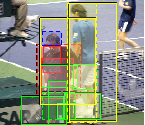}}
     {}
\\
\hline
\subf{\includegraphics[height=\detvistextheight]{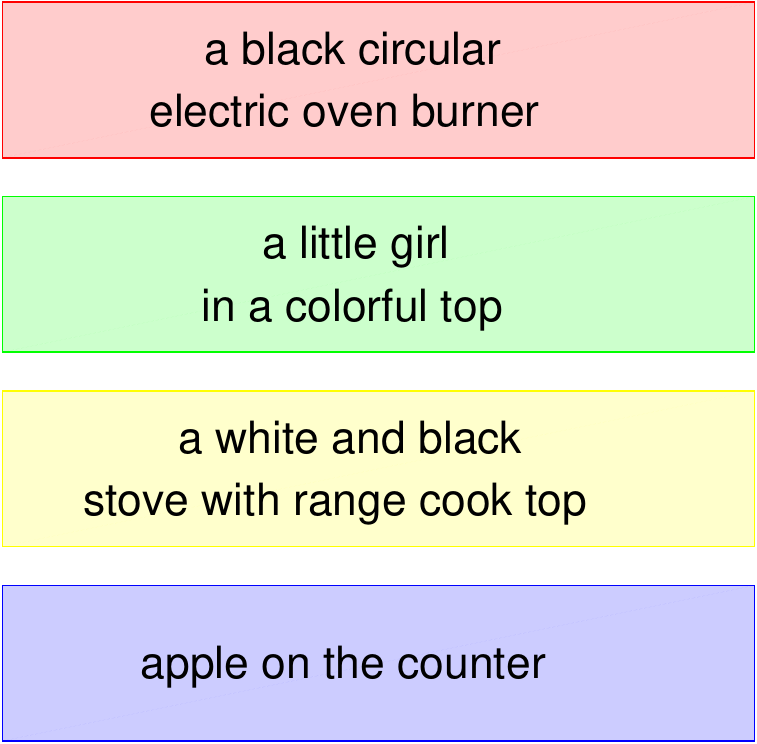}}
     {}
&
\subf{\includegraphics[height=\detvisheight]{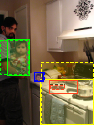}}
     {}    
&
\subf{\includegraphics[height=\detvisheight]{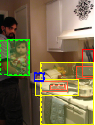}}
     {} 
&
\subf{\includegraphics[height=\detvisheight]{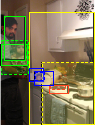}}
     {}
\end{tabular}
\caption{\detviscaption{\detviscase}{four}
\label{fig:det-vis-phrase-dependent} }
\end{figure}
\vspace*{\fill}
\end{centering}
\clearpage

\vspace*{\fill}
\begin{figure}[H]
\centering
\begin{tabular}{cccc}
\centering

\large Text phrases
&
\large DBNet
&
\large DenseCap
&
\large SCRC

\\
\hline
\subf{\includegraphics[height=\detvistextheight]{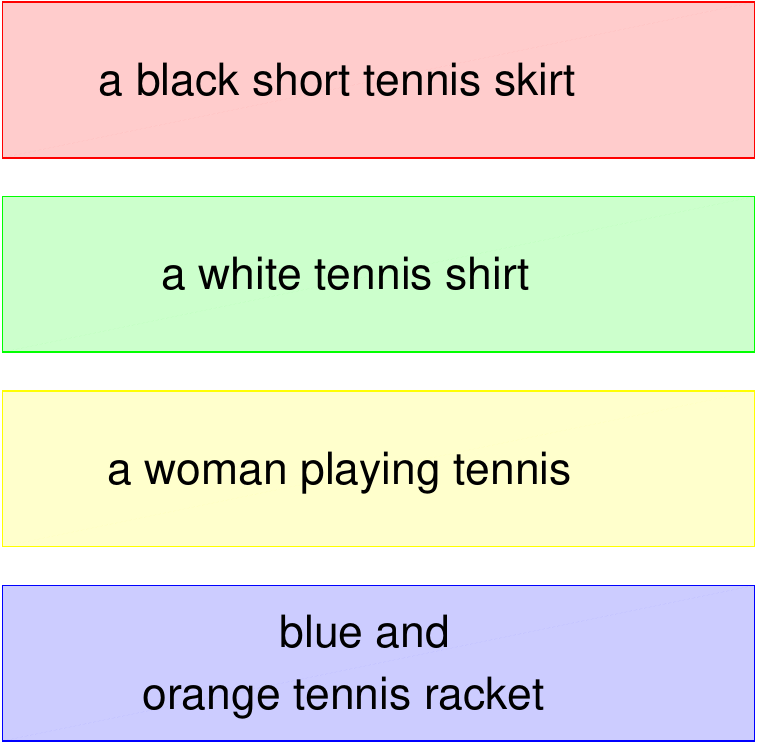}}
     {}
&
\subf{\includegraphics[height=\detvisheight]{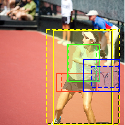}}
     {}    
&
\subf{\includegraphics[height=\detvisheight]{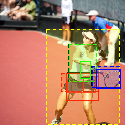}}
     {} 
&
\subf{\includegraphics[height=\detvisheight]{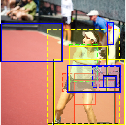}}
     {}
\\
\hline
\subf{\includegraphics[height=\detvistextheight]{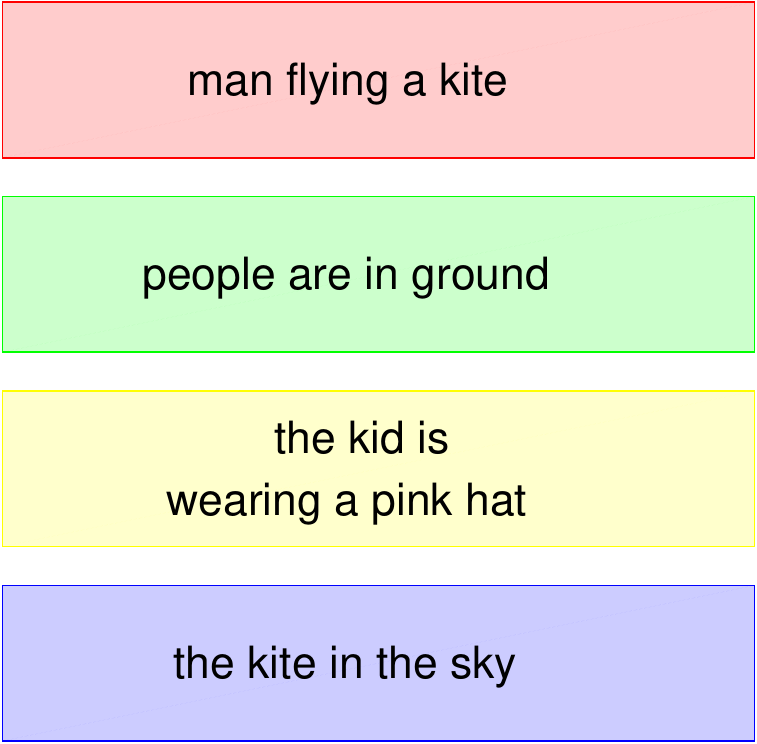}}
     {}
&
\subf{\includegraphics[height=\detvisheight]{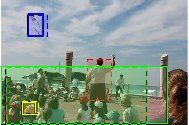}}
     {}    
&
\subf{\includegraphics[height=\detvisheight]{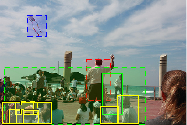}}
     {} 
&
\subf{\includegraphics[height=\detvisheight]{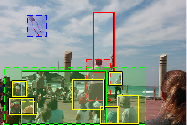}}
     {}
\\
\hline
\subf{\includegraphics[height=\detvistextheight]{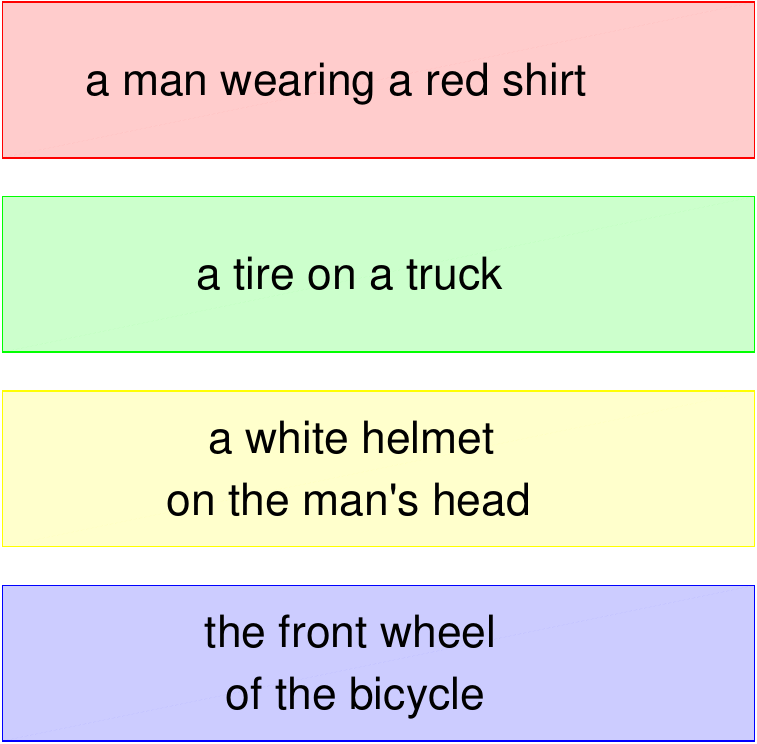}}
     {}
&
\subf{\includegraphics[height=\detvisheight]{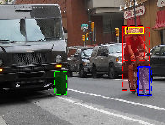}}
     {}    
&
\subf{\includegraphics[height=\detvisheight]{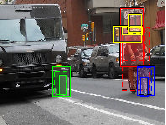}}
     {} 
&
\subf{\includegraphics[height=\detvisheight]{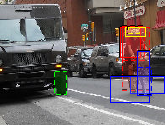}}
     {}
\end{tabular}
\repeatcaption{fig:det-vis-phrase-dependent}{\detviscaption{\detviscase}{four}}
\end{figure}
\vspace*{\fill}
\clearpage

\vspace*{\fill}
\begin{figure}[H]
\centering
\begin{tabular}{cccc}
\centering

\large Text phrases
&
\large DBNet
&
\large DenseCap
&
\large SCRC

\\
\hline
\subf{\includegraphics[height=\detvistextheight]{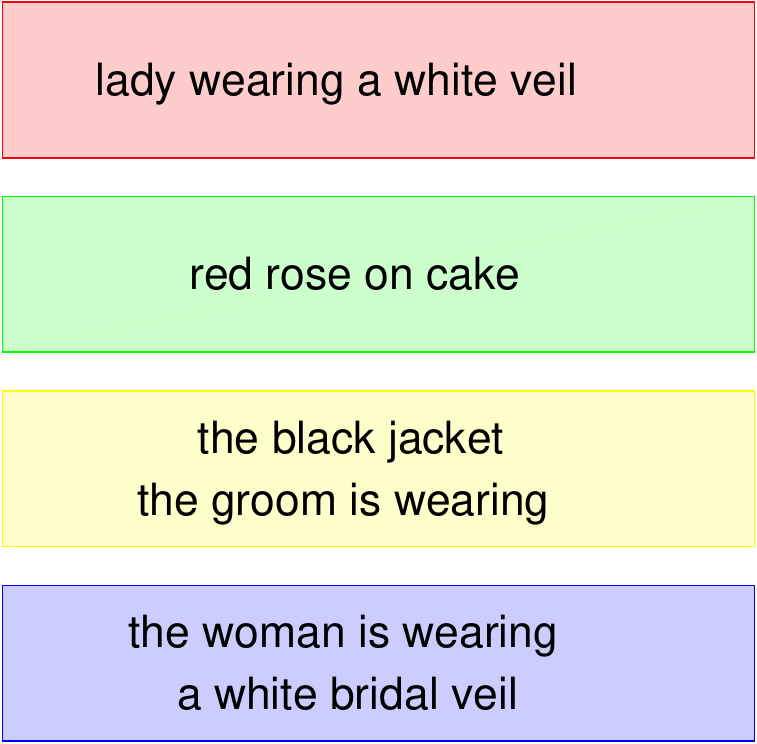}}
     {}
&
\subf{\includegraphics[height=\detvisheight]{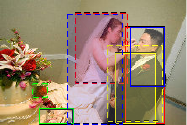}}
     {}    
&
\subf{\includegraphics[height=\detvisheight]{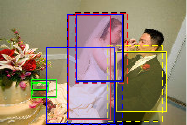}}
     {} 
&
\subf{\includegraphics[height=\detvisheight]{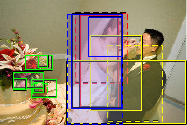}}
     {}
\\
\hline
\subf{\includegraphics[height=\detvistextheight]{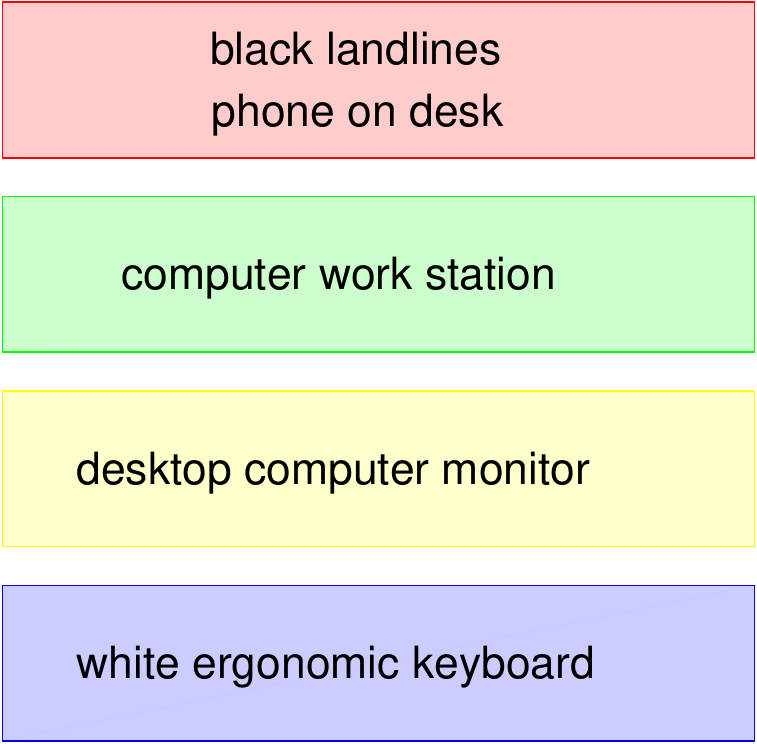}}
     {}
&
\subf{\includegraphics[height=\detvisheight]{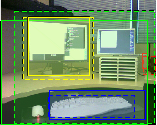}}
     {}    
&
\subf{\includegraphics[height=\detvisheight]{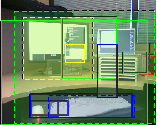}}
     {} 
&
\subf{\includegraphics[height=\detvisheight]{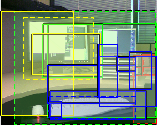}}
     {}
\\
\hline
\subf{\includegraphics[height=\detvistextheight]{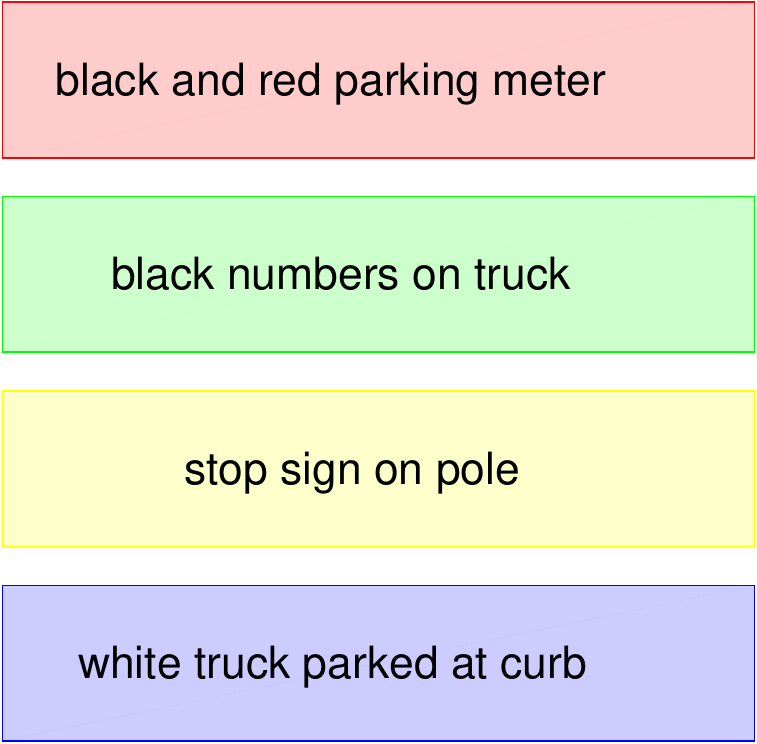}}
     {}
&
\subf{\includegraphics[height=\detvisheight]{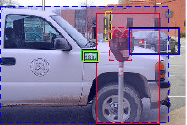}}
     {}    
&
\subf{\includegraphics[height=\detvisheight]{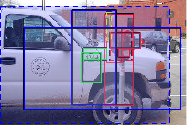}}
     {} 
&
\subf{\includegraphics[height=\detvisheight]{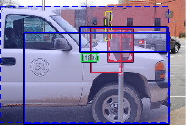}}
     {}
\end{tabular}
\repeatcaption{fig:det-vis-phrase-dependent}{\detviscaption{\detviscase}{four}}
\end{figure}
\vspace*{\fill}
\clearpage

\vspace*{\fill}
\begin{figure}[H]
\centering
\begin{tabular}{cccc}
\centering

\large Text phrases
&
\large DBNet
&
\large DenseCap
&
\large SCRC

\\
\hline
\subf{\includegraphics[height=\detvistextheight]{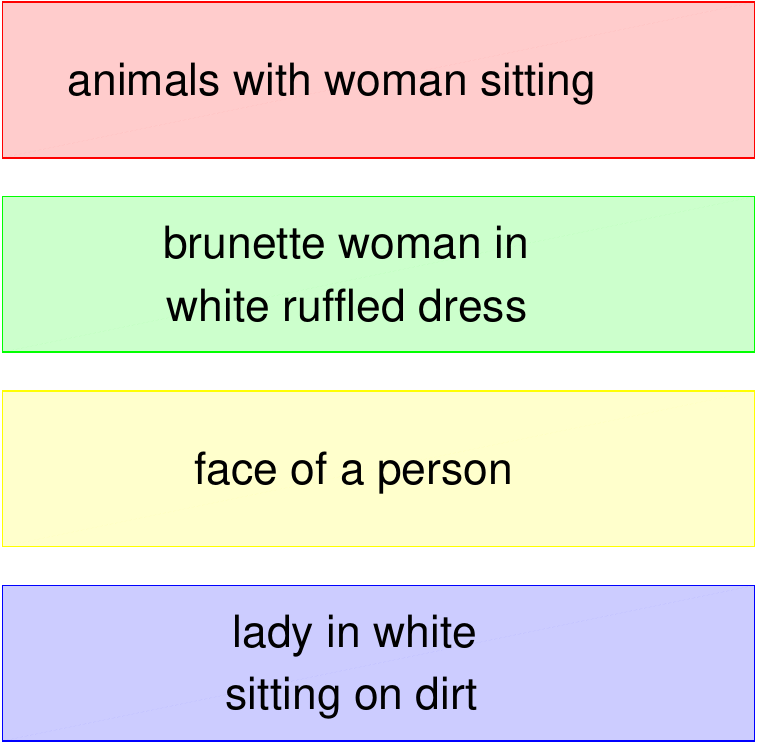}}
     {}
&
\subf{\includegraphics[height=\detvisheight]{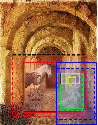}}
     {}    
&
\subf{\includegraphics[height=\detvisheight]{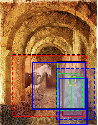}}
     {} 
&
\subf{\includegraphics[height=\detvisheight]{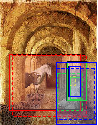}}
     {}
\\
\hline
\subf{\includegraphics[height=\detvistextheight]{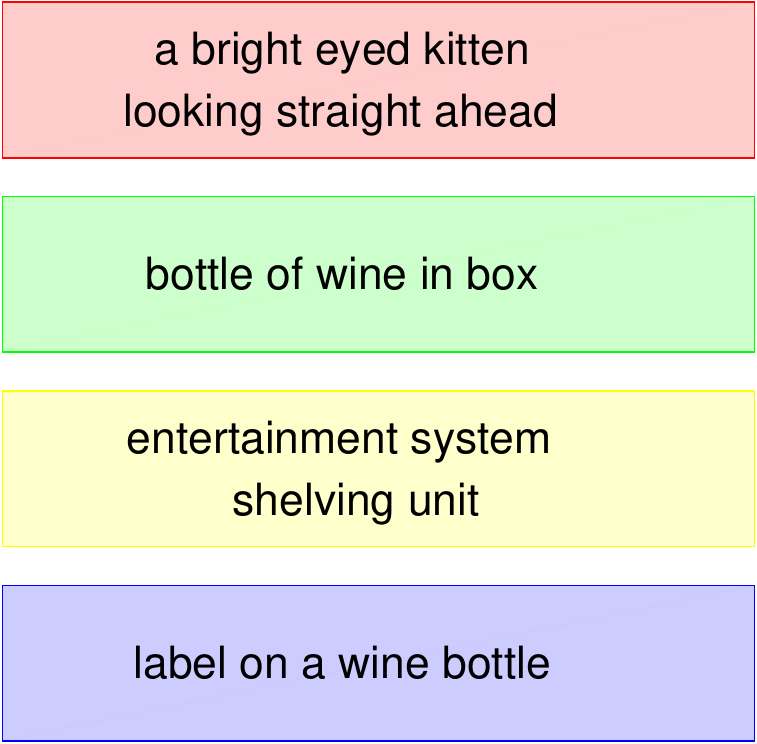}}
     {}
&
\subf{\includegraphics[height=\detvisheight]{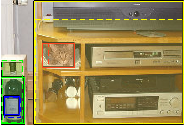}}
     {}    
&
\subf{\includegraphics[height=\detvisheight]{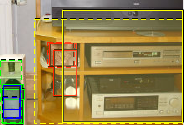}}
     {} 
&
\subf{\includegraphics[height=\detvisheight]{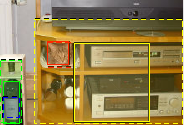}}
     {}
\\
\hline
\subf{\includegraphics[height=\detvistextheight]{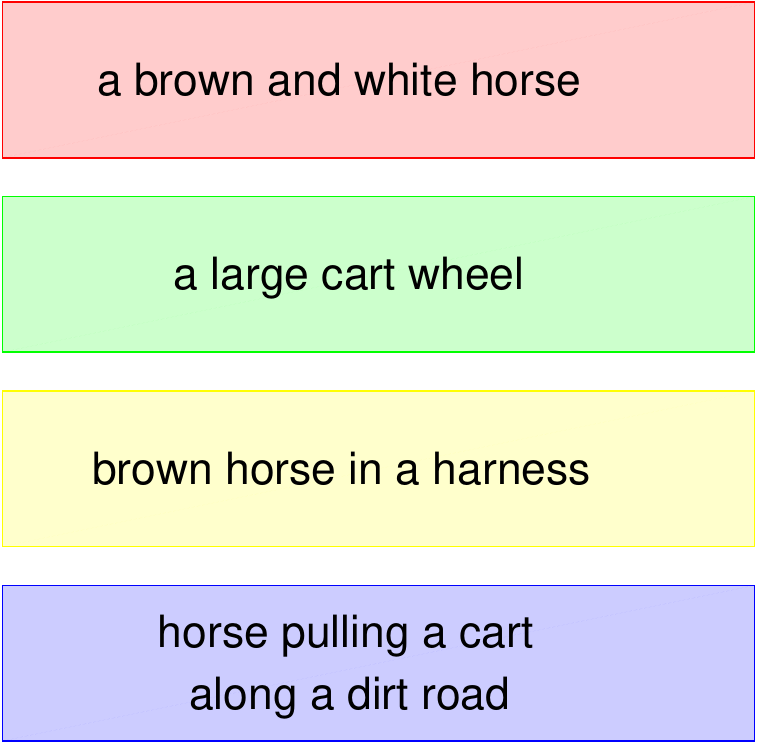}}
     {}
&
\subf{\includegraphics[height=\detvisheight]{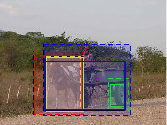}}
     {}    
&
\subf{\includegraphics[height=\detvisheight]{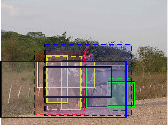}}
     {} 
&
\subf{\includegraphics[height=\detvisheight]{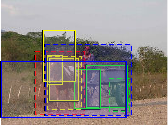}}
     {}
\end{tabular}
\repeatcaption{fig:det-vis-phrase-dependent}{\detviscaption{\detviscase}{four}}
\end{figure}
\vspace*{\fill}
\clearpage

\vspace*{\fill}
\begin{figure}[H]
\centering
\begin{tabular}{cccc}
\centering

\large Text phrases
&
\large DBNet
&
\large DenseCap
&
\large SCRC

\\
\hline
\subf{\includegraphics[height=\detvistextheight]{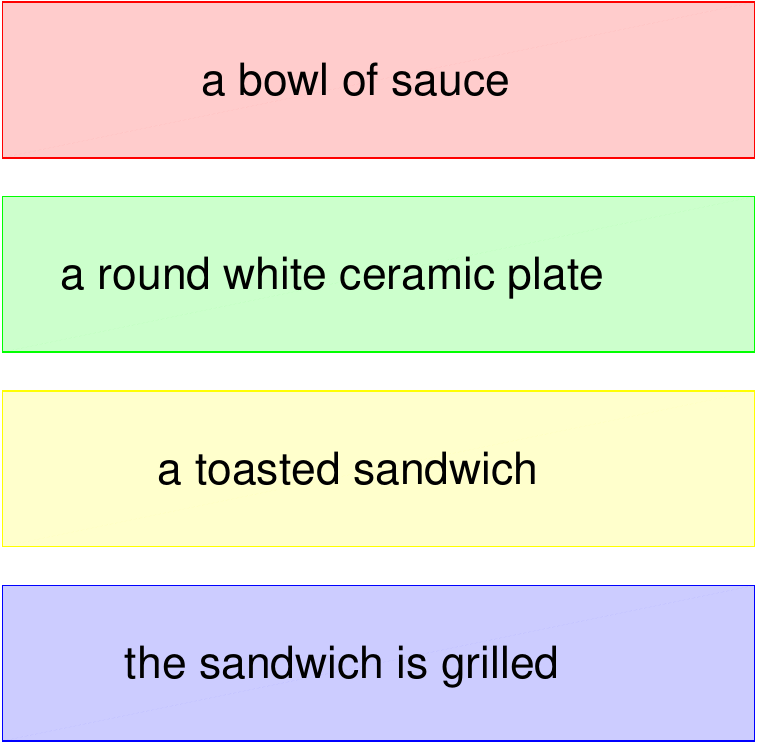}}
     {}
&
\subf{\includegraphics[height=\detvisheight]{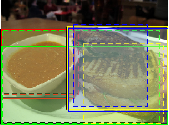}}
     {}    
&
\subf{\includegraphics[height=\detvisheight]{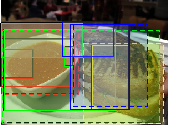}}
     {} 
&
\subf{\includegraphics[height=\detvisheight]{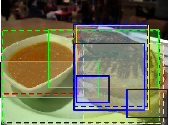}}
     {}
\\
\hline
\subf{\includegraphics[height=\detvistextheight]{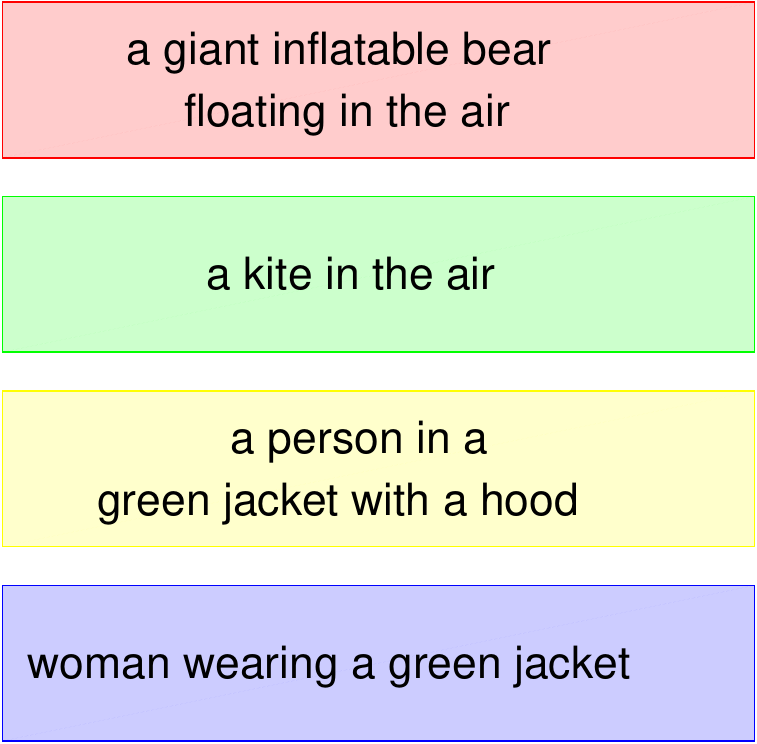}}
     {}
&
\subf{\includegraphics[height=\detvisheight]{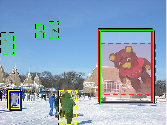}}
     {}    
&
\subf{\includegraphics[height=\detvisheight]{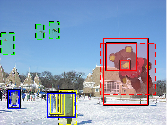}}
     {} 
&
\subf{\includegraphics[height=\detvisheight]{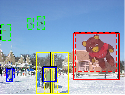}}
     {}
\\
\hline
\subf{\includegraphics[height=\detvistextheight]{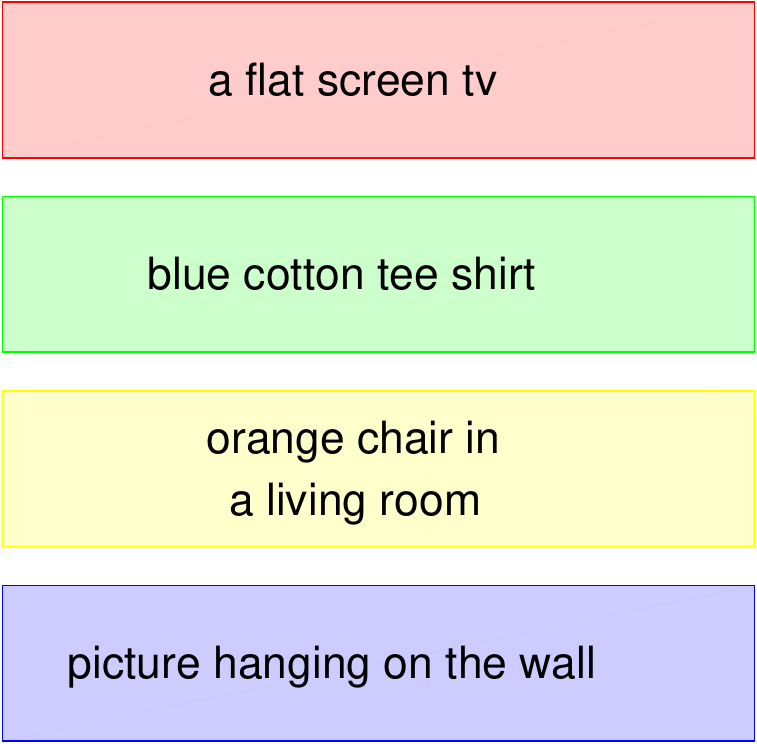}}
     {}
&
\subf{\includegraphics[height=\detvisheight]{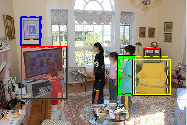}}
     {}    
&
\subf{\includegraphics[height=\detvisheight]{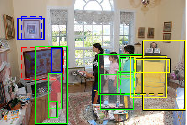}}
     {} 
&
\subf{\includegraphics[height=\detvisheight]{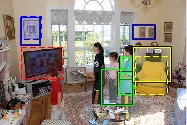}}
     {}
\end{tabular}
\repeatcaption{fig:det-vis-phrase-dependent}{\detviscaption{\detviscase}{four}}
\end{figure}
\vspace*{\fill}
\clearpage

\vspace*{\fill}
\begin{figure}[H]
\centering
\begin{tabular}{cccc}
\centering

\large Text phrases
&
\large DBNet
&
\large DenseCap
&
\large SCRC

\\
\hline
\subf{\includegraphics[height=\detvistextheight]{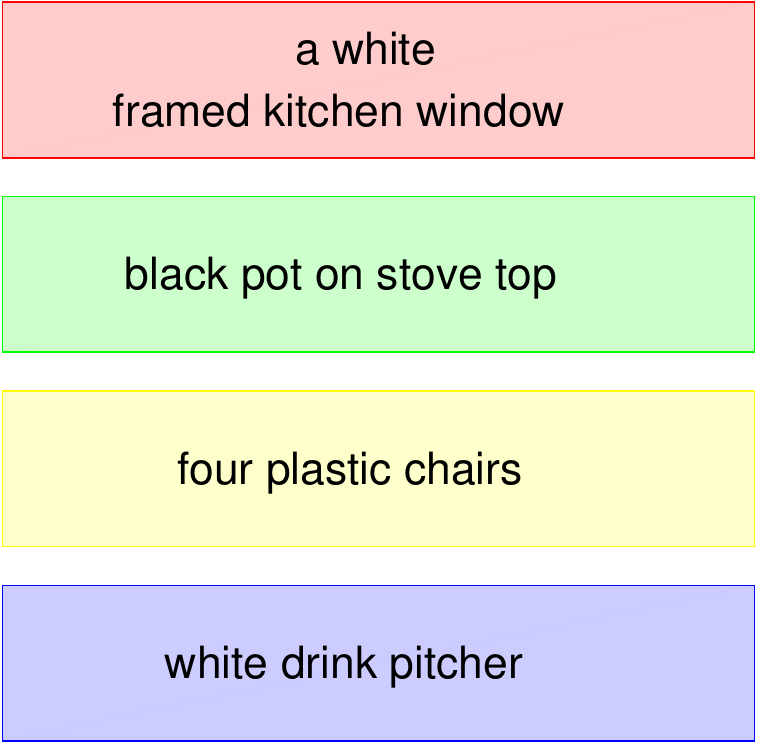}}
     {}
&
\subf{\includegraphics[height=\detvisheight]{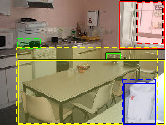}}
     {}    
&
\subf{\includegraphics[height=\detvisheight]{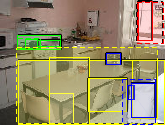}}
     {} 
&
\subf{\includegraphics[height=\detvisheight]{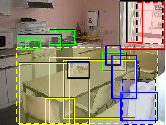}}
     {}
\\
\hline
\subf{\includegraphics[height=\detvistextheight]{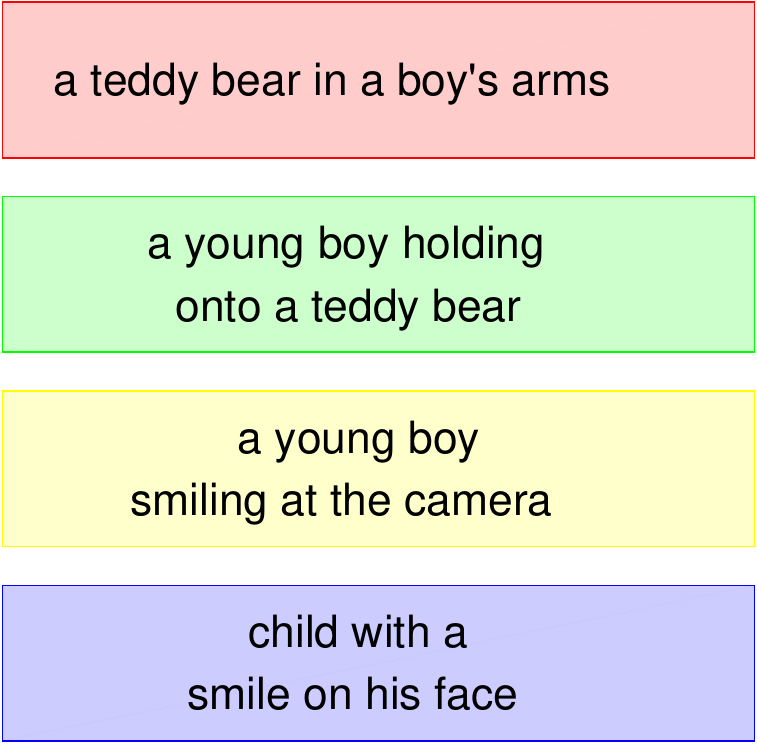}}
     {}
&
\subf{\includegraphics[height=\detvisheight]{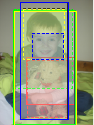}}
     {}    
&
\subf{\includegraphics[height=\detvisheight]{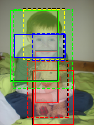}}
     {} 
&
\subf{\includegraphics[height=\detvisheight]{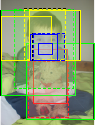}}
     {}
\\
\hline
\subf{\includegraphics[height=\detvistextheight]{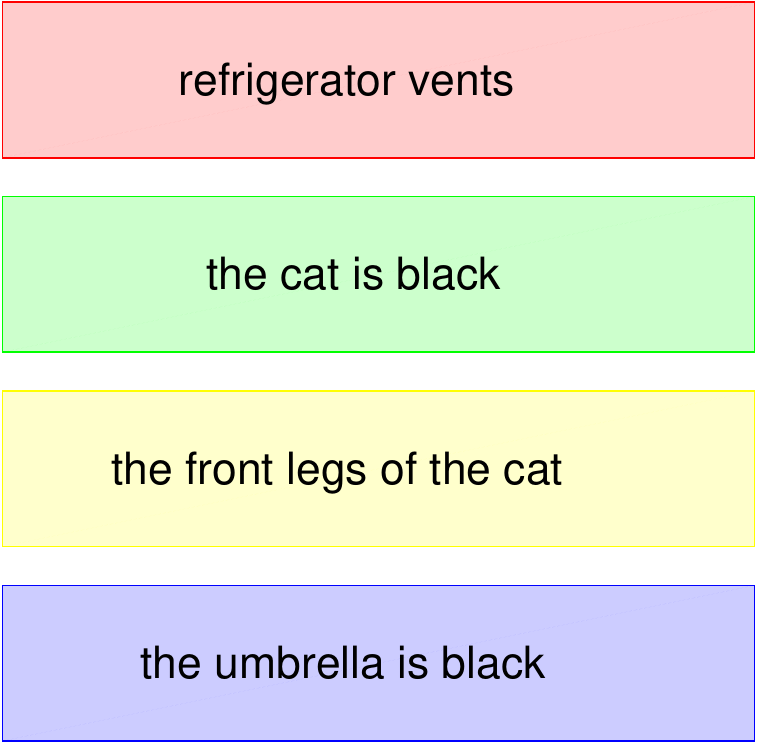}}
     {}
&
\subf{\includegraphics[height=\detvisheight]{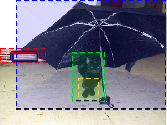}}
     {}    
&
\subf{\includegraphics[height=\detvisheight]{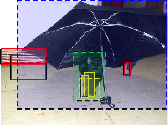}}
     {} 
&
\subf{\includegraphics[height=\detvisheight]{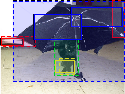}}
     {}
\end{tabular}
\repeatcaption{fig:det-vis-phrase-dependent}{\detviscaption{\detviscase}{four}}
\end{figure}
\vspace*{\fill}
\clearpage

\vspace*{\fill}
\begin{figure}[H]
\centering
\begin{tabular}{cccc}
\centering

\large Text phrases
&
\large DBNet
&
\large DenseCap
&
\large SCRC

\\
\hline
\subf{\includegraphics[height=\detvistextheight]{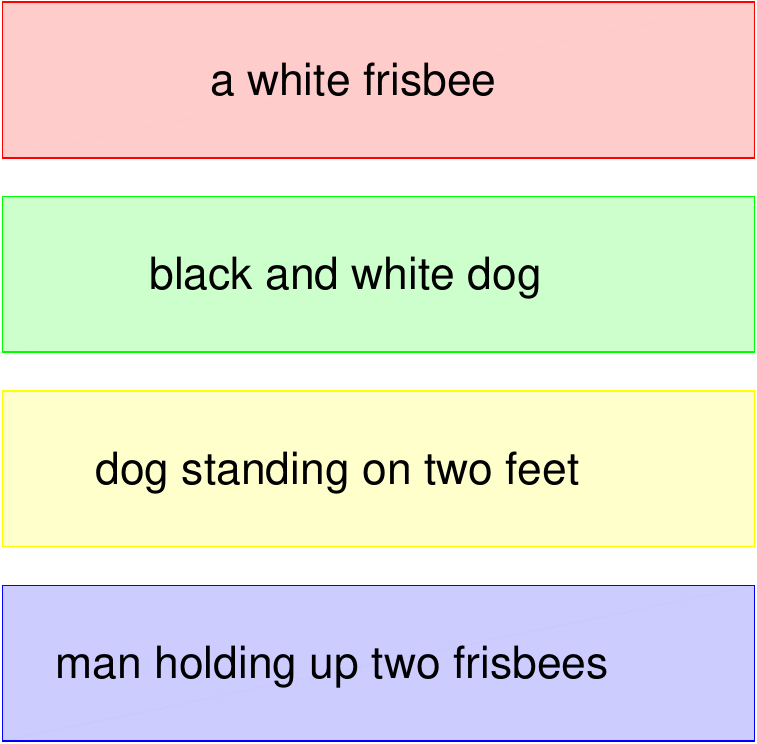}}
     {}
&
\subf{\includegraphics[height=\detvisheight]{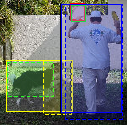}}
     {}    
&
\subf{\includegraphics[height=\detvisheight]{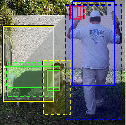}}
     {} 
&
\subf{\includegraphics[height=\detvisheight]{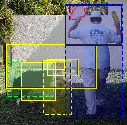}}
     {}
\\
\hline
\subf{\includegraphics[height=\detvistextheight]{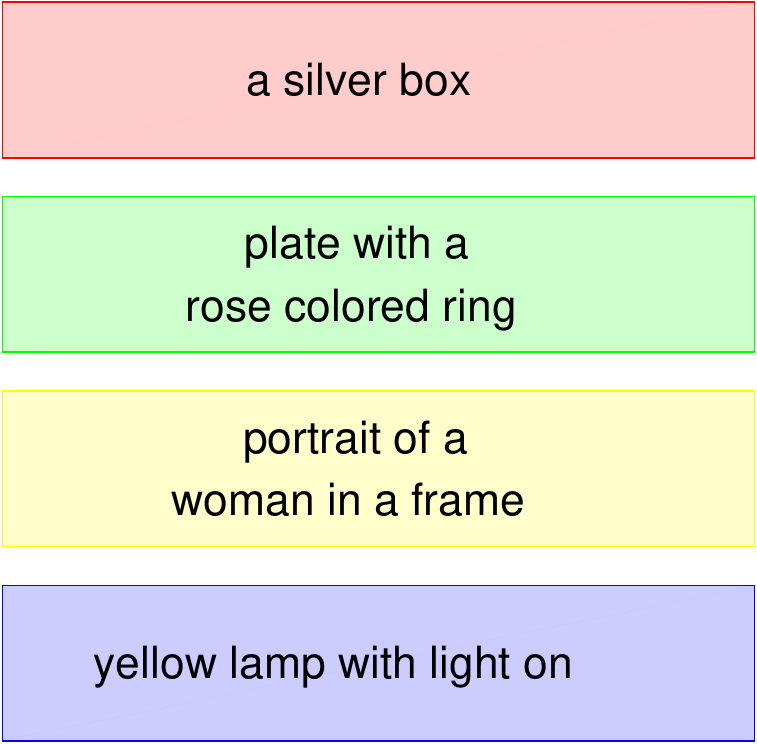}}
     {}
&
\subf{\includegraphics[height=\detvisheight]{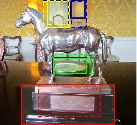}}
     {}    
&
\subf{\includegraphics[height=\detvisheight]{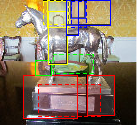}}
     {} 
&
\subf{\includegraphics[height=\detvisheight]{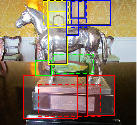}}
     {}
\\
\subf{\includegraphics[height=\detvistextheight]{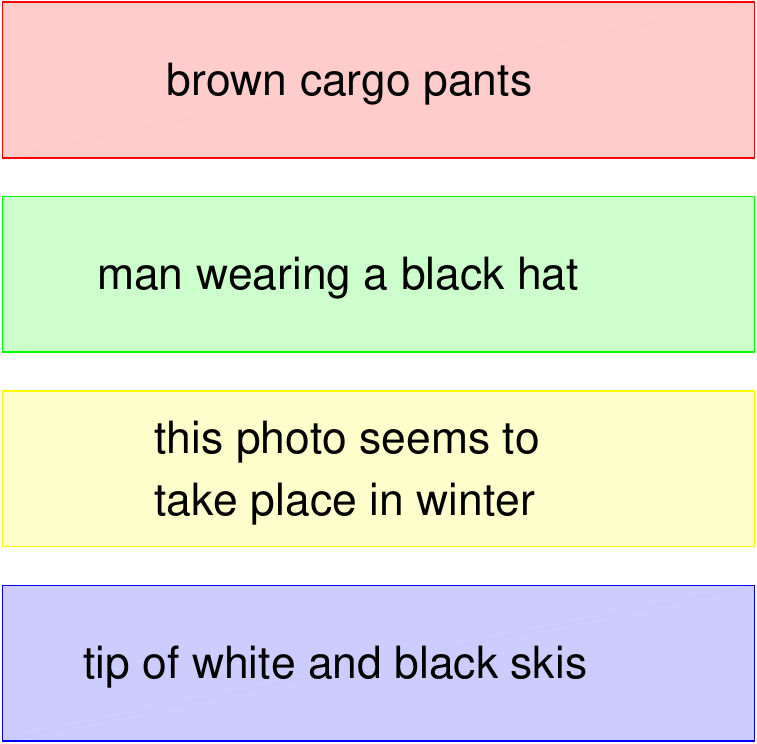}}
     {}
&
\subf{\includegraphics[height=\detvisheight]{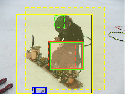}}
     {}    
&
\subf{\includegraphics[height=\detvisheight]{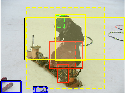}}
     {} 
&
\subf{\includegraphics[height=\detvisheight]{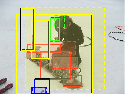}}
     {}
\end{tabular}
\repeatcaption{fig:det-vis-phrase-dependent}{\detviscaption{\detviscase}{four}}
\end{figure}
\vspace*{\fill}
\clearpage

\vspace*{\fill}
\begin{figure}[H]
\centering
\begin{tabular}{cccc}
\centering

\large Text phrases
&
\large DBNet
&
\large DenseCap
&
\large SCRC

\\
\hline
\subf{\includegraphics[height=\detvistextheight]{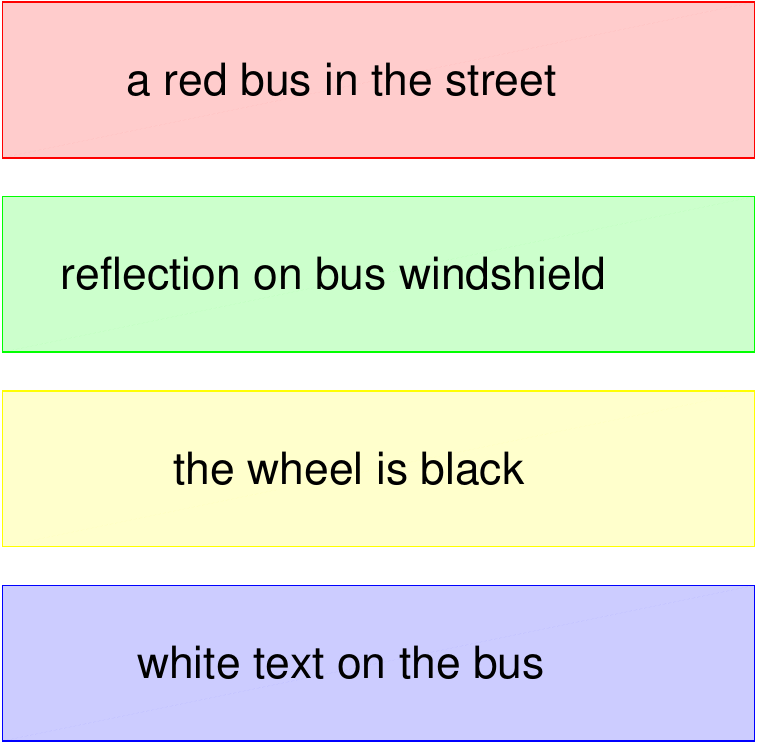}}
     {}
&
\subf{\includegraphics[height=\detvisheight]{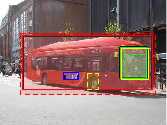}}
     {}    
&
\subf{\includegraphics[height=\detvisheight]{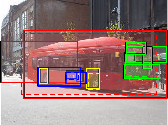}}
     {} 
&
\subf{\includegraphics[height=\detvisheight]{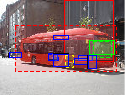}}
     {}
\\
\hline
\subf{\includegraphics[height=\detvistextheight]{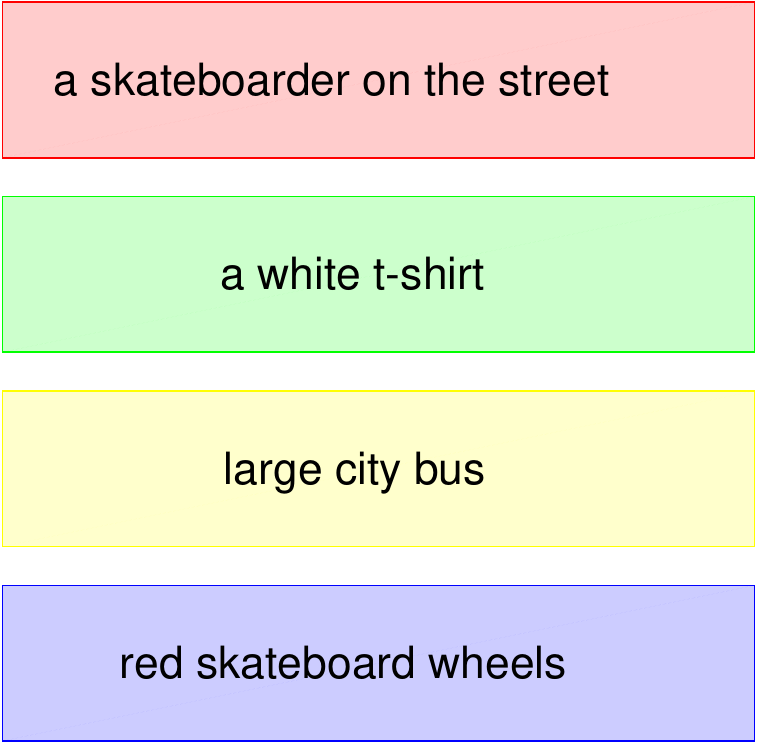}}
     {}
&
\subf{\includegraphics[height=\detvisheight]{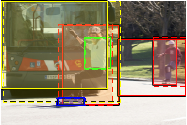}}
     {}    
&
\subf{\includegraphics[height=\detvisheight]{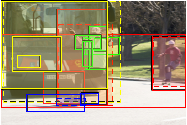}}
     {} 
&
\subf{\includegraphics[height=\detvisheight]{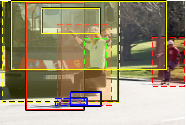}}
     {}
\\
\hline
\subf{\includegraphics[height=\detvistextheight]{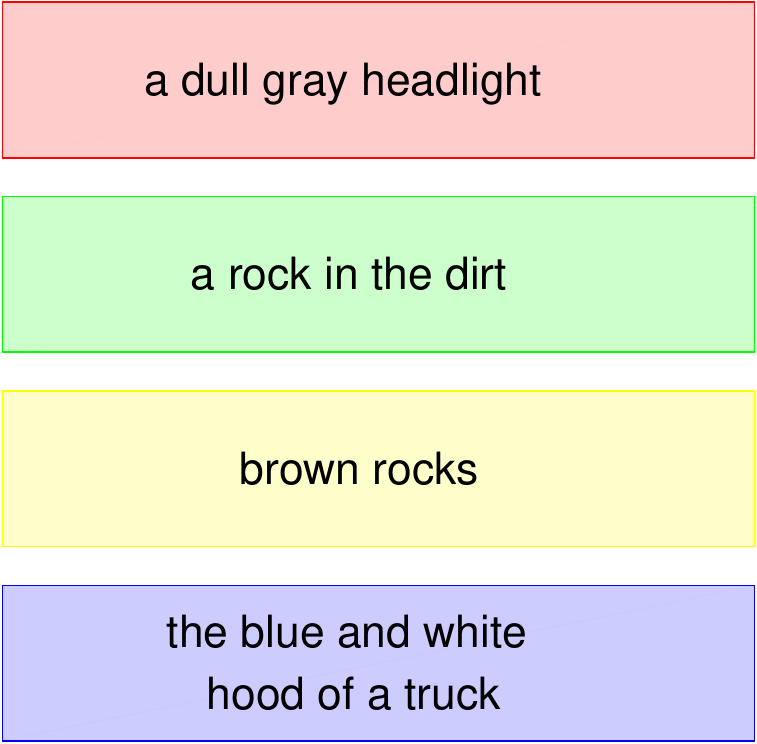}}
     {}
&
\subf{\includegraphics[height=\detvisheight]{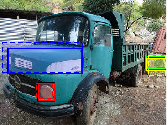}}
     {}    
&
\subf{\includegraphics[height=\detvisheight]{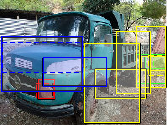}}
     {} 
&
\subf{\includegraphics[height=\detvisheight]{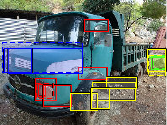}}
     {}
\end{tabular}
\repeatcaption{fig:det-vis-phrase-dependent}{\detviscaption{\detviscase}{four}}
\end{figure}
\vspace*{\fill}
\clearpage

\vspace*{\fill}
\begin{figure}[H]
\centering
\begin{tabular}{cccc}
\centering

\large Text phrases
&
\large DBNet
&
\large DenseCap
&
\large SCRC

\\
\hline
\subf{\includegraphics[height=\detvistextheight]{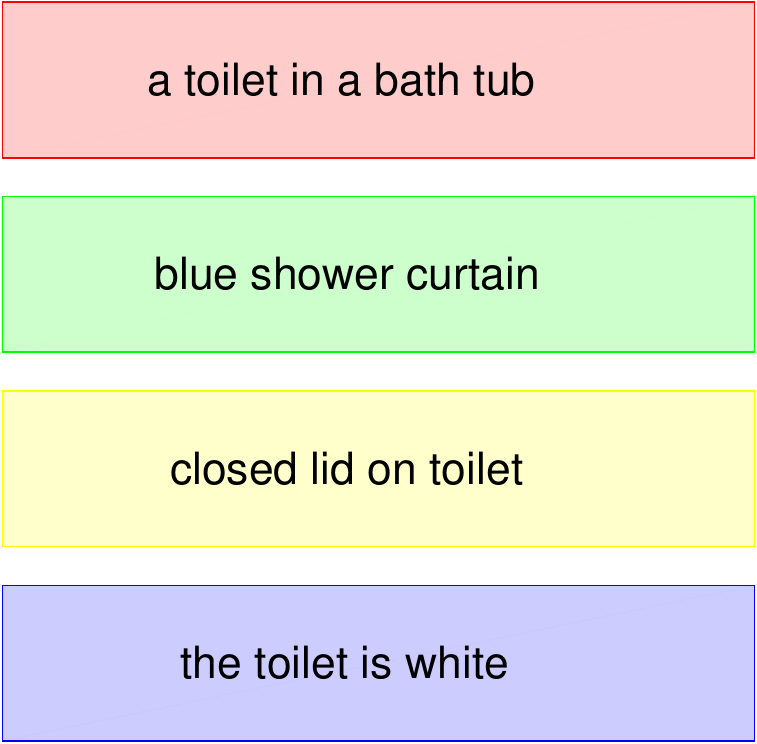}}
     {}
&
\subf{\includegraphics[height=\detvisheight]{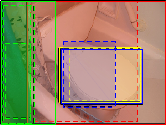}}
     {}    
&
\subf{\includegraphics[height=\detvisheight]{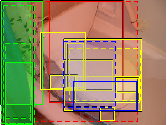}}
     {} 
&
\subf{\includegraphics[height=\detvisheight]{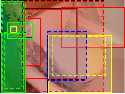}}
     {}
\\
\hline
\subf{\includegraphics[height=\detvistextheight]{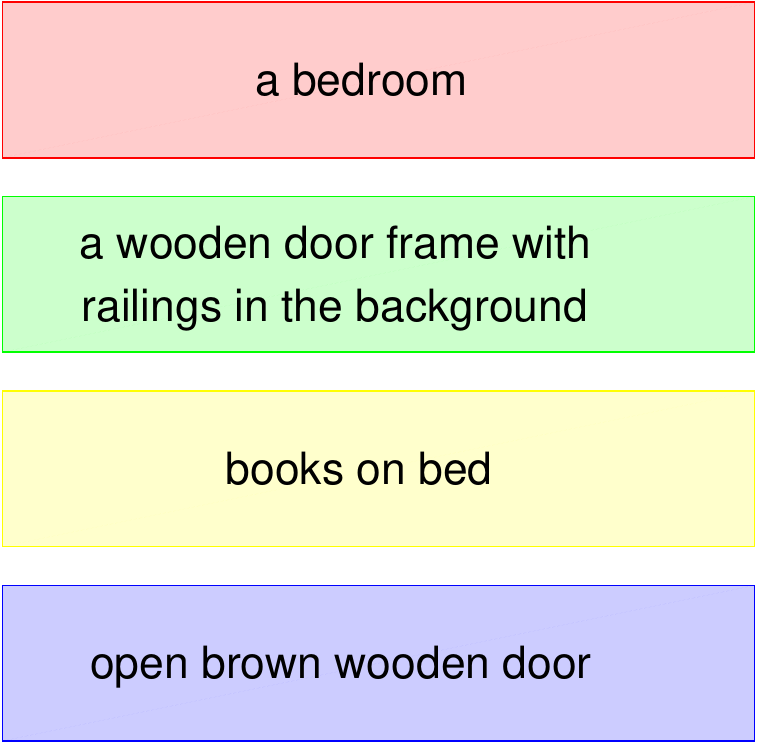}}
     {}
&
\subf{\includegraphics[height=\detvisheight]{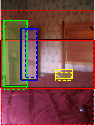}}
     {}    
&
\subf{\includegraphics[height=\detvisheight]{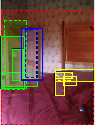}}
     {} 
&
\subf{\includegraphics[height=\detvisheight]{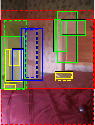}}
     {}
\\
\hline
\subf{\includegraphics[height=\detvistextheight]{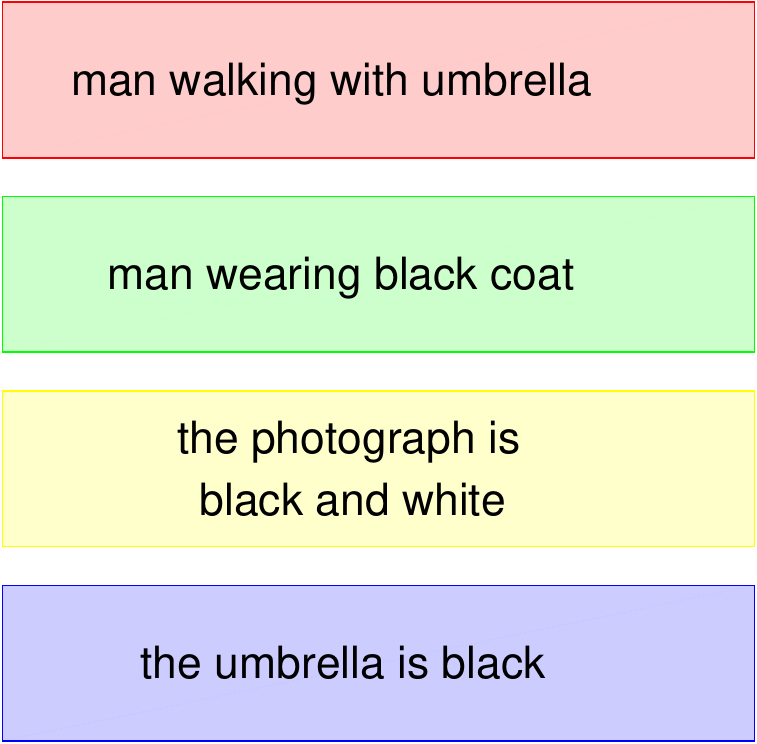}}
     {}
&
\subf{\includegraphics[height=\detvisheight]{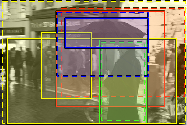}}
     {}    
&
\subf{\includegraphics[height=\detvisheight]{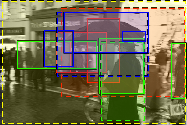}}
     {} 
&
\subf{\includegraphics[height=\detvisheight]{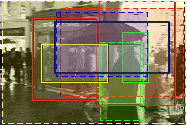}}
     {}
\end{tabular}
\repeatcaption{fig:det-vis-phrase-dependent}{\detviscaption{\detviscase}{four}}
\end{figure}
\vspace*{\fill}
\clearpage

\clearpage

\subsection{Failure cases for detection with phrase-dependent thresholds}
\label{sec:more-vis-det-failures}

\arxvspace{-6pt}

\newcommand{\psubf}[1]{\parbox{0.35\textheight}{\subf{#1}}\hspace*{0.5em}}
\newcommand{\labeltext}[1]{\begin{minipage}[c][\detvisheight]{\detvistextwidth}
\begin{flushleft}
#1
\par\end{flushleft}
\end{minipage}
}

In this section, we used phrase-dependent decision thresholds in the same way as in Section~\ref{sec:more-vis-det-phrase-dependent}, except for focusing on showing failure cases. 
We visualized randomly chosen testing images and phrases under the constraint that at least one of DBNet, DenseCap, and SCRC should significantly fail in detection (i.e., IoU with ground truth is less than 0.2).
In Figure~\ref{fig:det-vis-failures}, we categorized failure cases into three types: 1)~the false alarm (the detected box has no overlap with any ground truth), 2)~inaccurate localization (the IoU with ground truth is less than $0.5$), 3)~missing detection (no detection box has overlap with a ground truth region).
For each image, we showed only one phrase for visual clarity and displayed the failure types for comprehensiveness.
DBNet has significantly less failure cases than DenseCap and SCRC.

\renewcommand{\detviscase}{using phrase-dependent detection threshold}

\arxvspace{-10pt}

\begin{centering}
\vspace*{\fill}
\begin{figure}[H]
\begin{tabular}{c|ccc}
\centering

\large Text phrases
&
\large DBNet
&
\large DenseCap
&
\large SCRC

\\
\hline
\labeltext{  a man with dark hair eating outside}
&
\psubf{\includegraphics[height=\detvisheight]{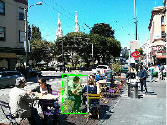}}
     {}    
&
\psubf{\includegraphics[height=\detvisheight]{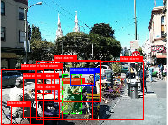}}
     {} 
&
\psubf{\includegraphics[height=\detvisheight]{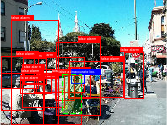}}
     {}
\\
\hline
\labeltext{  a group of swimmers in the ocean}
&
\psubf{\includegraphics[height=\detvisheight]{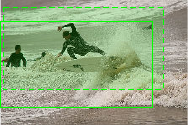}}
     {}    
&
\psubf{\includegraphics[height=\detvisheight]{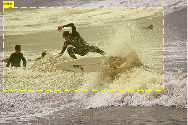}}
     {} 
&
\psubf{\includegraphics[height=\detvisheight]{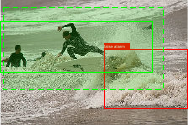}}
     {}
\\
\hline
\labeltext{ a multi colored towel in the cabinet}

&
\psubf{\includegraphics[height=\detvisheight]{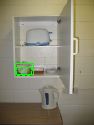}}
     {}    
&
\psubf{\includegraphics[height=\detvisheight]{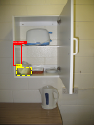}}
     {} 
&
\psubf{\includegraphics[height=\detvisheight]{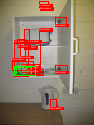}}
     {}
\end{tabular}
\caption{\failviscaption
\label{fig:det-vis-failures} }
\end{figure}
\arxvspace{-6pt}
\vspace*{\fill}
\end{centering}

\clearpage

\vspace*{\fill}
\begin{figure}[H]
\centering
\begin{tabular}{c|ccc}
\centering

\large Text phrases
&
\large DBNet
&
\large DenseCap
&
\large SCRC

\\
\hline
\labeltext{a black and white cat}
&
\psubf{\includegraphics[height=\detvisheight]{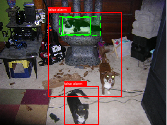}}
     {}    
&
\psubf{\includegraphics[height=\detvisheight]{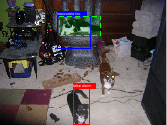}}
     {} 
&
\psubf{\includegraphics[height=\detvisheight]{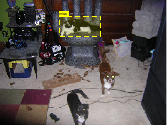}}
     {}
\\
\hline
\labeltext{a buckle is on the collar}
&
\psubf{\includegraphics[height=\detvisheight]{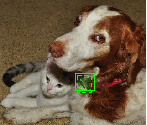}}
     {}    
&
\psubf{\includegraphics[height=\detvisheight]{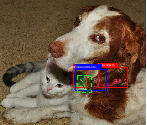}}
     {} 
&
\psubf{\includegraphics[height=\detvisheight]{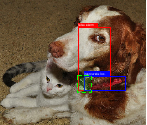}}
     {}
\\
\hline
\labeltext{a black shirt}
&
\psubf{\includegraphics[height=\detvisheight]{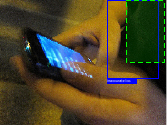}}
     {}    
&
\psubf{\includegraphics[height=\detvisheight]{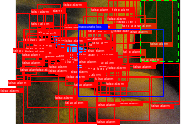}}
     {} 
&
\psubf{\includegraphics[height=\detvisheight]{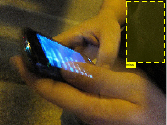}}
     {}
\end{tabular}
\repeatcaption{fig:det-vis-failures}{\failviscaption}
\end{figure}
\vspace*{\fill}
\clearpage

\vspace*{\fill}
\begin{figure}[H]
\centering
\begin{tabular}{c|ccc}
\centering

\large Text phrases
&
\large DBNet
&
\large DenseCap
&
\large SCRC

\\
\hline
\labeltext{a baseball tee}
&
\psubf{\includegraphics[height=\detvisheight]{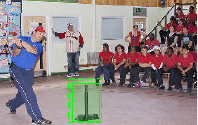}}
     {}    
&
\psubf{\includegraphics[height=\detvisheight]{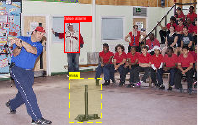}}
     {} 
&
\psubf{\includegraphics[height=\detvisheight]{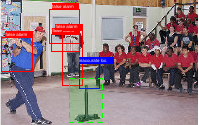}}
     {}
\\
\hline
\labeltext{airplane parked on tarmac}
&
\psubf{\includegraphics[height=\detvisheight]{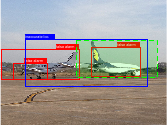}}
     {}    
&
\psubf{\includegraphics[height=\detvisheight]{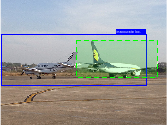}}
     {} 
&
\psubf{\includegraphics[height=\detvisheight]{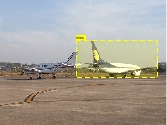}}
     {}
\\
\hline
\labeltext{a 2 toned blue winter jacket}
&
\psubf{\includegraphics[height=\detvisheight]{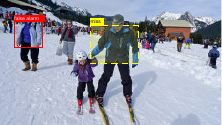}}
     {}    
&
\psubf{\includegraphics[height=\detvisheight]{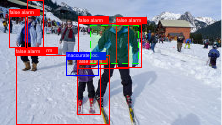}}
     {} 
&
\psubf{\includegraphics[height=\detvisheight]{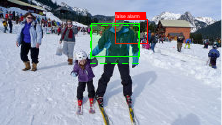}}
     {}
\end{tabular}
\repeatcaption{fig:det-vis-failures}{\failviscaption}
\end{figure}
\vspace*{\fill}

\clearpage

\end{landscape}

\endgroup

\section{Precision-recall curves}
\label{sec:more-pr}

\arxvspace{-6pt}

We show precision-recall curves for both global average precision (gAP) (Section~\ref{sec:gAP-pr}) and mean average precision (mAP) (Section~\ref{sec:mAP-pr}) calculation. 

\arxvspace{-6pt}

\subsection{Phrase-independent precision-recall curves}
\label{sec:gAP-pr}

\arxvspace{-6pt}

\begin{figure}[H]
\small
\centering
\begin{tabular}{ccc}
\subf{\includegraphics[width=0.3\textwidth]{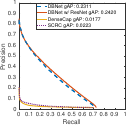}}
     {}
&
\subf{\includegraphics[width=0.3\textwidth]{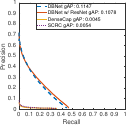}}
     {}
&
\subf{\includegraphics[width=0.3\textwidth]{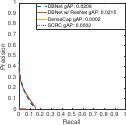}}
     {}
 \\
 Level 0, IoU@0.3 &  Level 0, IoU@0.5  &   Level 0, IoU@0.7  \\ \\
\subf{\includegraphics[width=0.3\textwidth]{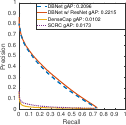}}
     {}
&
\subf{\includegraphics[width=0.3\textwidth]{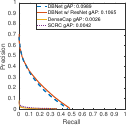}}
     {}
&
\subf{\includegraphics[width=0.3\textwidth]{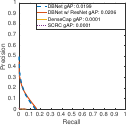}}
     {}
     \\
 Level 1, IoU@0.3 &  Level 1, IoU@0.5  &   Level 1, IoU@0.7  \\ \\
\subf{\includegraphics[width=0.3\textwidth]{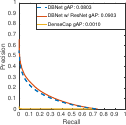}}
     {}
&
\subf{\includegraphics[width=0.3\textwidth]{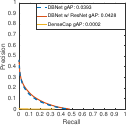}}
     {}
&
\subf{\includegraphics[width=0.3\textwidth]{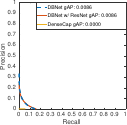}}
     {}
     \\
 Level 2, IoU@0.3 &  Level 2, IoU@0.5  &   Level 2, IoU@0.7  \\ \\
\end{tabular}
\unitcut\unitcut
\caption{Phrase-independent precision-recall curves for calculating gAP.}
\label{fig:global-pr}
\unitcut\unitcut\unitcut
\end{figure}

We reported precision-recall curves for different query set under different IoU threshold using the detection results for all test cases in Figure~\ref{fig:global-pr}. 
gAP was computed based on these precision-recall curves. 

\arxvspace{-6pt}

\clearpage

\subsection{Phrase-dependent precision-recall curves}
\label{sec:mAP-pr}

\begin{figure}[H]
\small
\centering
\begin{tabular}{ccc}
\subf{\includegraphics[width=0.3\textwidth]{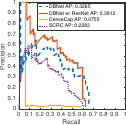}}
     {}
&
\subf{\includegraphics[width=0.3\textwidth]{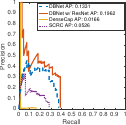}}
     {}
&
\subf{\includegraphics[width=0.3\textwidth]{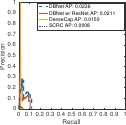}}
     {}
 \\
 Level 0, IoU@0.3 &  Level 0, IoU@0.5  &   Level 0, IoU@0.7  \\ \\
 \subf{\includegraphics[width=0.3\textwidth]{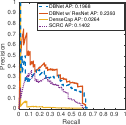}}
     {}
&
\subf{\includegraphics[width=0.3\textwidth]{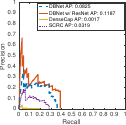}}
     {}
&
\subf{\includegraphics[width=0.3\textwidth]{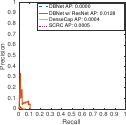}}
     {}
     \\
 Level 1, IoU@0.3 &  Level 1, IoU@0.5  &   Level 1, IoU@0.7  \\ \\
 \subf{\includegraphics[width=0.3\textwidth]{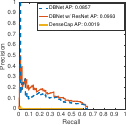}}
     {}
&
\subf{\includegraphics[width=0.3\textwidth]{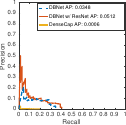}}
     {}
&
\subf{\includegraphics[width=0.3\textwidth]{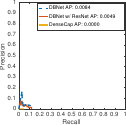}}
     {}
     \\
 Level 2, IoU@0.3 &  Level 2, IoU@0.5  &   Level 2, IoU@0.7  \\ \\
 
\end{tabular}
\caption{Precision-recall curves for text phrase ``head of a person''.
\label{fig:pr-curve-source-11} }
\end{figure}

\begin{figure}[H]
\small
\centering
\begin{tabular}{ccc}
\subf{\includegraphics[width=0.3\textwidth]{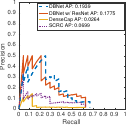}}
     {}
&
\subf{\includegraphics[width=0.3\textwidth]{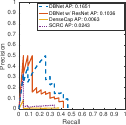}}
     {}
&
\subf{\includegraphics[width=0.3\textwidth]{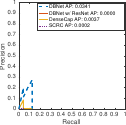}}
     {}
 \\
 Level 0, IoU@0.3 &  Level 0, IoU@0.5  &   Level 0, IoU@0.7  \\ \\
 \subf{\includegraphics[width=0.3\textwidth]{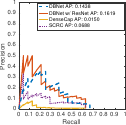}}
     {}
&
\subf{\includegraphics[width=0.3\textwidth]{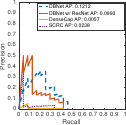}}
     {}
&
\subf{\includegraphics[width=0.3\textwidth]{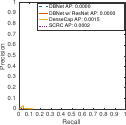}}
     {}
     \\
 Level 1, IoU@0.3 &  Level 1, IoU@0.5  &   Level 1, IoU@0.7  \\ \\
 \subf{\includegraphics[width=0.3\textwidth]{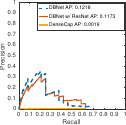}}
     {}
&
\subf{\includegraphics[width=0.3\textwidth]{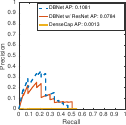}}
     {}
&
\subf{\includegraphics[width=0.3\textwidth]{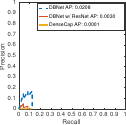}}
     {}
\\
 Level 2, IoU@0.3 &  Level 2, IoU@0.5  &   Level 2, IoU@0.7  \\ \\
  
\end{tabular}
\caption{Precision-recall curves for text phrase ``a window on the building''.
\label{fig:pr-curve-source-22} }
\end{figure}

\begin{figure}[H]
\small
\centering
\begin{tabular}{ccc}
\subf{\includegraphics[width=0.3\textwidth]{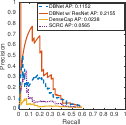}}
     {}
&
\subf{\includegraphics[width=0.3\textwidth]{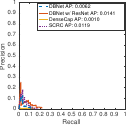}}
     {}
&
\subf{\includegraphics[width=0.3\textwidth]{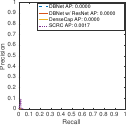}}
     {}
 \\
 Level 0, IoU@0.3 &  Level 0, IoU@0.5  &   Level 0, IoU@0.7  \\ \\
 \subf{\includegraphics[width=0.3\textwidth]{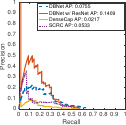}}
     {}
&
\subf{\includegraphics[width=0.3\textwidth]{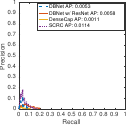}}
     {}
&
\subf{\includegraphics[width=0.3\textwidth]{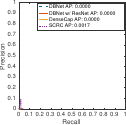}}
     {}
     \\
 Level 1, IoU@0.3 &  Level 1, IoU@0.5  &   Level 1, IoU@0.7  \\ \\
 \subf{\includegraphics[width=0.3\textwidth]{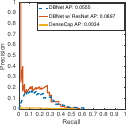}}
     {}
&
\subf{\includegraphics[width=0.3\textwidth]{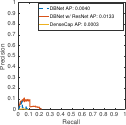}}
     {}
&
\subf{\includegraphics[width=0.3\textwidth]{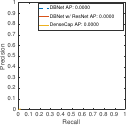}}
     {}
\\
 Level 2, IoU@0.3 &  Level 2, IoU@0.5  &   Level 2, IoU@0.7  \\ \\
  
\end{tabular}
\caption{Precision-recall curves for text phrase ``the water is calm''.
\label{fig:pr-curve-source-42} }
\end{figure}

\begin{figure}[H]
\small
\centering
\begin{tabular}{ccc}
\subf{\includegraphics[width=0.3\textwidth]{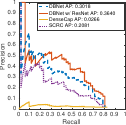}}
     {}
&
\subf{\includegraphics[width=0.3\textwidth]{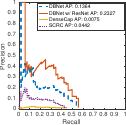}}
     {}
&
\subf{\includegraphics[width=0.3\textwidth]{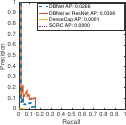}}
     {}
 \\
 Level 0, IoU@0.3 &  Level 0, IoU@0.5  &   Level 0, IoU@0.7  \\ \\
 \subf{\includegraphics[width=0.3\textwidth]{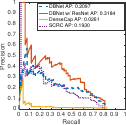}}
     {}
&
\subf{\includegraphics[width=0.3\textwidth]{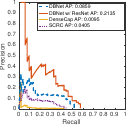}}
     {}
&
\subf{\includegraphics[width=0.3\textwidth]{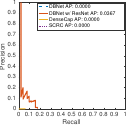}}
     {}
     \\
 Level 1, IoU@0.3 &  Level 1, IoU@0.5  &   Level 1, IoU@0.7  \\ \\
 \subf{\includegraphics[width=0.3\textwidth]{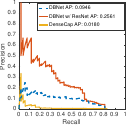}}
     {}
&
\subf{\includegraphics[width=0.3\textwidth]{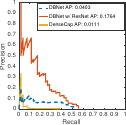}}
     {}
&
\subf{\includegraphics[width=0.3\textwidth]{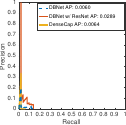}}
     {}
 \\
  Level 2, IoU@0.3 &  Level 2, IoU@0.5  &   Level 2, IoU@0.7  \\ \\
 
\end{tabular}
\caption{Precision-recall curves for text phrase ``man wearing blue jeans''.
\label{fig:pr-curve-source-72} }
\end{figure}

\begin{figure}[H]
\small
\centering
\begin{tabular}{ccc}
\subf{\includegraphics[width=0.3\textwidth]{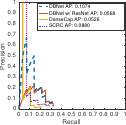}}
     {}
&
\subf{\includegraphics[width=0.3\textwidth]{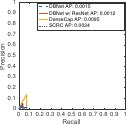}}
     {}
&
\subf{\includegraphics[width=0.3\textwidth]{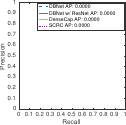}}
     {}
 \\
 Level 0, IoU@0.3 &  Level 0, IoU@0.5  &   Level 0, IoU@0.7  \\ \\
 \subf{\includegraphics[width=0.3\textwidth]{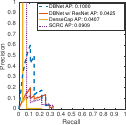}}
     {}
&
\subf{\includegraphics[width=0.3\textwidth]{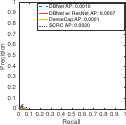}}
     {}
&
\subf{\includegraphics[width=0.3\textwidth]{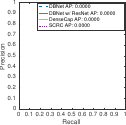}}
     {}
     \\
 Level 1, IoU@0.3 &  Level 1, IoU@0.5  &   Level 1, IoU@0.7  \\ \\
 \subf{\includegraphics[width=0.3\textwidth]{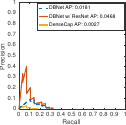}}
     {}
&
\subf{\includegraphics[width=0.3\textwidth]{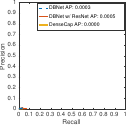}}
     {}
&
\subf{\includegraphics[width=0.3\textwidth]{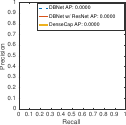}}
     {}
\\
 Level 2, IoU@0.3 &  Level 2, IoU@0.5  &   Level 2, IoU@0.7  \\ \\
  
\end{tabular}
\caption{Precision-recall curves for text phrase ``small ripples in the water''.
\label{fig:pr-curve-source-82} }
\end{figure}

We calculated precision-recall curves using various query sets under different IoU thresholds independently for different text phrases over the entire test set. 
mAP was computed based on these precision-recall curves.
We showed precision-recall curves for a few selected text phrases in Figure~\ref{fig:pr-curve-source-11}, \ref{fig:pr-curve-source-22}, \ref{fig:pr-curve-source-42}, \ref{fig:pr-curve-source-72}, and \ref{fig:pr-curve-source-82}. 

\clearpage

\clearpage

\clearpage

\clearpage

\clearpage

\end{document}